\documentclass[10pt,twocolumn,letterpaper]{article}

\usepackage{iccv}
\usepackage{times}
\usepackage{epsfig}
\usepackage{graphicx}
\usepackage{amsmath}
\usepackage{amssymb}
\usepackage[accsupp]{axessibility}  

\usepackage{kotex}
\usepackage{amsfonts,amsmath,amssymb}
\usepackage{booktabs}
\usepackage{colortbl}
\usepackage{comment}
\usepackage{dsfont}
\usepackage{graphicx}
\usepackage{makecell}
\usepackage{multirow}
\usepackage{rotating}
\usepackage{scalerel}
\usepackage{setspace} 
\usepackage[group-separator={,}]{siunitx}
\usepackage{tabularx}
\usepackage{wrapfig}
\usepackage{xspace}
\usepackage{capt-of}
\usepackage{enumitem}   
\setlist[itemize]{noitemsep,topsep=0pt,parsep=0pt,partopsep=0pt}
\setlist[enumerate]{noitemsep,topsep=0pt,parsep=0pt,partopsep=0pt}

\newcolumntype{L}{>{\raggedright\arraybackslash}X}
\newcolumntype{C}{>{\centering\arraybackslash}X}

\newcommand{\Eq}[1]  {Eq.\ (\ref{eq:#1})}
\newcommand{\Eqs}[1] {Eqs.\ (\ref{eq:#1})}
\newcommand{\Fig}[1] {Fig.\ \ref{fig:#1}}

\newcommand{\Tbl}[1]  {Table~\ref{tbl:#1}}
\newcommand{\Tbls}[1] {Tables~\ref{tbl:#1}}
\newcommand{\Sec}[1] {Sec.\ \ref{sec:#1}}

\newcommand{\Etal}   {{\textit{et al.}}}

\makeatletter
\DeclareRobustCommand\onedot{\futurelet\@let@token\@onedot}
\def\@onedot{\ifx\@let@token.\else.\null\fi\xspace}

\def\etal{\emph{et al}\onedot}
\makeatother

\newcommand{\parahead}[1]{\noindent\textbf{#1}.\ }

\definecolor{Goldenrod}{rgb}{0.85, 0.65, 0.13}
\definecolor{Orchid}{rgb}{0.85, 0.44, 0.84}
\definecolor{LimeGreen}{rgb}{0.2, 0.8, 0.2}
\definecolor{OliveGreen}{rgb}{0.25, 0.49, 0.19}
\definecolor{SkyBlue}{rgb}{0.53, 0.81, 0.92}
\definecolor{LightCyan}{rgb}{0.88,1,1}
\definecolor{LightGoldenrod}{rgb}{1.00, 0.93, 0.55}
\definecolor{LightSalmon}{rgb}{1.00, 0.63, 0.48}

\usepackage[breaklinks=true,bookmarks=false]{hyperref}

\iccvfinalcopy 

\def\iccvPaperID{7206} 

\ificcvfinal\fi

\begin{document}

\title{CTRL-C: Camera calibration TRansformer with Line-Classification}

\author{Jinwoo Lee \\
        \small Kakao Brain 
        \and 
        \hspace{-1.1em}
        Hyunsung Go \\
        \hspace{-1.1em}
        \small Kookmin University
        \and 
        \hspace{-1.1em}
        Hyunjoon Lee \\
        \hspace{-1.1em}
        \small Kakao Brain
        \and 
        \hspace{-1.1em}
        Sunghyun Cho \\
        \hspace{-1.1em}
        \small POSTECH
        \and 
        \hspace{-1.1em}
        Minhyuk Sung \\
        \hspace{-1.1em}
        \small KAIST
        \and 
        \hspace{-1.1em}
        Junho Kim\thanks{Corresponding author: \href{mailto:junho@kookmin.ac.kr}{junho@kookmin.ac.kr}} \\
        \hspace{-1.1em}
        \small Kookmin University
}



\maketitle
\ificcvfinal\thispagestyle{empty}\fi

\graphicspath{{./figs/}}

\begin{abstract}
Single image camera calibration is the task of estimating the camera parameters from a single input image, such as the vanishing points, focal length, and horizon line. In this work, we propose Camera calibration TRansformer with Line-Classification (CTRL-C), an end-to-end neural network-based approach to single image camera calibration, which directly estimates the camera parameters from an image and a set of line segments. Our network adopts the transformer architecture to capture the global structure of an image with multi-modal inputs in an end-to-end manner. We also propose an auxiliary task of line classification to train the network to extract the global geometric information from lines effectively. Our experiments demonstrate that CTRL-C outperforms the previous state-of-the-art methods on the Google Street View and SUN360 benchmark datasets. \renewcommand\UrlFont{\color{magenta}\rmfamily\itshape}Code is available at \url{https://github.com/jwlee-vcl/CTRL-C}.
\end{abstract}

\section{Introduction}

Single image camera calibration is a task of inferring intrinsic and extrinsic camera parameters by analyzing the distortion in an input image caused by the perspective projection.
It is a key problem in various computer vision applications, including image rotation correction~\cite{Fischer:2015,Samii:2015:CGF}, photo upright adjustment~\cite{Lee:2014,Chaudhury:2014:ICIP}, metrology~\cite{Criminisi:2000,Zhu:2020:ECCV}, visual aesthetics assessment~\cite{Bhattacharya:2010:MM,Samii:2015:CGF}, object composition for augmented reality~\cite{Hoiem:2008:IJCV,Karsch:2011:SIGASIA,Karsch:2014:TOG}, and so on. 

\begin{figure}
    \centering
    \includegraphics[width=0.95\linewidth]{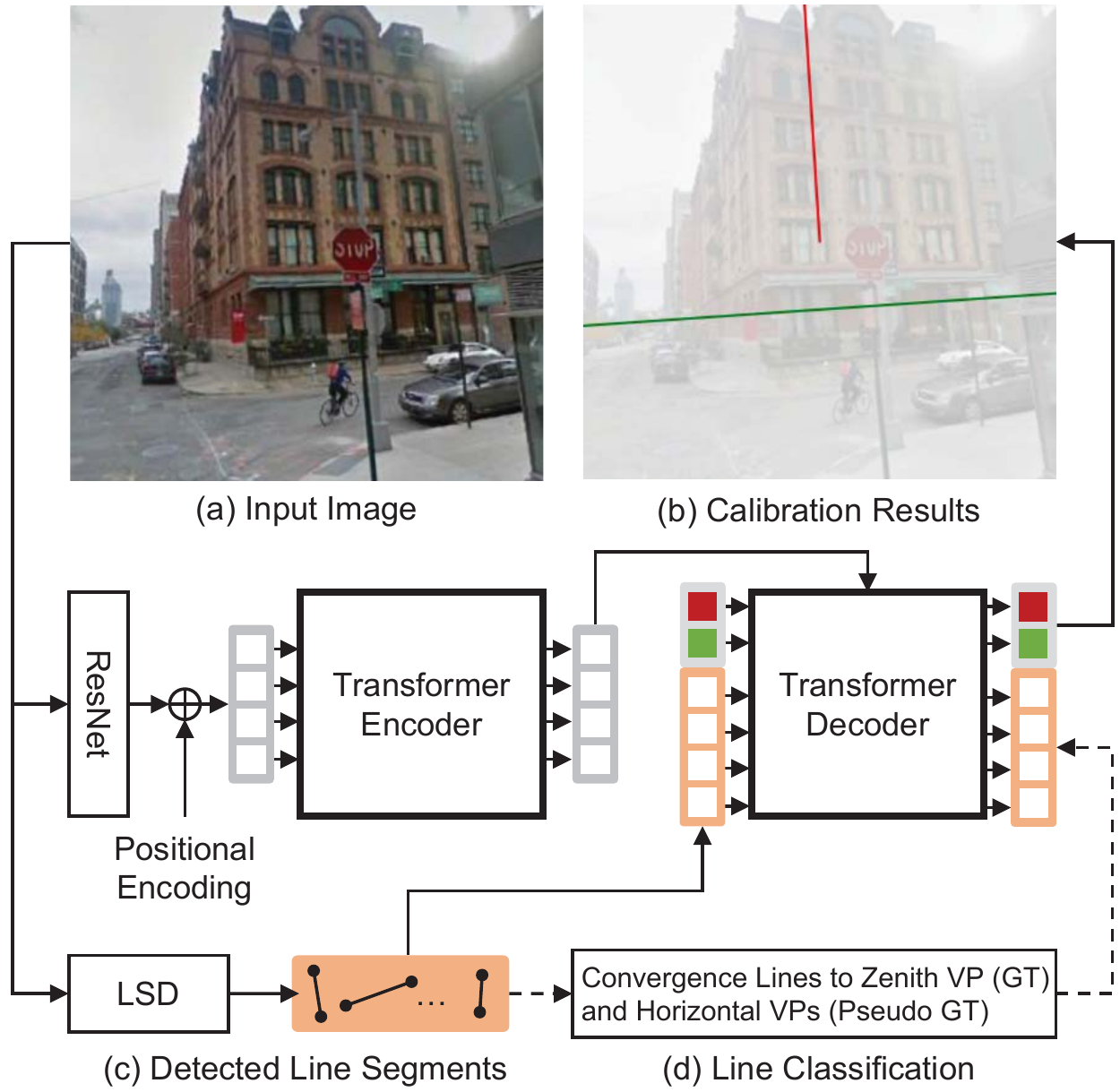} 
    \caption{
    Overview of CTRL-C.
    From a given image (a), CTRL-C predicts camera parameters including the zenith, FoV, and horizon line (b) by taking multi-modal cues; semantic ones from (a) and geometric ones from detected line segments in (c).
    While directly regressing the camera parameters, CTRL-C classifies line segments into vertical and horizontal convergence lines, as in (d).
    This auxiliary line classification task assists the network to have a better understanding of the geometric structure in the image and thus helps improve the accuracy of camera parameter prediction.
    }
    \label{fig:teaser}
\end{figure}

Single view geometry is highly related to projective geometric cues such as vanishing points (VPs) and the horizon line, which are the intersections of the world parallel lines and planes, respectively~\cite{MVG,Ma:3DV}. 
Hence, for single image camera calibration, a classical approach is first to detect line segments in an input image and then find inlier line segments that account for VPs and the horizon line using RANSAC or other sampling strategies~\cite{Schaffalitzky:2000,ECD:2012,Lee:2014,Simon:2018}.
However, this approach relies solely on lines and may degrade when inlier lines are falsely detected. 

Several deep learning-based approaches have recently been proposed to overcome such limitations~\cite{Workman:2016,Hold-Geoffroy:2018,Xian:2019}.
These methods directly infer the camera parameters from an input image using semantic cues learned by deep neural networks.
While they achieve more accurate calibration results than classical methods, their networks need to learn
geometric structures from an image without
any explicit supervision, limiting their performance.
Recently, a few deep learning-based methods~\cite{Zhai:2016,Lee:2020:ECCV} leveraging geometric and semantic cues have been proposed and achieved superior results. 
Nonetheless, their approaches to leveraging lines are still limited to conducting post-processing~\cite{Zhai:2016} or building multiple networks that are trained separately~\cite{Lee:2020:ECCV}.
Moreover, all the existing neural network-based approaches rely on \emph{convolutional} neural networks (CNNs), which are less effective in capturing long-term dependencies over an image, and consequently, global characteristics of an image such as the camera parameters.

To address the aforementioned issues, we propose to leverage \emph{transformers}~\cite{Transformer:2017,BERT:2019}, which have been recently adopted in multiple vision tasks~\cite{Survey:Transformer:2020,Survey:Transformer:2021}. We observe that transformers are particularly suitable for the following goals related to single image camera calibrations: i) exploiting both geometric and semantic cues, and ii) effectively learning their relationships and global contextual information of an image.
As transformers treat any types of input data as a sequence of tokens, both the semantic and geometric cues can be easily incorporated into a single end-to-end network.
Also, thanks to the attention mechanism, transformers can readily capture long-term dependencies across local image contexts and line segments.
Most importantly, we also empirically demonstrate that an \emph{auxiliary} task using the outputs of the transformers can even improve the performances by facilitating the interactions between these cues.

To this end, we propose a novel neural network named \emph{Camera calibration TRansformer with Line-Classification}, \emph{CTRL-C} in short, whose pipeline is illustrated in~\Fig{teaser}.
Our CTRL-C takes both an image and line segments as input and regresses the camera parameters based on the transformer encode-decoder architecture.
The input image is first fed to a ResNet~\cite{He:2016} and converted to a set of features of local patches. 
The transformer encoder then processes the image features with positional encoding to generate our \emph{semantic tokens}.
The line segments, extracted from the input image using the LSD algorithm~\cite{Gioi:2010}, are also mapped to \emph{geometric tokens}.
The subsequent transformer decoder aggregates both semantic and geometric tokens along with the queries for the camera parameters --- zenith VP, horizon line, and field of view (FoV) --- and learns the relationships across them. As an auxiliary task, the line segments are classified into convergence lines to either the zenith or horizontal VPs, which affects to improve the performance of camera parameter regression.

Our experimental results on the Google Street View~\cite{Lee:2020:ECCV} and SUN360~\cite{SUN360:2012} datasets show that CTRL-C outperforms previous single image camera calibration methods~\cite{Simon:2018,Workman:2016,Hold-Geoffroy:2018,Xian:2019,Lee:2020:ECCV} in multiple evaluation criteria. 
Even without the line classification task, our baseline transformer architecture already achieves competitive performance compared to the previous state-of-the-art (SotA) methods. Adopting the line classification task improves the performance, reducing the error of the up direction, pitch, roll, and FoV angles. Especially, CTRL-C increases the AUCs of the horizon line estimation with significant margins, from 83.12\% to 87.29\% (4.17\% gap) for the Google Street View test set and from 80.85\% to 85.45\% (4.6\% gap) for the SUN360 test set, compared to the results of the previous SotA methods.

\begin{figure*}[t!]
\includegraphics[width=1\linewidth]{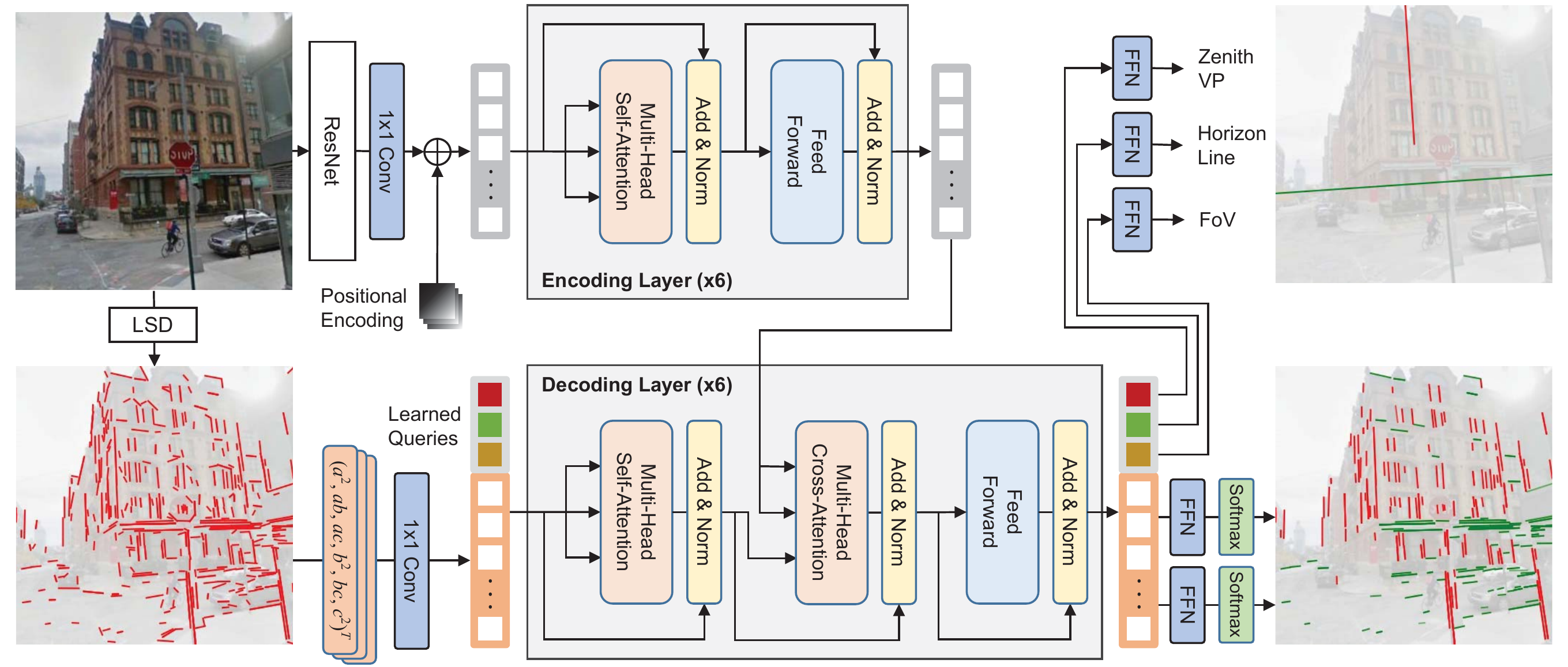} 
\caption{
Our network estimates parameters for camera calibration from an input image and a set of line segments. Image features are extracted with the ResNet backbone, flattened and positionally encoded, and then passed into a transformer encoder network. Learned query embedding vectors are fed into a transformer decoder network, alongside with the line embedding vectors, so that queries can attend to image and line features. Two auxiliary outputs, 
vertical and horizontal convergence line scores, 
are used to further guide the network to learn the scene geometry for calibration.}
\label{fig:overview}
\end{figure*}

\section{Related Work}
\paragraph{Single Image Camera Calibration}
Conventional methods typically find two or more VPs and estimate the camera's focal length and rotation from the relationship across them. Schaffalitzky and Zisserman~\cite{Schaffalitzky:2000} propose an automatic detection algorithm of VPs by grouping of planar geometric patterns. 
Tretyak~\Etal~\cite{ECD:2012} introduce an optimization framework that parses edge pixels into groups of the world parallel lines for different vanishing points and geometrically analyzes them.
Lee~\Etal~\cite{Lee:2014} present a maximum-a-posteriori (MAP)-based optimization method that estimates the VPs and a relative camera frame for a 3D scene, and then applies the camera calibration parameters for photo upright adjustment.
Simon~\Etal~\cite{Simon:2018} detect the zenith VP and the horizon by finding the maximally meaningful modes. 

Contrarily, recent neural network-based approaches propose to directly predict the camera parameters using semantic cues from an input image learned by convolutional networks.
Workman~\Etal~\cite{Workman:2016} propose the first neural network-based solution for estimating the horizon line from an image.
Hold-Geoffroy~\Etal~\cite{Hold-Geoffroy:2018} extend the idea to jointly predict both focal length and the horizon line, and provide perceptual studies of how humans perceive errors in camera calibration.
Xian~\Etal~\cite{Xian:2019} even step forward to predict the camera rotation as well, not by simply regressing the parameters, but by predicting the per-pixel 3D surface frames first and integrating the information.
While these neural approaches show improvement in accuracy compared to the conventional methods, they do not take any explicit geometric information of the input image, leaving room for further improvement.

A few neural approaches consider both geometric and semantic information for single image camera calibration~\cite{Zhai:2016,Lee:2020:ECCV}. 
Zhai~\Etal~\cite{Zhai:2016} obtain an initial horizon line using a CNN and improve it with a separate optimization step using line segments detected by the LSD algorithm~\cite{Gioi:2010}.
Lee~\Etal~\cite{Lee:2020:ECCV} propose a neural network-based geometric parsing framework for single image camera calibration, which uses both input image and lines detected from the image with the LSD algorithm~\cite{Gioi:2010}. They demonstrate that exploiting geometric information such as lines can help a network better understand the underlying perspective structure in an image and thus improve the camera calibration performance.
However, both of these methods are not end-to-end learnable. Zhai~\Etal's method requires a separate optimization step while Lee~\etal's method relies on two separately trained networks, and a complicated network structure to integrate line and image information.
Moreover, all the existing neural network-based approaches rely on CNNs, which are ineffective for capturing global information.

Inspired by \cite{Zhai:2016,Lee:2020:ECCV}, we also utilize lines as additional cue. 
In contrast to \cite{Zhai:2016,Lee:2020:ECCV}, however, we design CTRL-C based on the transformer architecture so that our network can be end-to-end learnable and effectively gather information from both semantic and geometric information.
We also show that our proposed auxiliary task of line classification significantly improves the camera calibration accuracy.

\paragraph{Image Transformers}
With remarkable success in natural language processing, transformers have been recently adopted to solve various tasks in computer vision. We refer the readers to~\cite{Survey:Transformer:2021,Survey:Transformer:2020} for thorough surveys. 
To name a few, Dosovitskiy~\Etal~\cite{ViT:2021} demonstrate that a standard transformer encoder treating images as a series of patches can classify images well without the image-specific induced bias. 
Carion~\Etal~\cite{DETR:2020} propose a transformer encoder-decoder architecture for object detection, called DETR, which predicts a set of object bounding boxes. 
Zhu~\Etal~\cite{DeformableDETR:2021} improve DETR~\cite{DETR:2020} with deformable attention modules both in accuracy and training time.
Xu~\Etal~\cite{LETR:2021} introduce a multi-scale transformer architecture for line segment detection, called LETR. 
VisualBERT~\cite{VisualBERT:2019}, ViLBERT~\cite{ViLBERT:2019:NeurIPS}, and their variants~\cite{LXMERT:2019:EMNLP,12-in-1:2020:CVPR,UNITER:2020:ECCV} also apply transformers to solve joint vision-and-language reasoning problems.
Along this direction, we also adopt the transformer in the single image calibration problem. Our main contribution is, however, in maximizing the advantage of the transformer architecture by providing image patches and line features as multimodal inputs. 
In contrast to \cite{DETR:2020,LETR:2021} that use transformers as a \emph{direct regressor}, e.g., for box parameters (DETR~\cite{DETR:2020}) and for line parameters (LETR~\cite{LETR:2021}),
we use the transformer to facilitate leveraging \emph{multi-modal cues}: image patches for semantic cues and line segments for geometric cues.

\section{Framework}

\Fig{overview} shows our framework. From a given input image, we estimate the vertical vanishing point (also known as the zenith VP), the horizon line, and the FoV for the focal length of the camera.
Our network has the following components: a CNN backbone network that extracts image features; an encoder transformer that encodes image features; a decoder transformer followed by FFNs that predicts all the outputs.

\subsection{Backbone Network}
We use ResNet (\texttt{resnet50} in \texttt{torchvision}) to compute image features \cite{He:2016}.
From a given input image of size $3 \times H_0 \times W_0$, we use the \texttt{block4} output of ResNet as image features of size $C \times H \times W$ where $H_0 = 32 H$ and $W_0 = 32 W$. Throughout our experiments we set $H_0$, $W_0$ and $C$ as 512, 512 and 2048, respectively.

\subsection{Encoder Network}
Our encoder architecture is similar to several previous works \cite{DETR:2020,ViT:2021,LETR:2021}; image features $F$ are first projected with a $1\times1$ convolution to have a smaller number of channels. The projected features of size $d \times H \times W$ are then spatially flattened and fed into the transformer encoder network. We set $d=256$ in our experiments. The encoder network consists of six self-attention blocks, and each block has eight attention heads. Positional encodings are added to features for each self-attention block; refer \cite{Transformer:2017,DETR:2020} for details of the architecture.

\subsection{Decoder Network}
\label{sec:decoder_network}
To estimate the calibration parameters, we query all the parameters to the transformer decoder.
Similarly to \cite{DETR:2020,LETR:2021}, all the queries for the zenith VP, horizon line, and FoV are decoded in parallel; we feed three $d$-dimensional vectors to the decoder as query embeddings. Multi-headed self-attention and cross-attention blocks are applied for several times to transform query embeddings to estimations of the parameters.

\paragraph{Utilizing Line Segments}
Although it is already possible to achieve SotA results only with the transformers and encoded image features (see \Sec{ablation}, \Tbls{test_googlestreetview} and~\ref{tbl:test_sun360}), we can further improve the estimation accuracy by utilizing line segments, especially for images with man-made structures. 

Many previous approaches~\cite{Tardif:2009,ECD:2012,Lee:2014} utilize line segments for camera calibration, mostly by detecting the zenith and horizontal VPs and classifying the lines into vertical and horizontal convergences (or none of the two).
Classification is often performed via energy minimization under some widely used assumptions \cite{Coughlan:2000,Schindler:2004}, and here we replace this optimization to a supervised classification problem. To the end, in addition to inferring calibration parameters, our network is trained to classify 
\emph{convergence lines} 
to the zenith and horizontal VPs in a set of line segments.

We first detect a set of line segments from a given input image \cite{Gioi:2010}. For each line segment, the corresponding line equation can be computed as a cross product of the two endpoints of the segment:
\begin{align}
    \mathbf{l} &= \left[\mathbf{p}_0\right]_\times\mathbf{p}_1,
\end{align}
where $\mathbf{p}_0$, $\mathbf{p}_1$ and $\mathbf{l}$ represent two endpoints of a line segment and its line equation, respectively. We sample at most 512 line segments, $L = \left\{ \mathbf{l}_0, \ldots, \mathbf{l}_{n<512} \right\}$, from the detected line segments. 
Directional ambiguities in line equations ($\mathbf{l}$ and $-\mathbf{l}$ represent the same line) are removed by taking upper triangular part of $\mathbf{ll}^T$ such that $\mathbf{s} = (a^2, ab, ac, b^2, bc, c^2)^T$ where $\mathbf{l} = (a, b, c)^T$.
We apply a $1\times1$ convolution to $\mathbf{s}$ to build $d$-dimensional line embedding vectors $\left\{ \mathbf{e}_0, \ldots \mathbf{e}_n \right\}$. 

Our network is then trained to classify input line segments into 
vertical and horizontal convergence lines 
(or none of the two) - lines that pass through the zenith VP and horizontal VPs, respectively. For this, we label each line segment $\mathbf{l}_i$ using the following equation:
\begin{align}
    c(\mathbf{l}_i, \mathbf{v}) &= \left\{
    \begin{array}{cl}
         1 & \mathrm{if~} d(\mathbf{l}_i, \mathbf{v}) \leq \delta_0 \\
         0 & \mathrm{if~} d(\mathbf{l}_i, \mathbf{v}) \geq \delta_1 \\
        -1 & \mathrm{otherwise}
    \end{array},
    \right.
\label{eq:classification_label}
\end{align}
where $\mathbf{v}$ represents the zenith VP or a horizontal VP. The point-line distance function $d(\cdot)$ is defined as:
\begin{align}
    d(\mathbf{l}, \mathbf{v}) &= \left| 
        \frac{\mathbf{v}^T\mathbf{l}}{\left\| \mathbf{l} \right\| \left\| \mathbf{v}\right\|}
    \right|,
\label{eq:line_vp_distance}
\end{align}
where $\left\| \cdot \right\|$ represents the $L_2$ norm of a vector. Throughout our experiments, we set $\delta_0$ and $\delta_1$ as $\sin(2^\circ)$ and $\sin(5^\circ)$, respectively.

\begin{figure}[t!]
\centering
\begin{tabular}{cc}
    \includegraphics[width=0.42\linewidth]{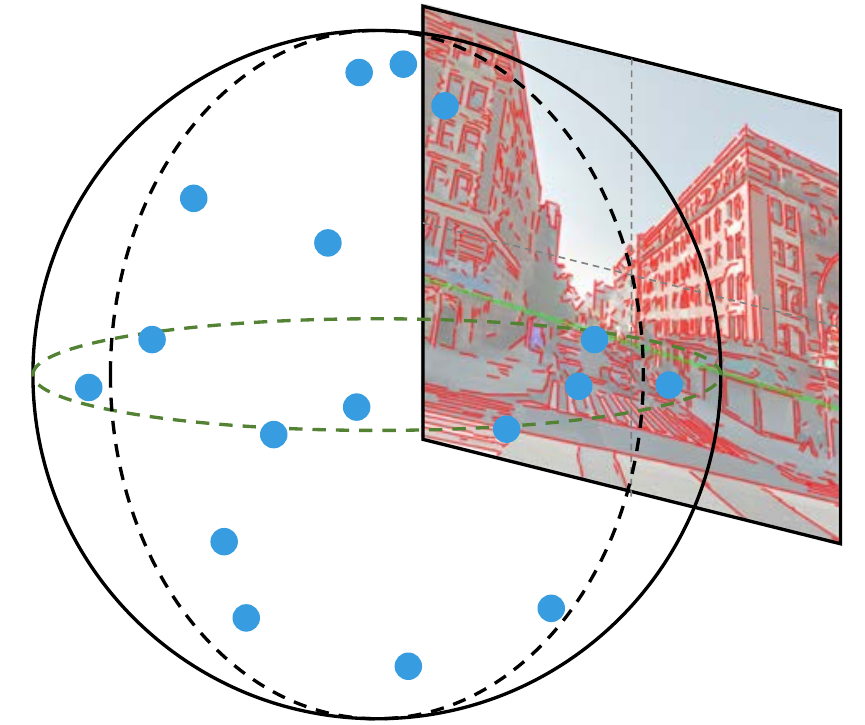} &
    \includegraphics[width=0.42\linewidth]{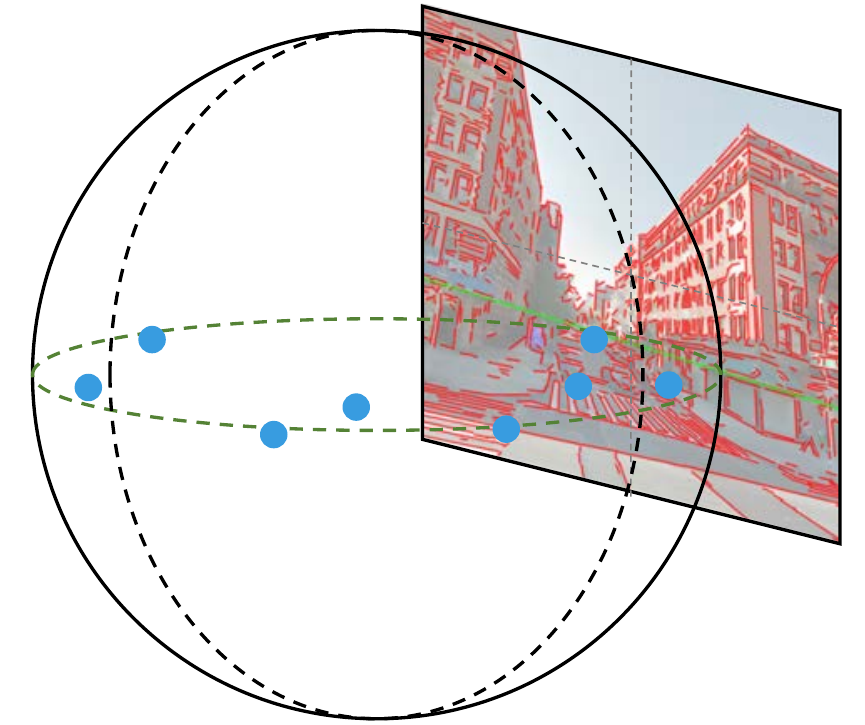} \\
    (a) & (b) \\
    \includegraphics[width=0.42\linewidth]{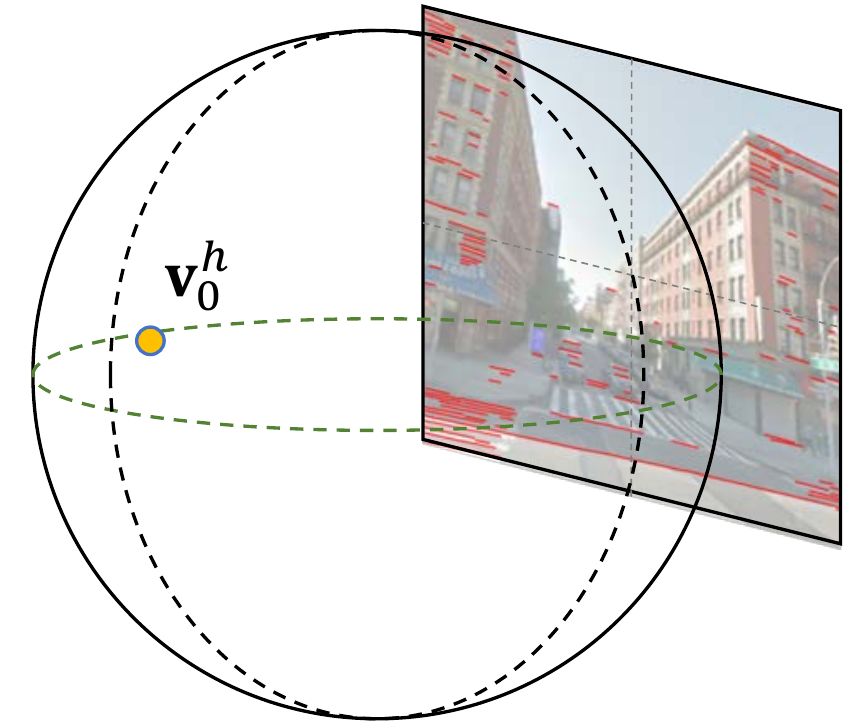} &
    \includegraphics[width=0.42\linewidth]{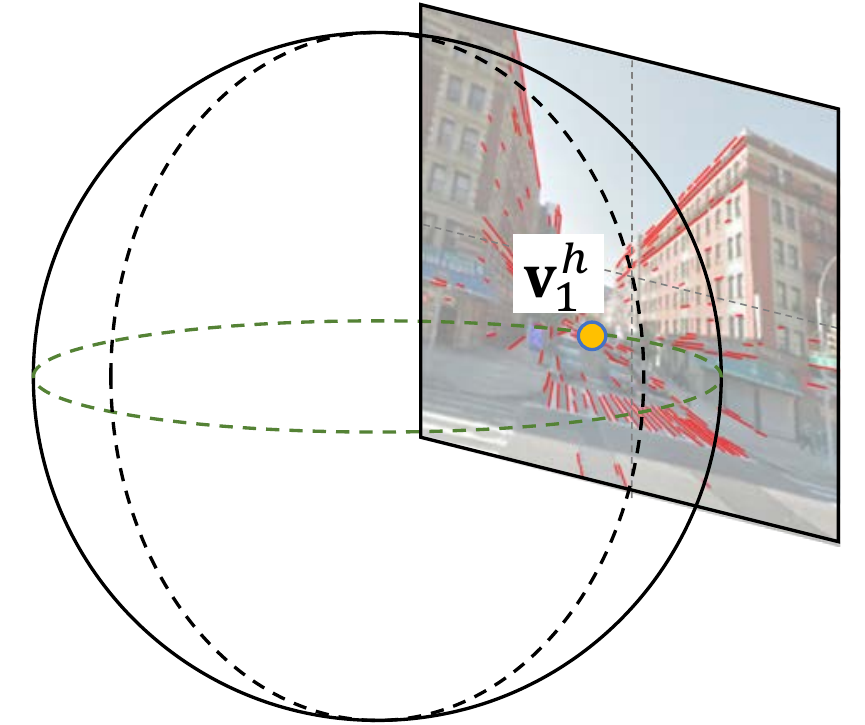} \\
    (c) & (d)
\end{tabular}

\caption{
Pseudo horizontal VP estimation process; (a) computing VP candidates with the intersection of different line pairs in $L$, (b) filtering the VP candidates in nearby the GT horizon line, (c)
selecting a VP candidate through which the most lines in $L$ pass nearby, (d) after filtering out the lines considered in (c) from $L$, selecting another VP candidate through which the most lines pass nearby. Pseudo horizontal VPs are set as the selected VP candidates in (c) and (d).
}
\label{fig:pseudo_vp}
\end{figure}

\paragraph*{Estimating Pseudo Horizontal VPs}
Once we have ground-truth (GT) values for the zenith VP and horizontal VPs, we can assign labels for 
vertical and horizontal convergence lines 
using \Eq{classification_label}. However, the Google Street View and SUN360 datasets only provide GT values for the zenith VP but not for horizontal VPs. 
To mitigate this issue, we create pseudo horizontal VPs;
1) extract a set of VP candidates, 2) filter out non-horizontal VP candidates, and 3) choose two VPs from the set of remaining candidates (see \Fig{pseudo_vp}).

To extract a set of initial VP candidates, we use a similar method with those of \cite{Lee:2014,Tardif:2009}.
From the set of line equations $L$, we randomly select a pair of line segments $(\mathbf{l}_0, \mathbf{l}_1)$ and compute their intersection points $\mathbf{v}$ as $\mathbf{v} = \mathbf{l}_0 \times \mathbf{l}_1$. We repeat this until we have the set of VP candidates 
$V_c$.

By definition, all horizontal VPs should be located on the horizon line. We remove the candidates from $V_c$ that are not near enough to the horizon line and keep the remains as the horizontal VP candidates
using the following equation:\hspace{-1.0em}
\begin{align}
    V_{\mathrm{in}} = \left\{ \mathbf{v} ~|~ d(\mathbf{h}, \mathbf{v}) < \delta ~\mathrm{and}~\mathbf{v} \in V_c \right\}, \label{eq:horizontal_vp_candidates}
\end{align}
where $\mathbf{h}$ represents the horizon line equation and $d(\cdot)$ is defined in \Eq{line_vp_distance}. 

From the horizontal VP candidates in $V_{\mathrm{in}}$, we select two dominants,
$\mathbf{v}^{h}_0$ and $\mathbf{v}^{h}_1$, 
as the pseudo horizontal VPs.
For each candidate in $V_{\mathrm{in}}$, we find line segments close enough to the candidate and compute the sum of their lengths:
\begin{align}
    \label{eq:pseudo_gt}
    m_i &= \sum_{\mathbf{l} \in L_\mathrm{in}^i} \mathrm{len}(\mathbf{l}), \mathrm{~where} \\
    \label{eq:convergence_lines}
    L_\mathrm{in}^i &= \left\{ \mathbf{l} ~|~ d(\mathbf{l}, \mathbf{v}_i) < \delta ~\mathrm{and}~\mathbf{v}_i \in V_{\mathrm{in}} \right\} 
\end{align}
where $\mathrm{len}(\mathbf{l})$ represents the length of the line segment whose line equation is  $\mathbf{l}$.
We select a pseudo horizontal VP $\mathbf{v}^{h}_0$ with the largest value of $m_i$, remove $L_\mathrm{in}^i$ from $L$, and find $\mathbf{v}^{h}_1$ by repeating the above process. 
We set $\delta$ in \Eqs{horizontal_vp_candidates} and \eqref{eq:convergence_lines} as $\sin(2.5^\circ)$ in our experiments.

The decoder network gets line embeddings as well as calibration queries as inputs ($n+3$ queries in total), transforms them, and feeds the results to the feed forward networks. Position embeddings are not added to line features, as lines contain positional information by construction.

\subsection{Feed-Forward Networks (FFNs)}
\label{sec:ffn}
We append separate FFNs to estimate calibration parameters $\mathbf{z}$, $\mathbf{h}$, and $f$, representing zenith VP, horizon line, and FoV, respectively.
To classify line embeddings into 
vertical and horizontal convergence lines, 
two additional FFNs are appended to line embeddings $\left\{ \mathbf{e}_0, \ldots, \mathbf{e}_{n-1} \right\}$, followed by softmax layers resulting $\left\{ s_0^z, \ldots, s_{n-1}^z \right\}$ and $\left\{ s_0^h, \ldots, s_{n-1}^h \right\}$, respectively, where $s^{\{z,h\}} \in [0, 1]$.

\subsection{Loss Functions}
\label{sec:loss}
Our network is trained with five loss terms:
\begin{align}
    l &= l_{z} + l_{h} + l_{f} + l_{zc} + l_{hc}.
\end{align}
For the zenith VP, we minimize angular distances between the GT points and their corresponding predictions:
\begin{align}
    l_{z} &= 1 - \left| \frac{ \mathbf{z}^T \Tilde{\mathbf{z}}}{\|\mathbf{z}\| \|\mathbf{\Tilde{z}}\|} \right|,
\end{align}
where $\mathbf{z}$ and $\Tilde{\mathbf{z}}$ represent a GT zenith VP and an estimate, respectively.
For the horizon line, we use a metric from \cite{Barinova:2010}.
Specifically, we compute intersection points between image boundaries and GT/estimated horizon lines.
Let $\mathbf{b}_l$ and $\Tilde{\mathbf{b}}_l$ be intersection points of the left image boundary and horizon lines $\mathbf{h}$ and $\Tilde{\mathbf{h}}$, respectively, and $\mathbf{b}_r$ and $\Tilde{\mathbf{b}}_r$ for the right image boundary. 
The loss function is defined as:
\begin{align}
    l_{h} &= \max \left( 
        \left\| \Tilde{\mathbf{b}_l} - \mathbf{b}_l \right\|_1, 
        \left\| \Tilde{\mathbf{b}_r} - \mathbf{b}_r \right\|_1
    \right),
\end{align}
where $\left\| \cdot \right\|_1$ represents the $L_1$ norm of a vector.
The loss function for the FoV parameter is defined as:
%
\begin{align}
    l_{f} &= \left| f - \Tilde{f} \right|.
\end{align}
%
As the classification losses of the vertical and horizontal convergence lines, 
$l_{zc}$ and $l_{hc}$, binary cross entropy loss is used:
\begin{align}
    l_{\{zc,hc\}} = -\frac{1}{n} \sum_i 
                    & \left\{ c_i^{\{z,h\}} \log s_i^{\{z,h\}} \right. \\ \nonumber
                    & \left. + \left(1 - c_i^{\{z,h\}}\right) \log \left(1 - s_i^{\{z,h\}}\right) \right\},
\end{align}
where $c_i^z$ and $c_i^h$ are defined using \Eq{classification_label} as follows:
\begin{align}
    c_i^z &= c(\mathbf{l}_i, \mathbf{z}) \\
    c_i^h &= \max \left\{ c(\mathbf{l}_i, \mathbf{v}^{h}_0), c(\mathbf{l}_i, \mathbf{v}^{h}_1) \right\}.
\end{align}
We remove any $c_i$ from computing losses if $c(\mathbf{l}_i, \cdot)$ is $-1$.


\begin{table*}[t!]
\renewcommand{\arraystretch}{0.9}
\caption{Quantitative evaluation results with Google Street View benchmark from Lee~\Etal~\cite{Lee:2020:ECCV}. Note that the results of previous methods are from Lee~\Etal~\cite{Lee:2020:ECCV} --- see Table 2 in their paper. The accuracy of FoV prediction is not provided for A-Contrario detection~\cite{Simon:2018}, DeepHorizon~\cite{Workman:2016}, and UprightNet~\cite{Xian:2019}, since they cannot predict FoV.
\emph{TR} indicates the case of ablating transformers but just using ResNet~\cite{He:2016} to directly predict the camera parameters.
}
\label{tbl:test_googlestreetview}
{\footnotesize
\begin{tabularx}{\textwidth}{>{\centering}m{0.3cm}|>{\centering}m{0.6cm}|>{\centering}m{0.6cm}|>{\centering}m{0.6cm}|CC|CC|CC|CC|C}
\toprule
\multicolumn{4}{l|}{\multirow{2}{*}{Method}} & \multicolumn{2}{c|}{Up Direction ($^\circ$) $\downarrow$}  & \multicolumn{2}{c|}{Pitch ($^\circ$) $\downarrow$} & \multicolumn{2}{c|}{Roll ($^\circ$) $\downarrow$} & \multicolumn{2}{c|}{FoV ($^\circ$) $\downarrow$}  & \multirow{2}{*}{\makecell{AUC\\($\%$) $\uparrow$}} \\
\cline{5-12}
\multicolumn{4}{l|}{} & Mean & Med. & Mean & Med. & Mean & Med. & Mean & Med. \\ 
\midrule
\multicolumn{4}{l|}{Upright~\cite{Lee:2014}}                 &  3.05 &  1.92 &  2.90 &  1.80 &  6.19 & 0.43 & 9.47 &  4.42 & 77.43 \\
\multicolumn{4}{l|}{A-Contrario~\cite{Simon:2018}}           &  3.93 & 1.85 &  3.51 &  1.64 & 13.98 &  0.52 &   -   &   -    & 74.25 \\
\multicolumn{4}{l|}{DeepHorizon~\cite{Workman:2016}}         &  3.58 &  3.01 &  2.76 &  2.12 &  1.78 &  1.67 &   -   &   -    & 80.29 \\
\multicolumn{4}{l|}{Perceptual~\cite{Hold-Geoffroy:2018}}    &  2.73 &  2.13 &  2.39 &  1.78 &  0.96 &  0.66 & 4.61 &  3.89  & 80.40 \\
\multicolumn{4}{l|}{UprightNet~\cite{Xian:2019}}             & 28.20 & 26.10 & 26.56 & 24.56 &  6.22 &  4.33 &   -   &   -   &   -   \\
\multicolumn{4}{l|}{GPNet~\cite{Lee:2020:ECCV}}              & 2.12 & 1.61 & 1.92 & 1.38 & 0.75 & \textbf{0.47} &  6.01 & 3.72 & 83.12 \\
\multicolumn{4}{l|}{\textbf{CTRL-C (Ours)}}                    & \textbf{1.80} & \textbf{1.52} & \textbf{1.58} & \textbf{1.31} & \textbf{0.66} & 0.53 & \textbf{3.59} & \textbf{2.72} & \textbf{87.29} \\
\midrule
Id & TR & $l_{vc}$ & $l_{hc}$ & \multicolumn{9}{c}{Ablation Study (TR: Transformers)} \\
\midrule
1 & & &                                 & 3.12 & 2.62 & 2.79 & 2.22 & 1.04 & 0.82 &  5.04 & 4.17 & 84.48 \\
2 & \checkmark & &                        & 2.17 & 1.84 & 1.87 & 1.51 & 0.82 & 0.60 & 3.71 & 2.88 & 85.65 \\
3 & \checkmark & \checkmark &  & 1.84 & 1.53 & 1.61 & 1.32 & 0.65 & \textbf{0.52} & \textbf{3.47} & \textbf{2.66} & 87.16 \\
4 & \checkmark & & \checkmark  & 2.05 & 1.75 & 1.75 & 1.43 & 0.83 & 0.63 & 3.83 & 3.00 & 86.09 \\
5 & \checkmark & \checkmark & \checkmark & \textbf{1.80} & \textbf{1.52} & \textbf{1.58} & \textbf{1.31} & \textbf{0.66} & 0.53 & 3.59 & 2.72 & \textbf{87.29} \\
\bottomrule
\end{tabularx}
}
\end{table*}

\begin{table*}[t!]
\renewcommand{\arraystretch}{0.9}
\caption{Quantitative evaluation results with SUN360 benchmark from Lee~\Etal~\cite{Lee:2020:ECCV}. Note that the results of previous methods from Lee~\Etal~\cite{Lee:2020:ECCV} --- see Table 2 in their paper. The accuracy of FoV prediction is not provided for A-Contrario detection~\cite{Simon:2018}, DeepHorizon~\cite{Workman:2016}, and UprightNet~\cite{Xian:2019}, since they cannot predict FoV. \emph{TR} indicates the case of ablating transformers but just using ResNet~\cite{He:2016} to directly predict the camera parameters.}
\label{tbl:test_sun360}
{\footnotesize
\begin{tabularx}{\textwidth}{>{\centering}m{0.3cm}|>{\centering}m{0.6cm}|>{\centering}m{0.6cm}|>{\centering}m{0.6cm}|CC|CC|CC|CC|C}
\toprule
\multicolumn{4}{l|}{\multirow{2}{*}{Method}} & \multicolumn{2}{c|}{Up Direction ($^\circ$) $\downarrow$}  & \multicolumn{2}{c|}{Pitch ($^\circ$) $\downarrow$} & \multicolumn{2}{c|}{Roll ($^\circ$) $\downarrow$} & \multicolumn{2}{c|}{FoV ($^\circ$) $\downarrow$}  & \multirow{2}{*}{\makecell{AUC\\($\%$) $\uparrow$}} \\
\cline{5-12}
\multicolumn{4}{l|}{} & Mean & Med. & Mean & Med. & Mean & Med. & Mean & Med. \\ 
\midrule
\multicolumn{4}{l|}{Upright~\cite{Lee:2014}}               &  3.43 &  1.43 &  3.03 &  1.13 & 6.85 & 0.47 &  8.62 & 3.21 & 79.16 \\
\multicolumn{4}{l|}{A-Contrario~\cite{Simon:2018}}         &  5.77 &  1.53 &  4.91 &  1.19 & 6.93 & 0.66 &   -   &  -   & 72.75 \\
\multicolumn{4}{l|}{DeepHorizon~\cite{Workman:2016}}       &  2.87 &  2.12 &  2.36 &  1.64 & 1.16 & 0.85 &   -   &  -   & 80.65 \\
\multicolumn{4}{l|}{Perceptual~\cite{Hold-Geoffroy:2018}}  &  2.54 &  1.93 &  2.11 &  1.49 & 1.06 & 0.77 & 5.29 & 3.93 & 80.85 \\
\multicolumn{4}{l|}{UprightNet~\cite{Xian:2019}}           & 34.72 & 34.67 & 35.31 & 33.72 & 4.92 & 2.88 &   -   &  -   &   -   \\
\multicolumn{4}{l|}{GPNet~\cite{Lee:2020:ECCV}}            &  2.33 &  \textbf{1.27} &  1.97 &  \textbf{0.96} & 0.97 & \textbf{0.51} & 5.66 & 3.16 & 80.07 \\
\multicolumn{4}{l|}{\textbf{CTRL-C (Ours)}}                  &  \textbf{1.91} &  1.57 &  \textbf{1.50} &  1.15 & \textbf{0.96} & 0.71 & \textbf{3.80} & \textbf{2.77} & \textbf{85.45} \\
\midrule
Id & TR & $l_{vc}$ & $l_{hc}$ & \multicolumn{9}{c}{Ablation Study (TR: Transformers)} \\
\midrule
1 & & &                                  & 2.34 & 1.83 & 1.90 & 1.41 & 1.09 & 0.80 & 4.60 & 3.55 & 83.94 \\
2 & \checkmark & &                       & 2.03 & 1.71 & 1.62 & 1.26 & 1.00 & 0.75 & 4.00 & 2.98 & 84.39 \\
3 & \checkmark & \checkmark &            & \textbf{1.80} & \textbf{1.45} & \textbf{1.44} & \textbf{1.05} & \textbf{0.88} & \textbf{0.63} & \textbf{3.76} & \textbf{2.68} & 85.09 \\
4 & \checkmark & & \checkmark            & 2.03 & 1.65 & 1.59 & 1.20 & 1.03 & 0.78 & 3.87 & 2.84 & 85.36 \\
5 & \checkmark & \checkmark & \checkmark & 1.91 & 1.57 & 1.50 & 1.15 & 0.96 & 0.71 & 3.80 & 2.77 & \textbf{85.45} \\
\bottomrule
\end{tabularx}
}
\end{table*}

\begin{figure}[t!]
    \centering
    \setlength\tabcolsep{1.5pt} 
    \begin{tabular}{cc}
    \includegraphics[width=0.45\linewidth]{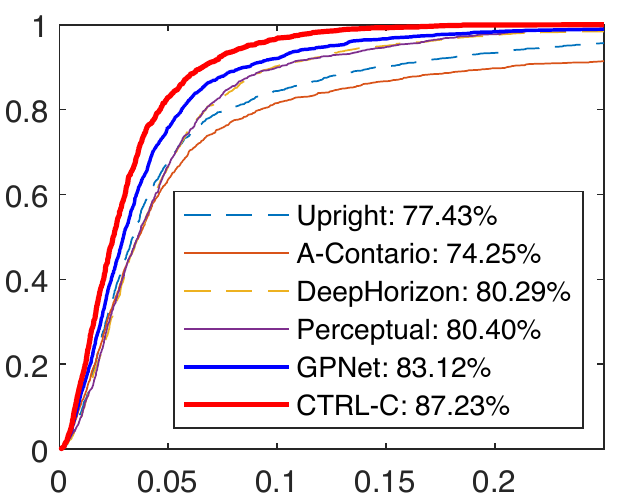} &
    \includegraphics[width=0.45\linewidth]{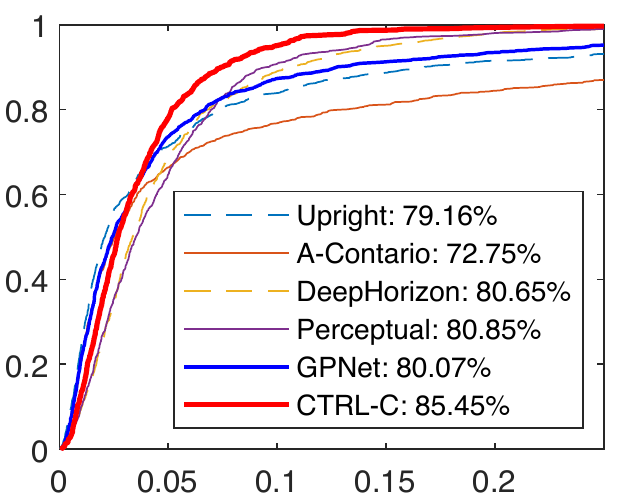} \\
    (a) & (b)
    \end{tabular}
    \caption{Comparison of the cumulative distributions of the horizon line error and their AUCs tested on Google Street View (a) and SUN360 (b). The AUCs are also reported in~\Tbls{test_googlestreetview} and \ref{tbl:test_sun360}.}
    \label{fig:auc}
\end{figure}

\begin{figure*}[t!]
    \centering
    \setlength\tabcolsep{1.5pt} 
    \begin{tabular}{ccccc}
    \includegraphics[width=0.19\linewidth]{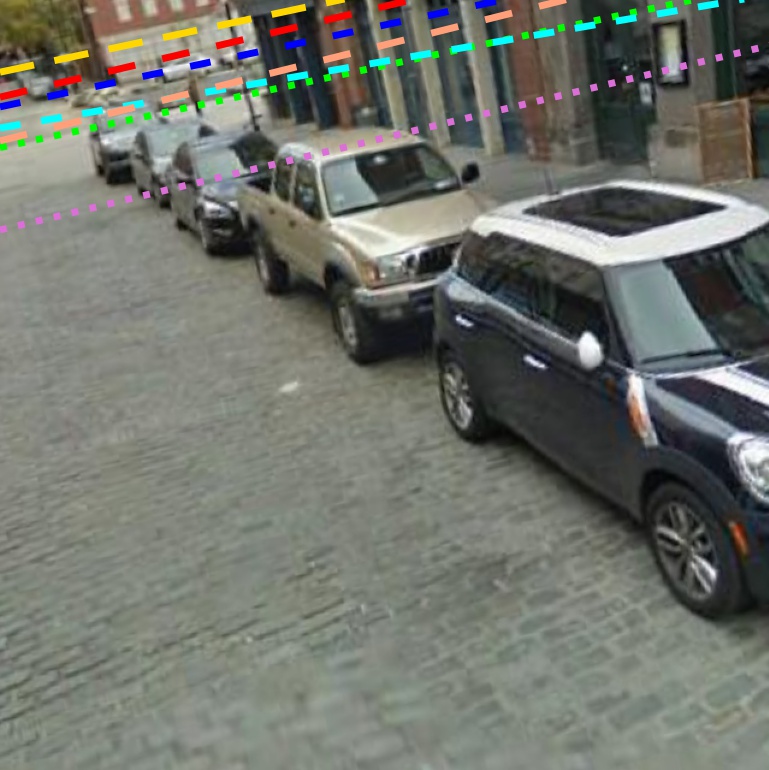} &
    \includegraphics[width=0.19\linewidth]{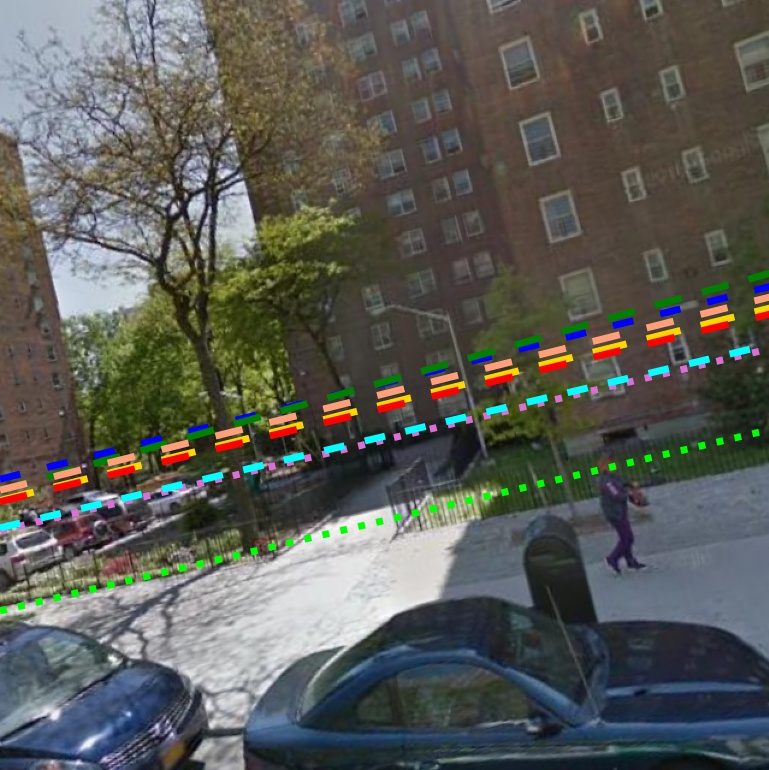} &
    \includegraphics[width=0.19\linewidth]{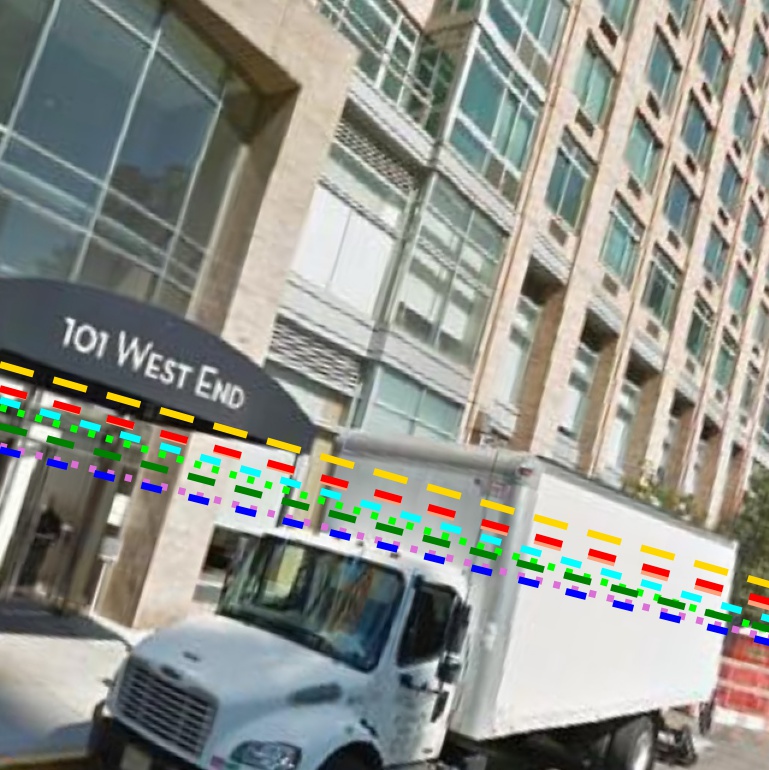} &
    \includegraphics[width=0.19\linewidth]{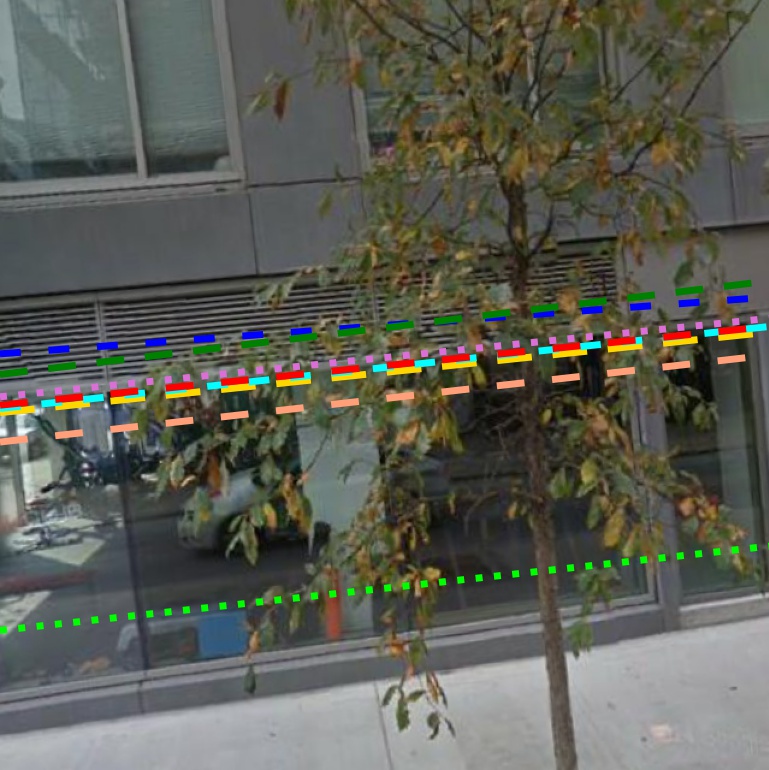} &
    \includegraphics[width=0.19\linewidth]{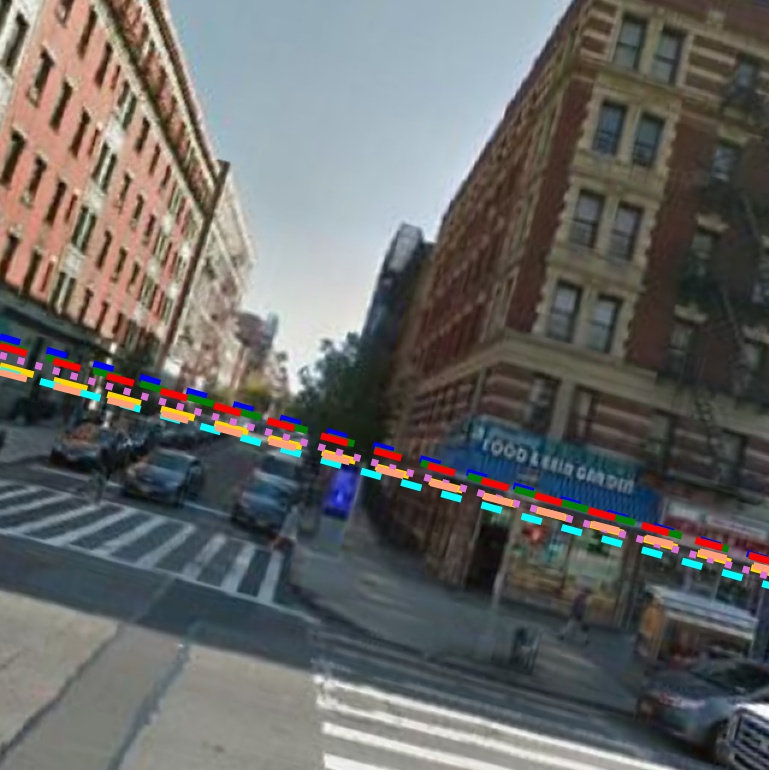} \\
    \includegraphics[width=0.19\linewidth]{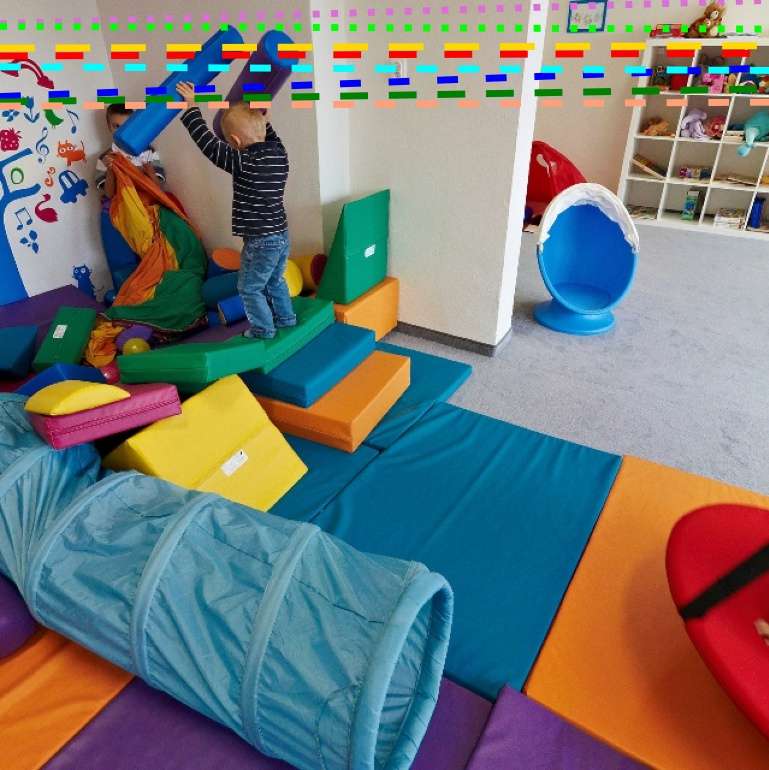} &
    \includegraphics[width=0.19\linewidth]{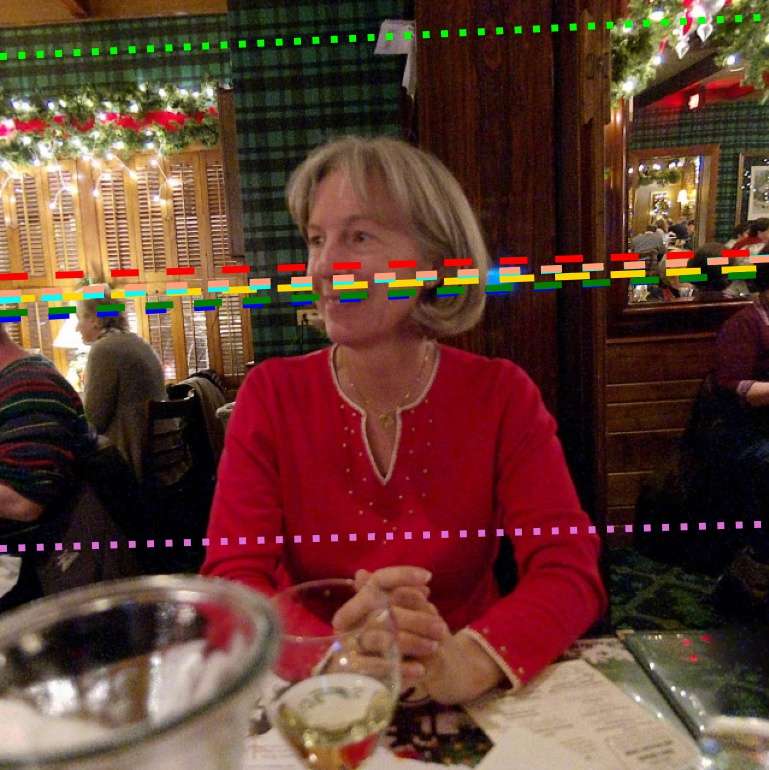} &
    \includegraphics[width=0.19\linewidth]{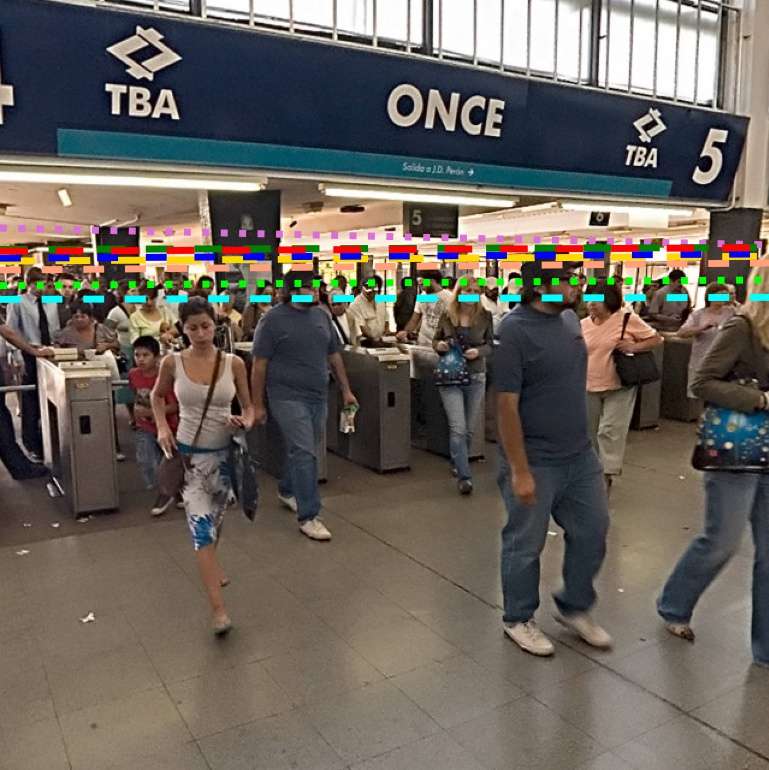} &
    \includegraphics[width=0.19\linewidth]{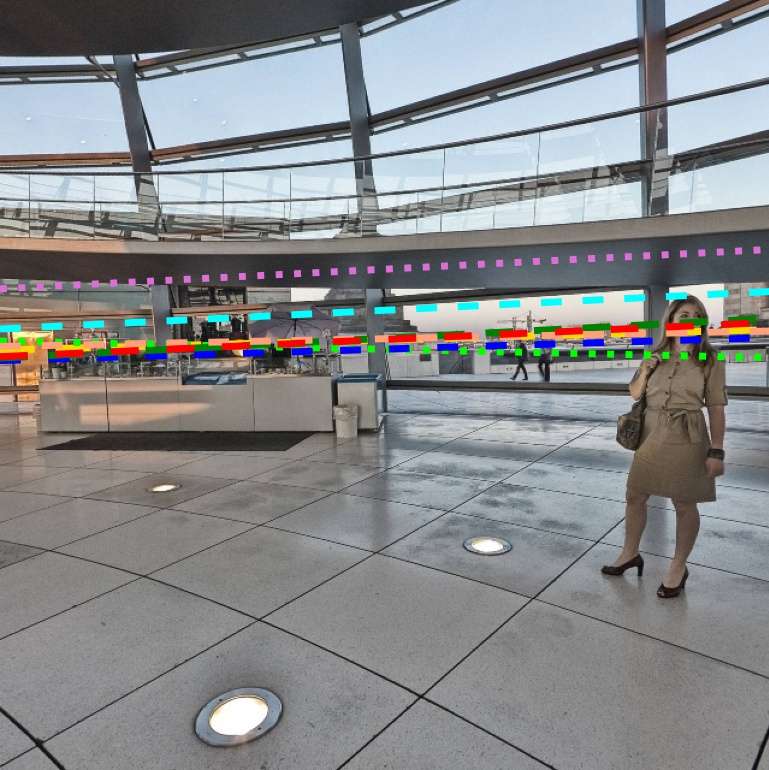} &
    \includegraphics[width=0.19\linewidth]{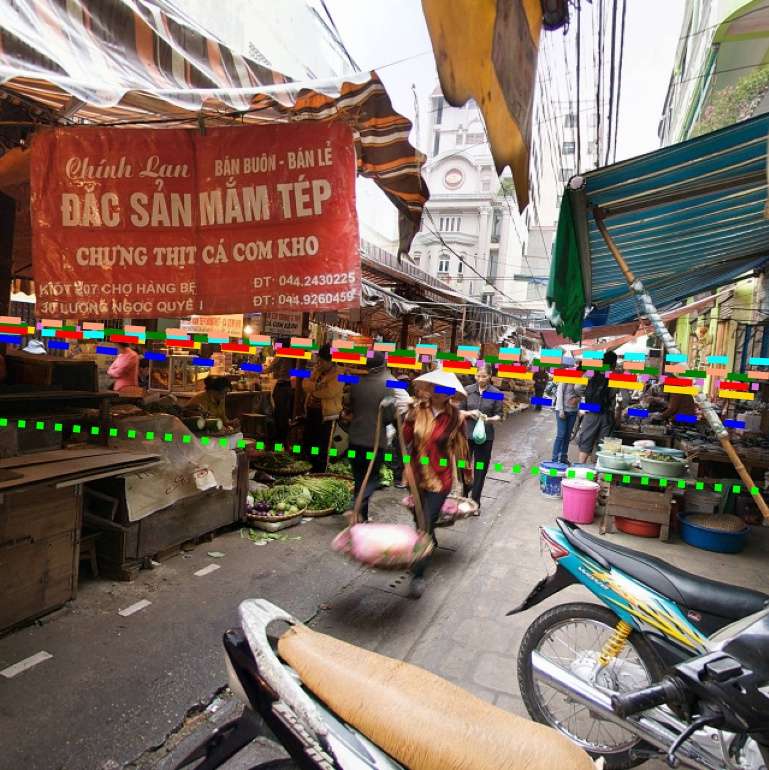} \\
    \end{tabular}
    \newcommand{\crule}[3][red]{\textcolor{#1}{\rule{#2}{#3} \rule{#2}{#3} \rule{#2}{#3} \rule{#2}{#3}}}
    {\scriptsize
    \begin{tabular}{llll}
    \crule[Goldenrod]{0.01\linewidth}{0.01\linewidth} Ground Truth & 
    \crule[Orchid]{0.01\linewidth}{0.01\linewidth} Upright \cite{Lee:2014} & 
    \crule[LimeGreen]{0.01\linewidth}{0.01\linewidth} A-Contario \cite{Simon:2018} & 
    \crule[blue]{0.01\linewidth}{0.01\linewidth} DeepHorizon \cite{Workman:2016} \\ 
    \crule[OliveGreen]{0.01\linewidth}{0.01\linewidth} Perceptual \cite{Hold-Geoffroy:2018} & 
    \crule[SkyBlue]{0.01\linewidth}{0.01\linewidth} GPNet \cite{Lee:2020:ECCV} &
    \crule[LightSalmon]{0.01\linewidth}{0.01\linewidth} ResNet &
    \crule[red]{0.01\linewidth}{0.01\linewidth} CTRL-C (Ours) \\
    \end{tabular}
    }
    \caption{Examples of horizon line prediction on the Google Street View test set (top row) and the SUN360 test set (bottom row).}
    \label{fig:horizon_line_predictions}
\end{figure*}

\section{Experiments}
\label{sec:experiments}

In the experiments, we evaluate the performance of our CTRL-C and other baselines using the benchmarks generated based on Google Street View (GSV)~\cite{GSVI} and SUN360~\cite{SUN360:2012} datasets. The benchmarks curated by Lee~\Etal~\cite{Lee:2020:ECCV} are constructed by
rectifying the original panoramic images with random samples of FoV, pitch, and roll in the ranges of $40 \sim 80 ^{\circ}$ ($40 \sim 90 ^{\circ}$ for SUN360), $-30 \sim 40^{\circ}$, and $-20 \sim 20^{\circ}$, respectively. Each sampled image's zenith VP and horizon line are then computed from sampled FoV, pitch and roll. The training and test sets contain 13,214 and 1,333 images for the GSV benchmark and 30,837 and 878 images for the SUN360 benchmark.
Refer to the supplementary material for more experimental and qualitative results on other datasets.

\subsection{Comparison}

We compare our method with the following previous methods: Upright~\cite{Lee:2014}, A-Contrario detection~\cite{Simon:2018}, DeepHorizon~\cite{Workman:2016}, Perceptual measure~\cite{Hold-Geoffroy:2018}, UprightNet~\cite{Xian:2019}, and GPNet~\cite{Lee:2020:ECCV}. For the comparison,
we refer to the results reported by GPNet~\cite{Lee:2020:ECCV},
where all the networks are retrained with the GSV dataset with the same ResNet architecture as the backbone (except for the UprightNet whose code is not publicly available). Note that A-Contrario detection~\cite{Simon:2018}, DeepHorizon~\cite{Workman:2016}, and UprightNet~\cite{Xian:2019} cannot predict FoV, and thus their results for FoV are not provided.

\Tbls{test_googlestreetview} and~\ref{tbl:test_sun360} show the quantitative results with the GSV~\cite{GSVI} dataset and the SUN360~\cite{SUN360:2012} dataset, respectively. We report the angle differences of up direction, pitch, roll, and FoV. For the horizon line, as done by 
Barinova~\Etal~\cite{Barinova:2010}, 
we calculate the area under the curve (AUC) of the cumulative distribution graph as shown in~\Fig{auc}, in which the $x$-axis indicates the distance between the predicted and GT horizon lines, and the $y$-axis indicates the percentage. Note that our CTRL-C outperforms all the baseline methods with significant margins in all the evaluation criteria, both in the GSV and SUN360 benchmarks. As shown in \Tbl{test_googlestreetview}, for the GSV benchmark, our CTRL-C reduces the mean errors of the up direction, pitch, roll, and FoV angles by 15.1\%, 17.7\%, 12.0\%, and 40.3\%, respectively, from the results of the previous SotA, GPNet~\cite{Lee:2020:ECCV}. The AUC of the horizon line errors is also improved from 83.12\% to 87.29\%, which has a 4.17\% gap. The results with the SUN360 benchmark in \Tbl{test_sun360} also show a similar trend.
Compared to the results of Perceptual~\cite{Hold-Geoffroy:2018}, 
the AUC of the horizon line errors is improved from 80.85\% to 85.45\%, which has a 4.6\% gap. 
\Fig{horizon_line_predictions} shows qualitative comparisons.

\begin{table}[t!]
\renewcommand{\arraystretch}{0.9}
\caption{Classification accuracies of convergence lines.}
\label{tbl:test_line_classification}
{\footnotesize
\begin{tabularx}{\linewidth}{C|C|C|C}
\toprule
\multirow{2}{*}{Training} & \multirow{2}{*}{Test} & \multicolumn{2}{c}{Accuracy  ($\%$) $\uparrow$} \\
\cline{3-4}
   &  & Vertical & Horizontal \\
\midrule
\multirow{2}{*}{GSV}         & GSV       & 99.73 & 93.35 \\
                                    & SUN360    & 93.76 & 85.18 \\
\midrule
\multirow{2}{*}{SUN360}      & GSV       & 97.96 & 92.16 \\
                                    & SUN360    & 99.49 & 91.54 \\
\bottomrule
\end{tabularx}
}
\end{table}

\subsection{Line Classification Results}
While the convergence line classification is 
an auxiliary task of our network, we also report the accuracy of the classification in~\Tbl{test_line_classification}.
When measuring the accuracy, 
the GT labels of convergence lines 
are given by the angle thresholds shown in~\Eq{classification_label}.
The results show very high accuracies and also good generalization capabilities even when the training and test datasets are different. For examples, training with GSV and testing with SUN360, the classification accuracies for 
vertical and horizontal convergence lines 
are 93.76\% and 85.18\%, respectively. 
In training with SUN360 and testing with GSV, the classification accuracies of 
vertical and horizontal convergence lines 
are better generalized to 97.96\% and 92.16\%, respectively.

\begin{figure*}[t!]
    \centering
    \setlength\tabcolsep{1.5pt} 
    \begin{tabular}{ccccccc}
    (a) &
    \includegraphics[width=0.15\linewidth]{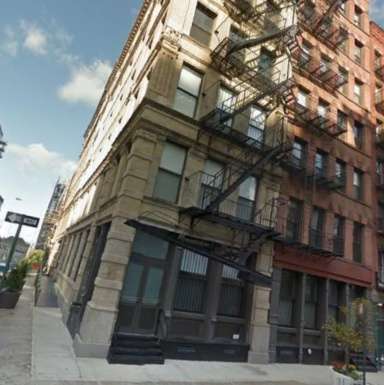} &
    \includegraphics[width=0.15\linewidth]{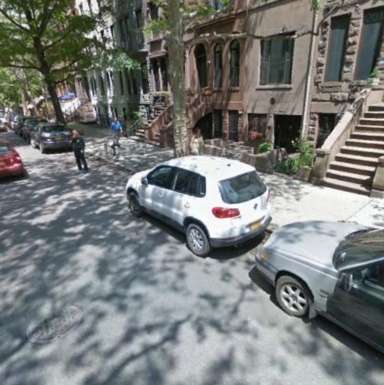} &
    \includegraphics[width=0.15\linewidth]{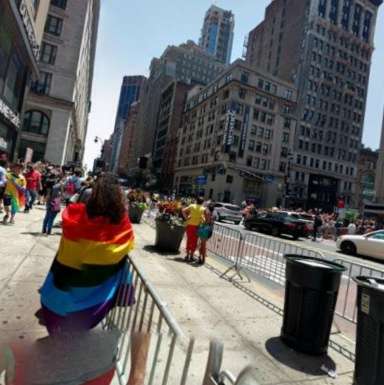} &
    \includegraphics[width=0.15\linewidth]{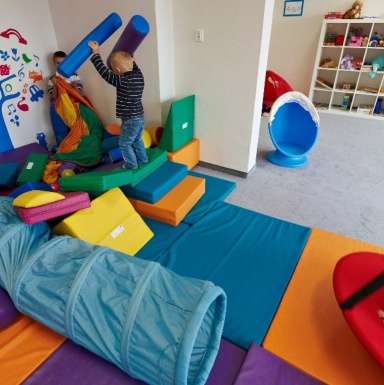} &
    \includegraphics[width=0.15\linewidth]{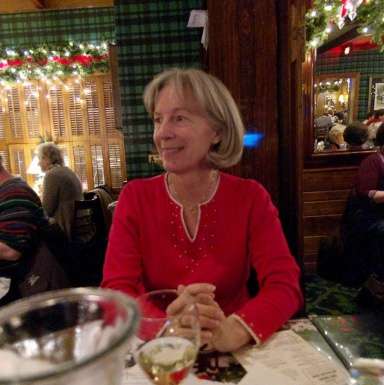} &
    \includegraphics[width=0.15\linewidth]{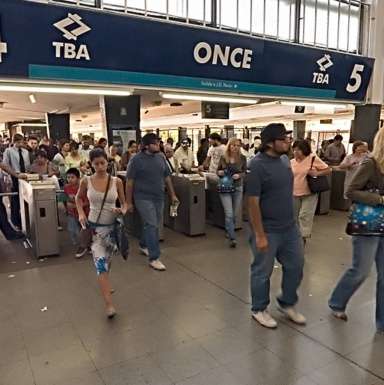} \\
    (b) &
    \includegraphics[width=0.15\linewidth]{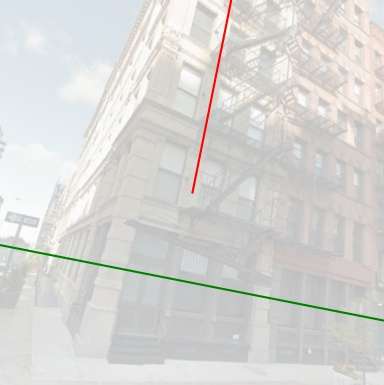} &
    \includegraphics[width=0.15\linewidth]{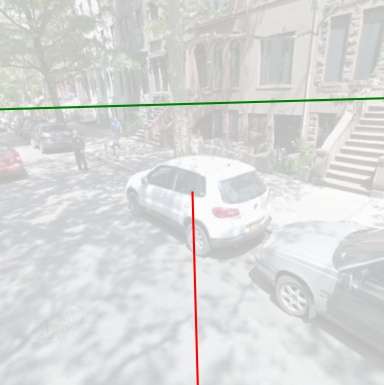} &
    \includegraphics[width=0.15\linewidth]{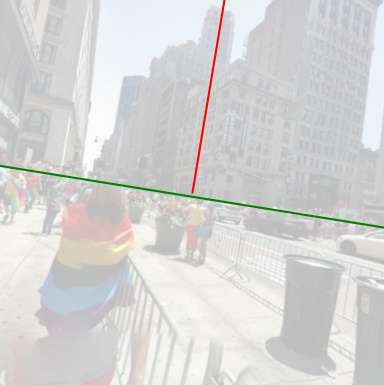} &
    \includegraphics[width=0.15\linewidth]{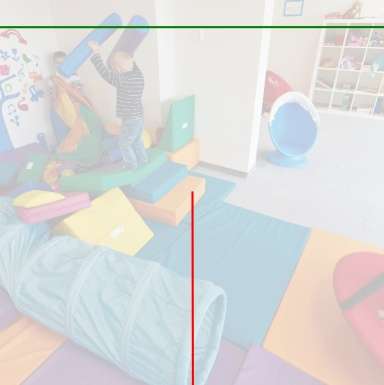} &
    \includegraphics[width=0.15\linewidth]{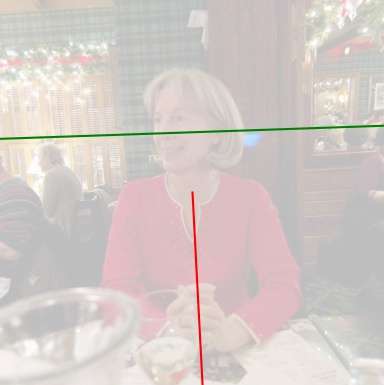} &
    \includegraphics[width=0.15\linewidth]{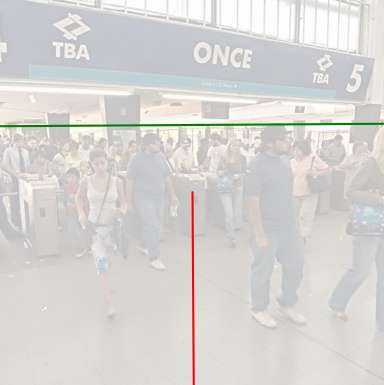} \\
    (c) &
    \includegraphics[width=0.15\linewidth]{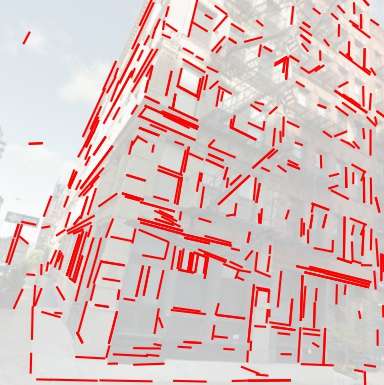} &
    \includegraphics[width=0.15\linewidth]{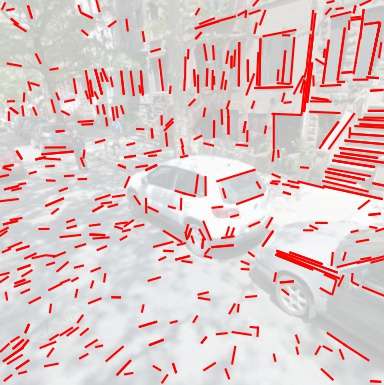} &
    \includegraphics[width=0.15\linewidth]{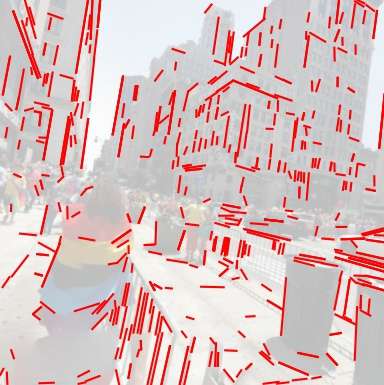} &
    \includegraphics[width=0.15\linewidth]{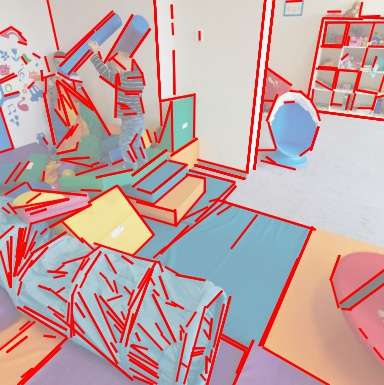} &
    \includegraphics[width=0.15\linewidth]{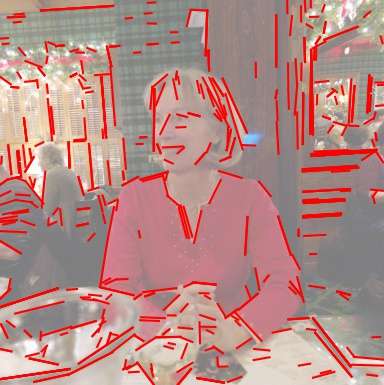} &
    \includegraphics[width=0.15\linewidth]{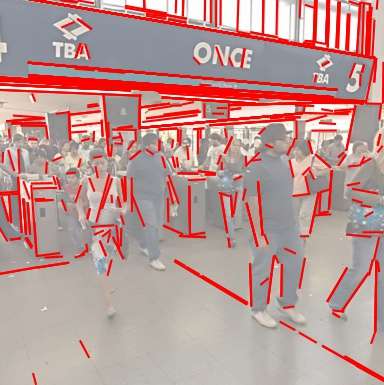} \\
    (d) &
    \includegraphics[width=0.15\linewidth]{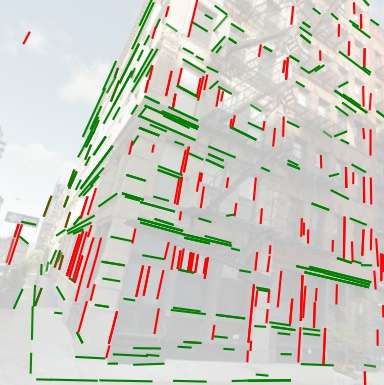} &
    \includegraphics[width=0.15\linewidth]{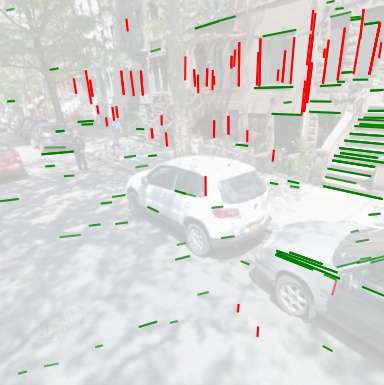} &
    \includegraphics[width=0.15\linewidth]{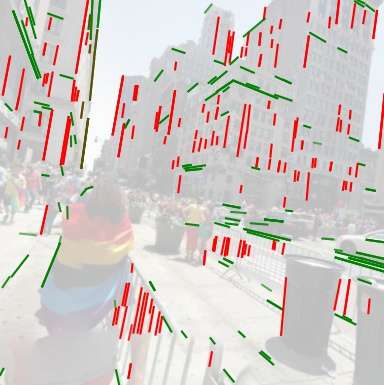} &
    \includegraphics[width=0.15\linewidth]{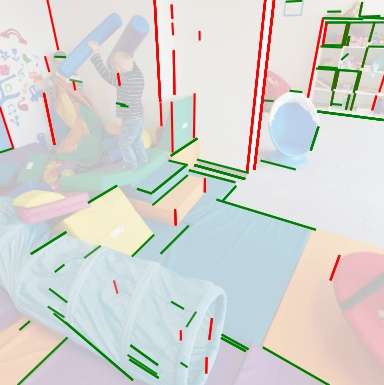} &
    \includegraphics[width=0.15\linewidth]{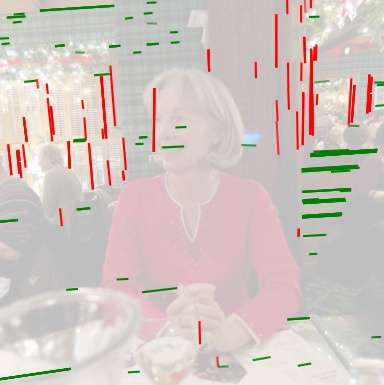} &
    \includegraphics[width=0.15\linewidth]{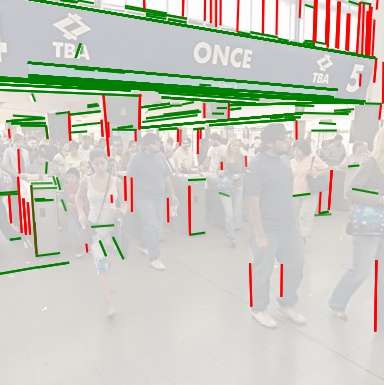} \\
    \end{tabular}
    \caption{More results with the Google Street View~\cite{Lee:2020:ECCV} test set (the first three columns) and SUN360~\cite{SUN360:2012} test set (the last three columns): (a) input image, (b) estimated horizon line (green) and vertical direction along with the zenith VP (red), 
    (c) detected lines with LSD \cite{Gioi:2010}, 
    (d) estimated vertical (red) and horizontal (green) convergence line segments of (c).
    }
    \label{fig:additional_results}
\end{figure*}

\Fig{additional_results} shows the qualitative results of the convergence line classification, along with the results of the horizon line and vertical direction prediction.
\Fig{additional_results} (a) and (c) show the input image and the line segments detected by LSD~\cite{Gioi:2010}, which are fed to our network. \Fig{additional_results} (b) shows the output vertical direction toward the zenith VP (red) and the horizon line (green). 
\Fig{additional_results} (d) illustrates the classification results of vertical and horizontal convergence lines performed as an auxiliary task; red and green indicate line segments classified as vertical and horizontal convergence lines (with a threshold 0.5), respectively. 
Notice that the classified line segments in (d) are vanishing toward the zenith VP or one of the two horizontal VPs laying on the horizon line in (b).

\subsection{Ablation Study} 
\label{sec:ablation}
We further demonstrate the impact of each component in our network by conducting an ablation study. The quantitative evaluation results are reported at the bottom of \Tbls{test_googlestreetview} and~\ref{tbl:test_sun360}. We first ablate the transformers~\cite{Transformer:2017} and just use ResNet~\cite{He:2016} as an image encoder that directly outputs the camera parameters. The network is trained with the same losses for the camera parameters ($l_z$, $l_h$, and $l_f$ in \Sec{loss}). As shown in the first row at the bottom of each table,
the performance is comparable but slightly worse than that of GPNet~\cite{Lee:2020:ECCV} in some criteria,
including the up direction, pitch, and roll. When the transformers are employed to process the patch features coming from ResNet (the second row), the performance becomes similar or even better than GPNet. Further improvements are made when feeding lines as additional inputs to the transformer decoders and
adding each classification loss for 
vertical and horizontal convergence lines 
($l_{vc}$ or $l_{hc}$, 
the third and fourth rows). Particularly, we found that 
the classifier for vertical convergence lines
makes meaningful improvements in the prediction of all camera parameters. The best performance can be achieved when 
the both classifiers for vertical and horizontal convergence lines
are used (the fifth row).

\subsection{Evaluation of Intermediate Decoding Layers}
The decoder of the transformer architecture is designed
to repeat
a layer including self-attention and cross-attention blocks --- typically six layers are used~\cite{DETR:2020,LETR:2021}. Since the inputs and outputs of the decoding layer are the same in the recursive structure, the camera parameters can also be predicted from \emph{each} layer by appending the FFN block used in the last layer. The network is then trained to predict calibration parameters from outputs of each decoder layer. \Tbl{test_layer} illustrates the quantitative evaluation results for the camera parameters predicted from each decoding layer. The results show that prediction power is increased as the image patch and line tokens are passed through more decoding layers, but the performance gain also converges as more layers are used.

\subsection{Results on Natural Scenes}
\begin{figure}
    \setlength{\tabcolsep}{1pt}
    \centering
    \begin{tabular}{ccc}
    \includegraphics[height=0.245\linewidth]{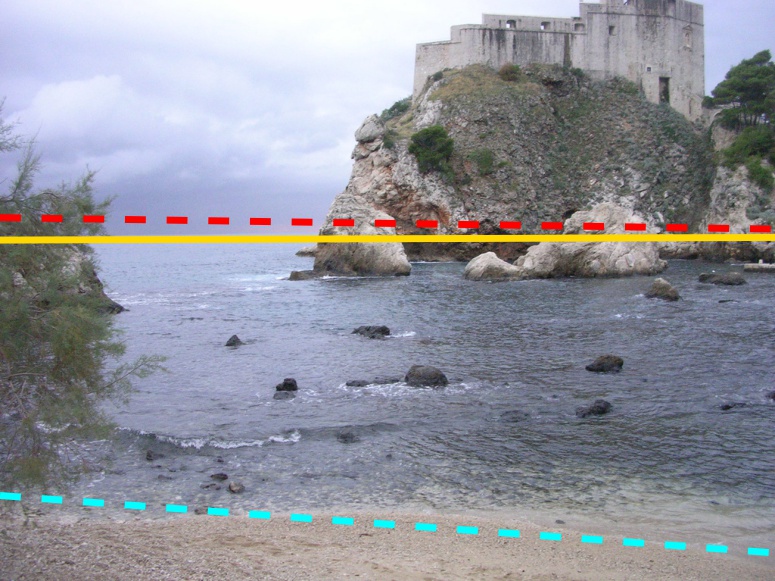} &
    \includegraphics[height=0.245\linewidth]{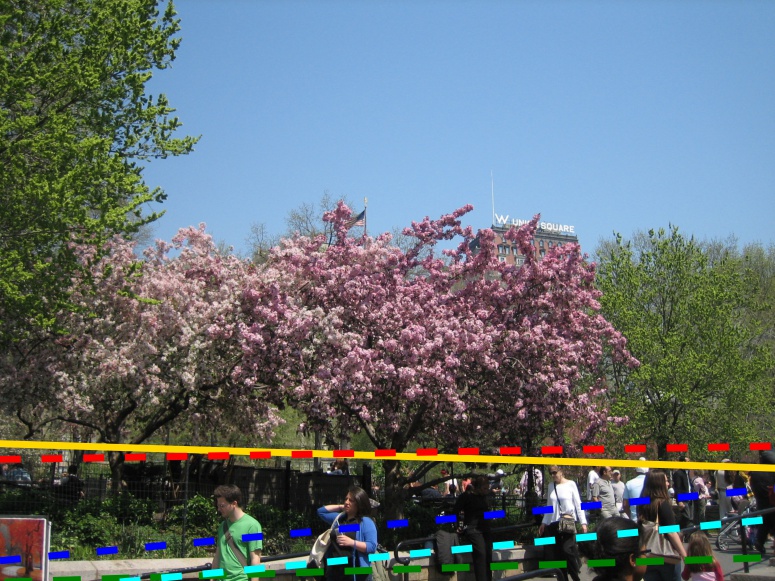} &
    \includegraphics[height=0.245\linewidth]{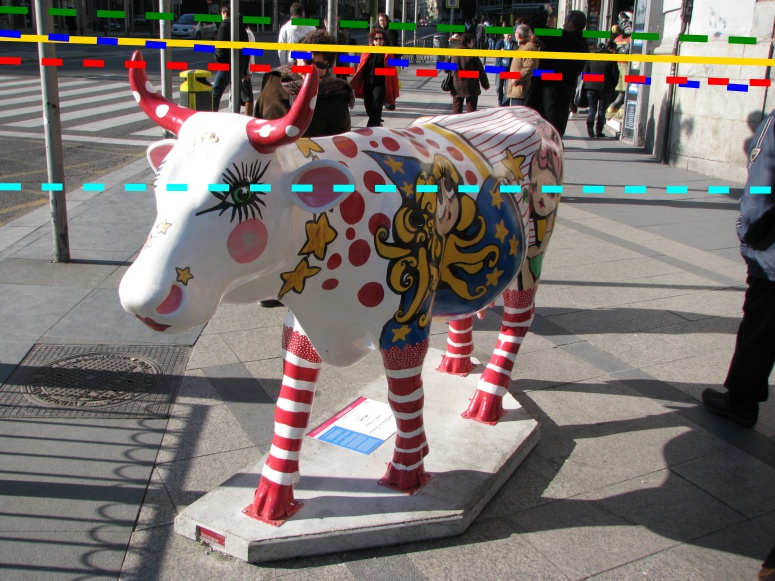} \\
    \end{tabular}
    \caption{Examples of horizon prediction on natural scenes in the HLW~\cite{Workman:2016} test set. Refer to \Fig{horizon_line_predictions} for the horizon line color coding.}
    \label{fig:horizon_line_predictions_hlw}
    \vspace{-3pt}
\end{figure}

Since we utilize semantic information learned from the ImageNet pre-trained ResNet, our network performs well even in natural scenes with few line segments, as some results with the HLW~\cite{Workman:2016} dataset are shown in \Fig{horizon_line_predictions_hlw}. 
For the quantitative analyses with the HLW~\cite{Workman:2016} dataset, refer to the supplementary material.

\begin{table}[t!]
\renewcommand{\arraystretch}{0.9}
\caption{
Quantitative evaluation of the different numbers of layers in the transformer decoder (`Id' indicates the index of decoding layers).
More decoding layers achieve better performance.
}
\label{tbl:test_layer}
{\footnotesize
\begin{tabularx}{\linewidth}{C|CC|CC|CC|CC|C}
\toprule
\multirow{2}{*}{Id} & \multicolumn{2}{c|}{Up Dir. ($^\circ$) $\downarrow$}  & \multicolumn{2}{c|}{Pitch ($^\circ$) $\downarrow$} & \multicolumn{2}{c|}{Roll ($^\circ$) $\downarrow$} & \multicolumn{2}{c|}{FoV ($^\circ$) $\downarrow$}  & \multirow{2}{*}{\makecell{AUC\\($\%$) $\uparrow$}} \\
\cline{2-9}
 & Mean & Med. & Mean & Med. & Mean & Med. & Mean & Med. \\ 
\midrule
1    & 2.34 & 1.99 & 1.95 & 1.57 & 1.02 & 0.78 & 3.83 & 3.00 & 86.46 \\
2    & 2.14 & 1.84 & 1.86 & 1.53 & 0.81 & 0.65 & 3.70 & 2.76 & 87.20 \\
3    & 1.98 & 1.70 & 1.72 & 1.42 & 0.74 & 0.58 & 3.67 & 2.84 & 87.22 \\
4    & 1.85 & 1.59 & 1.61 & 1.35 & 0.69 & 0.55 & 3.68 & 2.78 & 87.21 \\
5    & 1.82 & 1.53 & 1.60 & 1.32 & \textbf{0.66} & \textbf{0.52} & 3.66 & 2.76 & \textbf{87.29}  \\
6    & \textbf{1.80} & \textbf{1.52} & \textbf{1.58} & \textbf{1.31} & \textbf{0.66} & 0.53 & \textbf{3.59} & \textbf{2.72} & \textbf{87.29} \\
\bottomrule
\end{tabularx}
}
\end{table}

\section{Conclusion}

We presented CTRL-C, a novel method for single image camera calibration based on transformer encoder-decoder architecture using multi-modal tokens from the input image. The approach achieves the state-of-the-art performances, by capturing long-term dependencies among image tokens and geometric tokens with the self-/cross-attention modules.
For future work, we investigate to improve the generalization capability of the proposed network using the self-supervised based pre-training stage training \cite{Survey:SSVFL:2020:arXiv,Survey:SSL:2020:arXiv}.

\bigbreak
{\footnotesize
\begin{spacing}{1.0}
\parahead{Acknowledgements}
This was was supported by the National Research Foundation of Korea (NRF) grant funded by the Korea government (MSIT) (2017R1D1A1B03034907, 2021R1F1A1045604, 2020R1C1C1014863) and Institute of Information \& communications Technology Planning \& Evaluation (IITP) grant funded by the Korea government (MSIT) (No.2020-0-01826, Problem-Based Learning Program for Researchers to Proactively Solve Practical AI Problems (Kookmin University) and No.2019-0-01906, Artificial Intelligence Graduate School Program (POSTECH)).
\end{spacing}
}

{\small
\bibliographystyle{ieee_fullname}
\bibliography{references}

\begin{thebibliography}{10}\itemsep=-1pt

\bibitem{GSVI}
{Google Street View Images API}.
\newblock \url{https://developers.google.com/maps/}.

\bibitem{Barinova:2010}
Olga Barinova, Victor Lempitsky, Elena Tretiak, and Pushmeet Kohli.
\newblock {Geometric Image Parsing in Man-Made Environments}.
\newblock In {\em Proc.\ ECCV}, pages 57--70, 2010.

\bibitem{Bhattacharya:2010:MM}
Subhabrata Bhattacharya, Rahul Sukthankar, and Mubarak Shah.
\newblock {A Framework for Photo-Quality Assessment and Enhancement based on
  Visual Aesthetics}.
\newblock In {\em Proc.\ ACM Multimedia}, pages 271--280, 2010.

\bibitem{DETR:2020}
Nicolas Carion, Francisco Massa, Gabriel Synnaeve, Nicolas Usunier, Alexander
  Kirillov, and Sergey Zagoruyko.
\newblock {End-to-End Object Detection with Transformers}.
\newblock In {\em Proc.\ ECCV}, pages 213--229, 2020.

\bibitem{Chaudhury:2014:ICIP}
Krishnendu Chaudhury, Stephen DiVerdi, and Sergey Ioffe.
\newblock {Auto-rectification of user photos}.
\newblock In {\em Proc.\ ICIP}, pages 3479--3483, 2014.

\bibitem{UNITER:2020:ECCV}
Yen-Chun Chen, Linjie Li, Licheng Yu, Ahmed~El Kholy, Faisal Ahmed, Zhe Gan, Yu
  Cheng, and Jingjing Liu.
\newblock {UNITER: UNiversal Image-TExt Representation Learning}.
\newblock In {\em Proc.\ ECCV}, pages 104--120, 2020.

\bibitem{Coughlan:2000}
James~M. Coughlan and Alan~L. Yuille.
\newblock {The Manhattan World Assumption: Regularities in scene statistics
  which enable Bayesian inference}.
\newblock In {\em Proc.\ NIPS}, pages 845--851, 2000.

\bibitem{Criminisi:2000}
Antonio Criminisi, Ian Reid, and Andrew Zisserman.
\newblock {Single View Metrology}.
\newblock {\em International Journal of Computer Vision}, 40(2):123--148, 2000.

\bibitem{BERT:2019}
Jacob Devlin, Ming-Wei Chang, Kenton Lee, and Kristina Toutanova.
\newblock {BERT: Pre-training of Deep Bidirectional Transformers for Language
  Understanding}.
\newblock In {\em Proc.\ NAACL-HLT}, pages 4171--4186, 2019.

\bibitem{ViT:2021}
Alexey Dosovitskiy, Lucas Beyer, Alexander Kolesnikov, Dirk Weissenborn,
  Xiaohua Zhai, Thomas Unterthiner, Mostafa Dehghani, Matthias Minderer, Georg
  Heigold, Sylvain Gelly, Jakob Uszkoreit, and Neil Houlsby.
\newblock {An Image is Worth 16x16 Words: Transformers for Image Recognition at
  Scale}.
\newblock In {\em Proc.\ ICLR}, 2021.

\bibitem{Fischer:2015}
Philipp Fischer, Alexey Dosovitskiy, and Thomas Brox.
\newblock {Image Orientation Estimation with Convolutional Networks}.
\newblock In {\em Proc.\ GCPR}, pages 368--378, 2015.

\bibitem{Survey:Transformer:2020}
Kai Han, Yunhe Wang, Hanting Chen, Xinghao Chen, Jianyuan Guo, Zhenhua Liu,
  Yehui Tang, An Xiao, Chunjing Xu, Yixing Xu, Zhaohui Yang, Yiman Zhang, and
  Dacheng Tao.
\newblock {A Survey on Visual Transformer}.
\newblock {\em arXiv}, 2020.

\bibitem{MVG}
Richard Hartley and Andrew Zisserman.
\newblock {\em {Multiple View Geometry in Computer Vision}}.
\newblock Cambridge University Press, 2nd edition, 2004.

\bibitem{He:2016}
Kaiming He, Xiangyu Zhang, Shaoqing Ren, and Jian Sun.
\newblock {Deep Residual Learning for Image Recognition}.
\newblock In {\em Proc.\ CVPR}, pages 770--778, 2016.

\bibitem{Hoiem:2008:IJCV}
Derek Hoiem, Alexei~A. Efros, and Martial Hebert.
\newblock {Putting Objects in Perspective}.
\newblock {\em International Journal of Computer Vision}, 80:3--15, 2008.

\bibitem{Hold-Geoffroy:2018}
Yannick Hold-Geoffroy, Kalyan Sunkavalli, Jonathan Eisenmann, Matthew Fisher,
  Emiliano Gambaretto, Sunil Hadap, and Jean-François Lalonde.
\newblock {A Perceptual Measure for Deep Single Image Camera Calibration}.
\newblock In {\em Proc.\ CVPR}, pages 2354--2363, 2018.

\bibitem{Survey:SSVFL:2020:arXiv}
Longlong Jing and Yingli Tian.
\newblock {Self-supervised Visual Feature Learning with Deep Neural Networks: A
  Survey}.
\newblock {\em arXiv}, 2020.

\bibitem{Karsch:2011:SIGASIA}
Kevin Karsch, Varsha Hedau, David Forsyth, and Derek Hoiem.
\newblock {Rendering Synthetic Objects into Legacy Photographs}.
\newblock {\em ACM Trans.\ Graphics (Proc.\ SIGGRAPH Asia 2011)}, 30(6):Article
  157, 2011.

\bibitem{Karsch:2014:TOG}
Kevin Karsch, Kalyan Sunkavalli, Sunil Hadap, Nathan Carr, Hailin Jin, Rafael
  Fonte, Michael Sittig, and David Forsyth.
\newblock {Automatic Scene Inference for 3D Object Compositing}.
\newblock {\em ACM Trans.\ Graphics}, 33(3):Article 32, 2014.

\bibitem{Survey:Transformer:2021}
Salman Khan, Muzammal Naseer, Munawar Hayat, Syed~Waqas Zamir, Fahad~Shahbaz
  Khan, and Mubarak Shah.
\newblock {Transformers in Vision: A Survey}.
\newblock {\em arXiv}, 2021.

\bibitem{Lee:2014}
Hyunjoon Lee, Eli Shechtman, Jue Wang, and Seungyong Lee.
\newblock {Automatic Upright Adjustment of Photographs with Robust Camera
  Calibration}.
\newblock {\em IEEE Trans.\ Pattern Analysis Machine Intelligence},
  36(5):833--844, 2014.

\bibitem{Lee:2020:ECCV}
Jinwoo Lee, Minhyuk Sung, Hyunjoon Lee, and Junho Kim.
\newblock {Neural Geometric Parser for Single Image Camera Calibration}.
\newblock In {\em Proc.\ ECCV}, pages 541--557, 2020.

\bibitem{VisualBERT:2019}
Liunian~Harold Li, Mark Yatskar, Da Yin, Cho-Jui Hsieh, and Kai-Wei Chang.
\newblock {VisualBERT: A Simple and Performant Baseline for Vision and
  Language}.
\newblock {\em arXiv}, 2019.

\bibitem{Survey:SSL:2020:arXiv}
Xiao Liu, Fanjin Zhang, Zhenyu Hou, Li Mian, Zhaoyu Wang, Jing Zhang, and Jie
  Tang.
\newblock {Self-supervised Learning: Generative or Contrastive}.
\newblock {\em arXiv}, 2020.

\bibitem{ViLBERT:2019:NeurIPS}
Jiasen Lu, Dhruv Batra, Devi Parikh, and Stefan Lee.
\newblock {ViLBERT: Pretraining Task-Agnostic Visiolinguistic Representations
  for Vision-and-Language Tasks}.
\newblock In {\em Proc.\ NeurIPS}, 2019.

\bibitem{12-in-1:2020:CVPR}
Jiasen Lu, Vedanuj Goswami, Marcus Rohrbach, Devi Parikh, and Stefan Lee.
\newblock {12-in-1: Multi-Task Vision and Language Representation Learning}.
\newblock In {\em Proc.\ CVPR}, pages 10437--10446, 2020.

\bibitem{Ma:3DV}
Yi Ma, Stefano Soatto, Jana Ko\u{s}eck\'{a}, and S.~Shankar Sastry.
\newblock {\em {An Invitation to 3-D Vision: From Images to Geometric Models}}.
\newblock Springer, 2004.

\bibitem{Samii:2015:CGF}
Armin Samii, Radom\'{i}r M\v{e}ch, and Zhe Lin.
\newblock {Data‐Driven Automatic Cropping Using Semantic Composition Search}.
\newblock {\em Computer Graphics Forum}, 34(1):141--151, 2015.

\bibitem{Schaffalitzky:2000}
Frederik Schaffalitzky and Andrew Zisserman.
\newblock {Planar grouping for automatic detection of vanishing lines and
  points}.
\newblock {\em Image and Vision Computing}, 18(9):647--658, 2000.

\bibitem{Schindler:2004}
Grant Schindler and Frank Dellaert.
\newblock {Atlanta World: An Expectation Maximization Framework for
  Simultaneous Low-level Edge Grouping and Camera Calibration in Complex
  Man-made Environments}.
\newblock In {\em Proc.\ CVPR}, 2004.

\bibitem{Simon:2018}
Gilles Simon, Antoine Fond, and Marie-Odile Berger.
\newblock {A-Contrario Horizon-First Vanishing Point Detection Using
  Second-Order Grouping Laws}.
\newblock In {\em Proc.\ ECCV}, pages 318--333, 2018.

\bibitem{LXMERT:2019:EMNLP}
Hao Tan and Mohit Bansal.
\newblock {LXMERT: Learning Cross-Modality Encoder Representations from
  Transformers}.
\newblock In {\em Proc.\ EMNLP-IJCNLP}, pages 5100--5111, 2019.

\bibitem{Tardif:2009}
Jean-Philippe Tardif.
\newblock {Non-Iterative Approach for Fast and Accurate Vanishing Point
  Detection}.
\newblock In {\em Proc.\ ICCV}, pages 1250--1257, 2009.

\bibitem{ECD:2012}
Elena Tretyak, Olga Barinova, Pushmeet Kohli, and Victor Lempitsky.
\newblock {Geometric Image Parsing in Man-Made Environments}.
\newblock {\em International Journal of Computer Vision}, 97:305--321, 2012.

\bibitem{Transformer:2017}
Ashish Vaswani, Noam Shazeer, Niki Parmar, Jakob Uszkoreit, Llion Jones,
  Aidan~N. Gomez, Łukasz Kaiser, and Illia Polosukhin.
\newblock {Attention is All you Need}.
\newblock In {\em Proc.\ NIPS}, pages 6000--6010, 2017.

\bibitem{Gioi:2010}
Rafael~Grompone von Gioi, J{\'e}r{\'e}mie Jakubowicz, Jean-Michel Morel, and
  Gregory Randall.
\newblock {LSD: A Fast Line Segment Detector with a False Detection Control}.
\newblock {\em IEEE Trans.\ Pattern Analysis Machine Intelligence},
  32(4):722--732, 2010.

\bibitem{Workman:2016}
Scott Workman, Menghua Zhai, and Nathan Jacobs.
\newblock {Horizon Lines in the Wild}.
\newblock In {\em Proc.\ BMVC}, pages 20.1--20.12, 2016.

\bibitem{Xian:2019}
Wenqi Xian, Zhengqi Li, Matthew Fisher, Jonathan Eisenmann, Eli Shechtman, and
  Noah Snavely.
\newblock {UprightNet: Geometry-Aware Camera Orientation Estimation From Single
  Images}.
\newblock In {\em Proc.\ ICCV}, pages 9974--9983, 2019.

\bibitem{SUN360:2012}
Jianxiong Xiao, Krista~A. Ehinger, Aude Oliva, and Antonio Torralba.
\newblock {Recognizing Scene Viewpoint using Panoramic Place Representation}.
\newblock In {\em Proc.\ CVPR}, pages 2695--2702, 2012.

\bibitem{LETR:2021}
Yifan Xu, Weijian Xu, David Cheung, and Zhuowen Tu.
\newblock {Line Segment Detection Using Transformers without Edges}.
\newblock In {\em Proc.\ CVPR}, pages 4257--4266, 2021.

\bibitem{Zhai:2016}
Menghua Zhai, Scott Workman, and Nathan Jacobs.
\newblock {Detecting Vanishing Points using Global Image Context in a
  Non-Manhattan World}.
\newblock In {\em Proc.\ CVPR}, pages 5657--5665, 2016.

\bibitem{HoliCity:2020:arXiv}
Yichao Zhou, Jingwei Huang, Xili Dai, Linjie Luo, Zhili Chen, and Yi Ma.
\newblock {HoliCity: A City-Scale Data Platform for Learning Holistic 3D
  Structures}.
\newblock {\em arXiv}, 2020.

\bibitem{Zhu:2020:ECCV}
Rui Zhu, Xingyi Yang, Yannick Hold-Geoffroy, Federico Perazzi, Jonathan
  Eisenmann, Kalyan Sunkavalli, and Manmohan Chandraker.
\newblock {Single View Metrology in the Wild}.
\newblock In {\em Proc.\ ECCV}, pages 316--333, 2020.

\bibitem{DeformableDETR:2021}
Xizhou Zhu, Weijie Su, Lewei Lu, Bin Li, Xiaogang Wang, and Jifeng Dai.
\newblock {Deformable DETR: Deformable Transformers for End-to-End Object
  Detection}.
\newblock In {\em Proc.\ ICLR}, 2021.

\end{thebibliography}
}

\clearpage

\renewcommand{\thesection}{A}
\setcounter{table}{0}
\renewcommand{\thetable}{A\arabic{table}}
\setcounter{figure}{0}
\renewcommand{\thefigure}{A\arabic{figure}}

\newif\ifpaper
\papertrue

\ifpaper
\section*{Appendix}

\ifpaper
  \newcommand{\refpaper}{}
\else
  \makeatletter
  \newcommand{\manuallabel}[2]{\def\@currentlabel{#2}\label{#1}}
  \makeatother
  \manuallabel{fig:auc}{4}
  \manuallabel{fig:horizon_line_predictions}{5}
  \manuallabel{fig:additional_results}{6}
  \newcommand{\refpaper}{ in the paper}
\fi

\subsection{Generalization Tests}
We compare the generalization capacity of our network with those of three network-based approaches, DeepHorizon~\cite{Workman:2016}, Perceptual~\cite{Hold-Geoffroy:2018} and GPNet~\cite{Lee:2020:ECCV}.
Specifically, we train our CTRL-C and other networks with either the Google Street View (GSV)~\cite{GSVI} or SUN360~\cite{SUN360:2012} datasets and then test them with the other datasets, including HoliCity~\cite{HoliCity:2020:arXiv} dataset, Horizon Lines in the Wild (HLW)~\cite{Workman:2016} dataset, and Eurasian Cities (ECD)~\cite{ECD:2012} dataset.
For the HoliCity~\cite{HoliCity:2020:arXiv} dataset, we evaluate networks trained with the SUN360~\cite{SUN360:2012} dataset for comparison, since the sampling range of FoV in the SUN360~\cite{SUN360:2012} dataset covers $90 ^{\circ}$, which is the FoV used in the entire HoliCity~\cite{HoliCity:2020:arXiv} dataset.
For HLW~\cite{Workman:2016} and ECD~\cite{ECD:2012} datasets, we only measure the horizon line prediction accuracy since these datasets do not provide GT for the other camera parameters.

\parahead{HoliCity~\cite{HoliCity:2020:arXiv} dataset}
Compared with the other methods, our CTRL-C provides better accuracy overall except for pitch. The AUC of the horizon line errors is improved from 81.72\% of the previous SotA to 84.16\% of ours, which has a 2.44\% gap.
\Fig{horizon_line_predictions_holicity} shows experimental results on the HoliCity~\cite{HoliCity:2020:arXiv} test set, as in \Fig{horizon_line_predictions}\refpaper, visualizing qualitative evaluations on horizon line predictions.  
\Fig{additional_results_holicity} shows further examples illustrating the inputs and outputs of our network, as in \Fig{additional_results}\refpaper.

\parahead{HLW~\cite{Workman:2016} dataset}
\Tbl{test_hlw} shows the comparison results evaluated on the HLW~\cite{Workman:2016} test set.
Our CTRL-C provides better overall accuracy and less sensitivity to the choice of the training datasets compared to the other network-based approaches.
For instance, the difference of AUCs with the different training datasets (GSV~\cite{GSVI} and SUN360~\cite{SUN360:2012}) is 5.15\% for our CTRL-C while 8.37\% for GPNet~\cite{Lee:2020:ECCV}.

\parahead{ECD~\cite{ECD:2012} dataset}
\Tbl{test_ecd} shows the comparison results evaluated on the ECD~\cite{ECD:2012} dataset. 
Our CTRL-C shows better generalization performance than those of the other neural approaches, in both cases of training with the GSV~\cite{GSVI} and SUN360~\cite{SUN360:2012} datasets. 

\subsection{Additional Results}

\parahead{Google Street View~\cite{GSVI} dataset}
\Fig{horizon_line_predictions_gsv} shows additional results on the Google Street View~\cite{GSVI} test set, as in \Fig{horizon_line_predictions}\refpaper, visualizing qualitative evaluations on horizon line predictions. \Fig{additional_results_gsv} shows additional results with the Google Street View~\cite{GSVI} test set, as in \Fig{additional_results}\refpaper, illustrating the inputs and outputs of our network.

\smallskip
\parahead{SUN360~\cite{SUN360:2012} dataset}
\Fig{horizon_line_predictions_sun360} shows additional results on the SUN360~\cite{SUN360:2012} test set, as in \Fig{horizon_line_predictions}\refpaper, visualizing qualitative evaluations on horizon line predictions. \Fig{additional_results_sun360} shows additional results with the SUN360~\cite{SUN360:2012} test set, as in \Fig{additional_results}\refpaper, illustrating the inputs and outputs of our network.

\begin{table}[t!]
\renewcommand{\arraystretch}{0.9}
\caption{Quantitative evaluation results on the HoliCity~\cite{HoliCity:2020:arXiv} dataset. All the networks are trained with the SUN360~\cite{SUN360:2012} dataset.}
\label{tbl:test_holicity_1c}
{\footnotesize
\setlength\tabcolsep{1.5pt} 
\begin{tabularx}{\linewidth}{>{\centering}m{0.15cm}|CC|CC|CC|CC|C}
\toprule
\multicolumn{1}{l|}{\multirow{2}{*}{Method}} & \multicolumn{2}{c|}{Up Dir ($^\circ$) $\downarrow$}  & \multicolumn{2}{c|}{Pitch ($^\circ$) $\downarrow$} & \multicolumn{2}{c|}{Roll ($^\circ$) $\downarrow$} & \multicolumn{2}{c|}{FoV ($^\circ$) $\downarrow$}  & \multirow{2}{*}{\makecell{AUC\\($\%$) $\uparrow$}} \\
\cline{2-9}
\multicolumn{1}{l|}{} & Mean & Med. & Mean & Med. & Mean & Med. & Mean & Med. \\ 
\midrule
\multicolumn{1}{l|}{DeepHorizon~\cite{Workman:2016}}       & 7.82 & 3.99 & 6.10 & 2.73 & 3.97 & 2.67 &   -   &   -   & 70.13 \\
\multicolumn{1}{l|}{Perceptual~\cite{Hold-Geoffroy:2018}}  & 7.37 & 3.29 & 6.32 & 2.86 & 3.10 & 1.82 &  5.48 &  2.80 & 70.80 \\
\multicolumn{1}{l|}{GPNet~\cite{Lee:2020:ECCV}}            & 4.17 & \textbf{1.73} & \textbf{1.46} & \textbf{0.74} & 3.65 & 1.36 & 10.03 &  4.29 & 81.72 \\
\midrule
\multicolumn{1}{l|}{\textbf{CTRL-C}}                  & \textbf{2.90} & 1.99 & 2.43 & 1.50 & \textbf{1.36} & \textbf{0.95} &  \textbf{2.68} &  \textbf{1.56} & \textbf{84.16} \\
\bottomrule
\end{tabularx}
}
\end{table}

\begin{table}[t!]
\renewcommand{\arraystretch}{0.9}
\caption{Horizon line prediction results on the HLW~\cite{Workman:2016} dataset.}
\label{tbl:test_hlw}
{\footnotesize
\begin{tabularx}{\linewidth}{m{0.4\linewidth}|C|C}
\toprule
Method & \multicolumn{2}{c}{AUC (\%) $\uparrow$} \\
\hline
Training Sets & GSV & SUN360 \\
\midrule
DeepHorizon \cite{Workman:2016}         & 45.63 & 40.63 \\
Perceptual \cite{Hold-Geoffroy:2018}    & 38.29 & 46.70 \\
GPNet \cite{Lee:2020:ECCV}              & \textbf{48.90} & 40.53 \\
\midrule
CTRL-C                                  & 44.93 & \textbf{50.08} \\
\bottomrule
\end{tabularx}
}
\end{table}

\begin{table}[t!]
\renewcommand{\arraystretch}{0.9}
\caption{Horizon line prediction results on the ECD dataset~\cite{ECD:2012}.}
\label{tbl:test_ecd}
{\footnotesize
\begin{tabularx}{\linewidth}{m{0.4\linewidth}|C|C}
\toprule
Method & \multicolumn{2}{c}{AUC (\%) $\uparrow$} \\
\hline
Training Sets & GSV & SUN360 \\
\midrule
DeepHorizon \cite{Workman:2016}      & 74.26 & 74.60 \\
Perceptual \cite{Hold-Geoffroy:2018} & 67.97 & 76.53 \\
GPNet \cite{Lee:2020:ECCV}           & 77.61 & 75.04 \\
\midrule
CTRL-C                               & \textbf{77.66} & \textbf{79.83} \\
\bottomrule
\end{tabularx}
}
\end{table}

\begin{figure*}[h!]
    \centering
    \setlength\tabcolsep{1.5pt} 
    \begin{tabular}{cccccc}
    \includegraphics[width=0.15\linewidth]{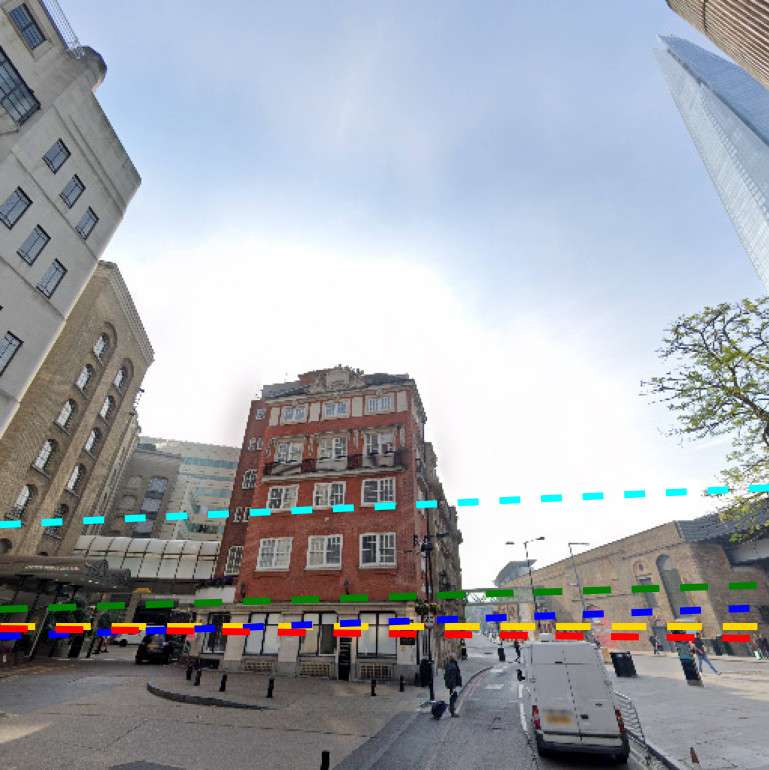} &
    \includegraphics[width=0.15\linewidth]{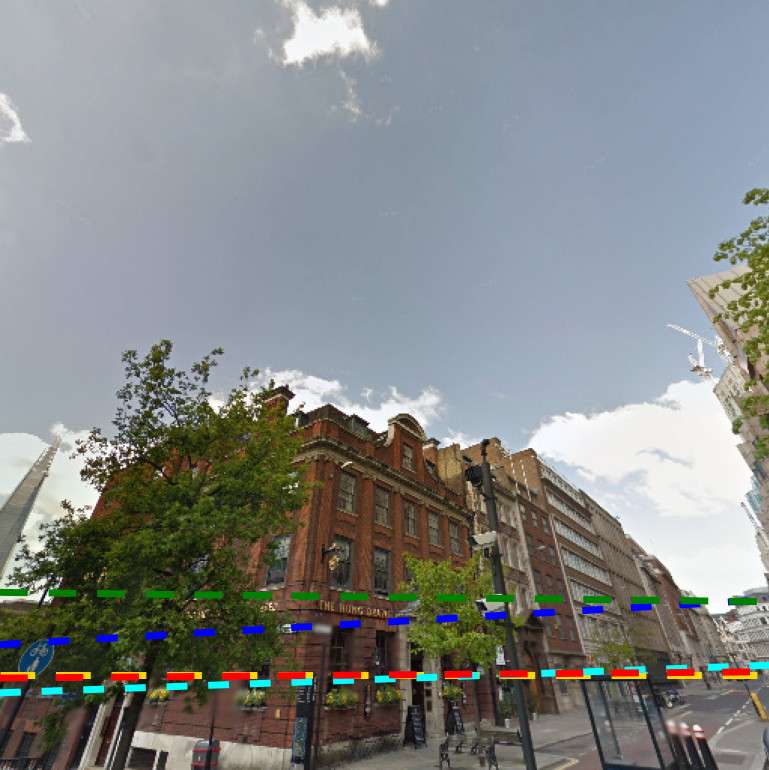} &
    \includegraphics[width=0.15\linewidth]{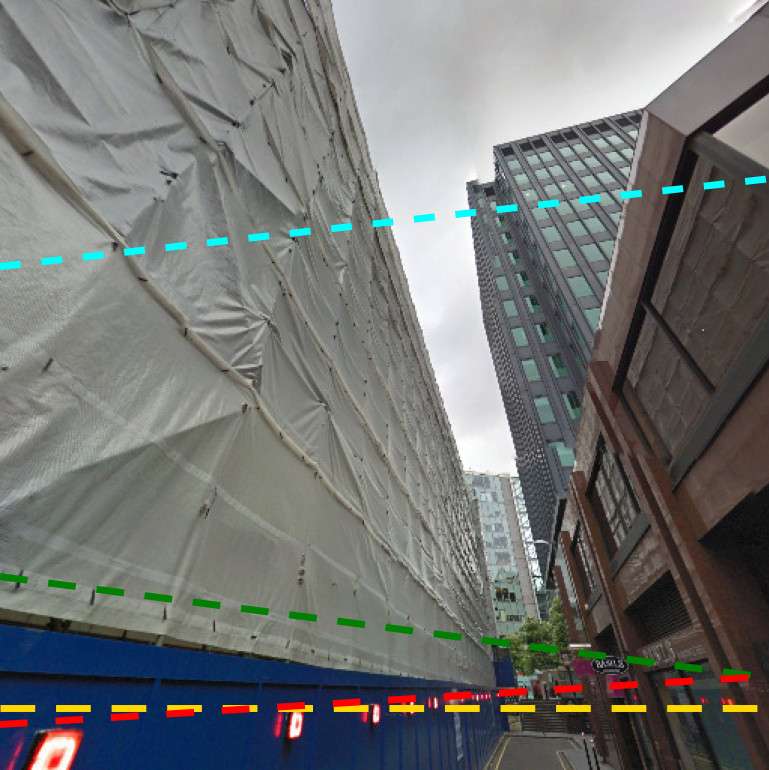} &
    \includegraphics[width=0.15\linewidth]{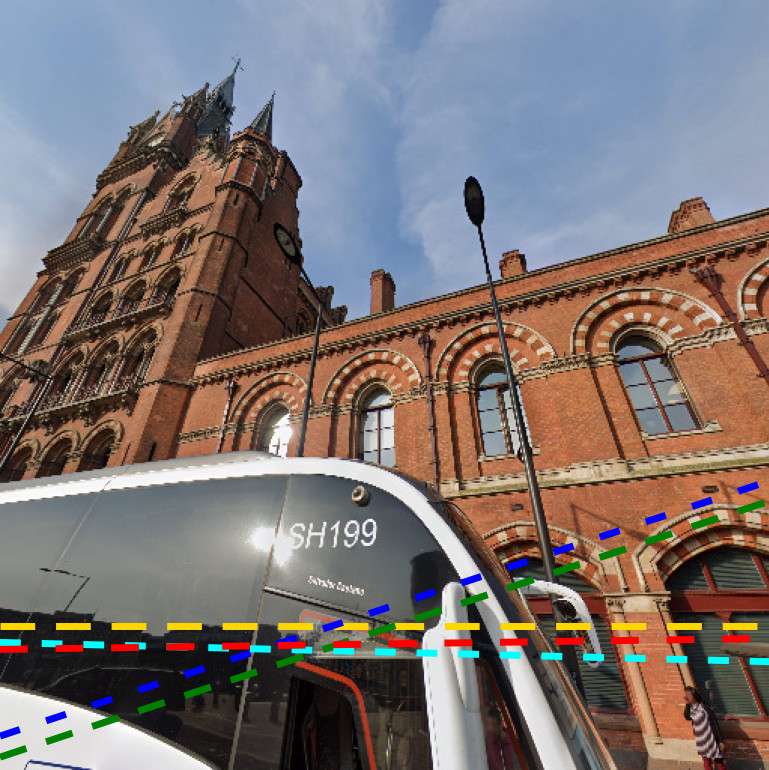} &
    \includegraphics[width=0.15\linewidth]{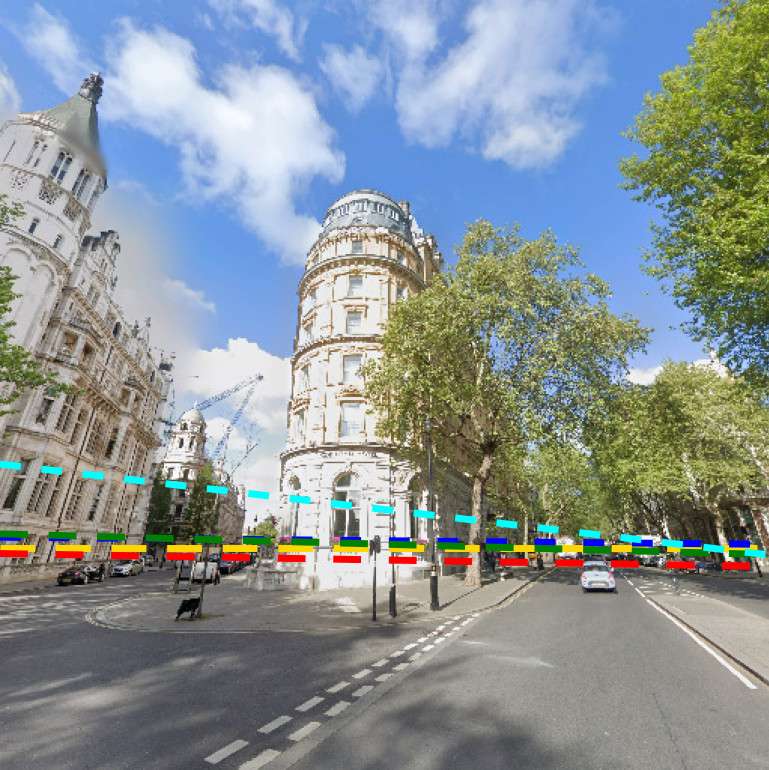} &
    \includegraphics[width=0.15\linewidth]{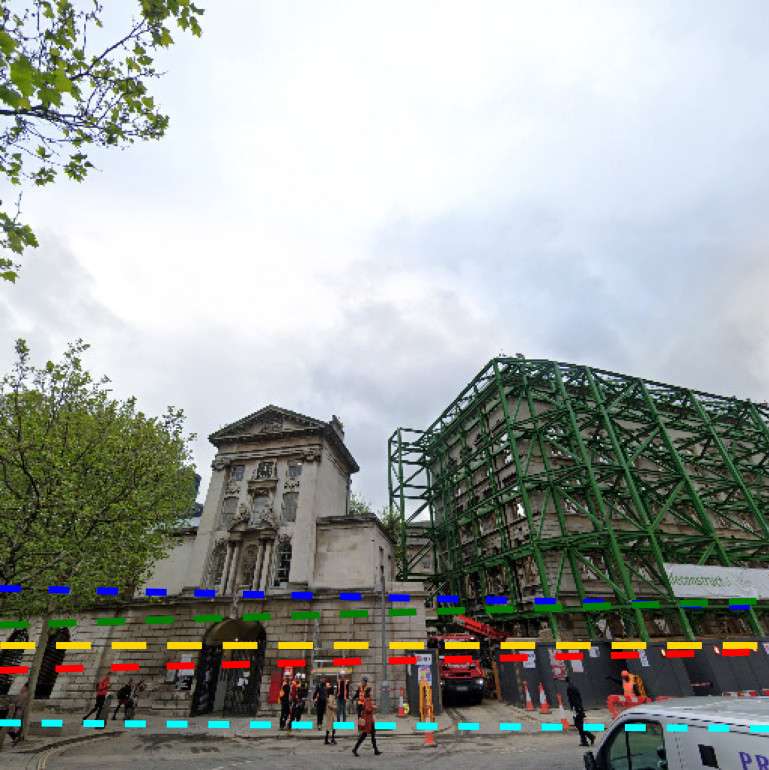} \\
    \includegraphics[width=0.15\linewidth]{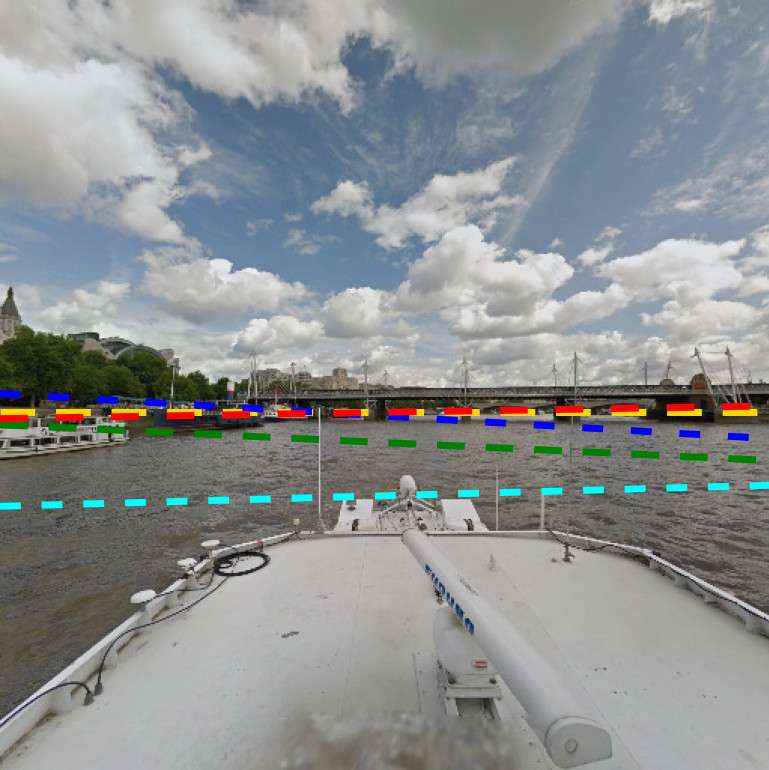} &
    \includegraphics[width=0.15\linewidth]{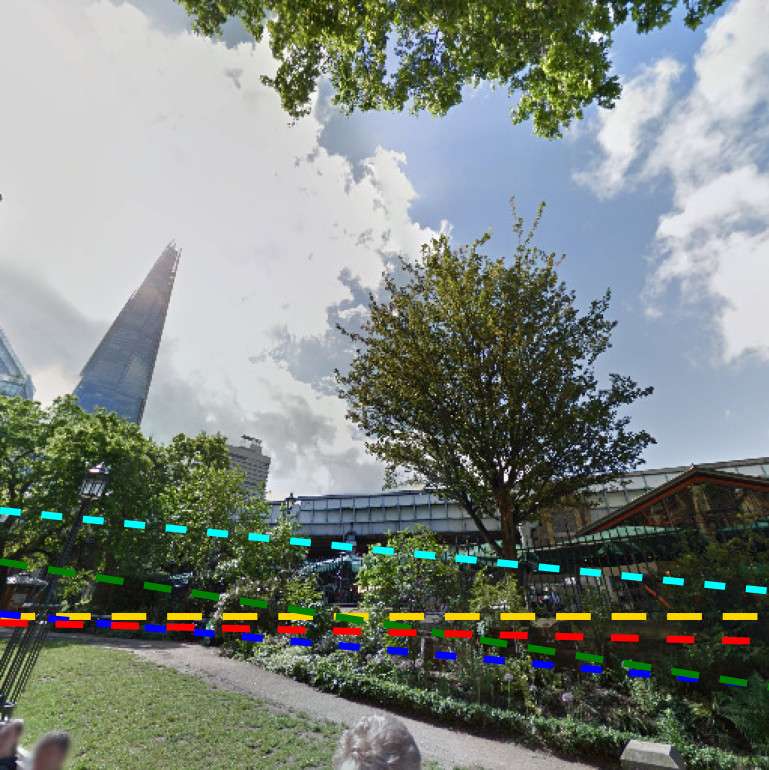} &
    \includegraphics[width=0.15\linewidth]{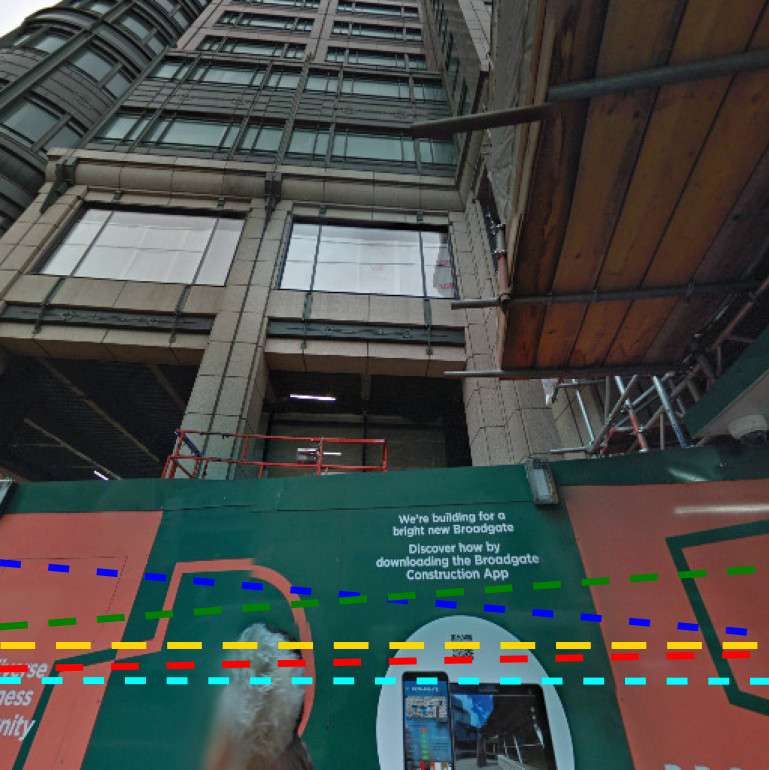} &
    \includegraphics[width=0.15\linewidth]{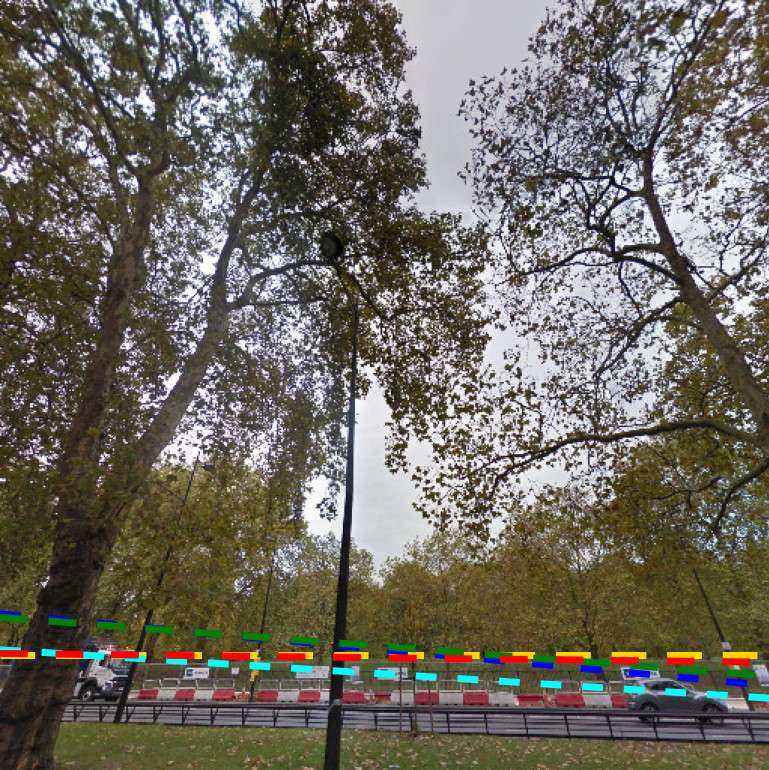} &
    \includegraphics[width=0.15\linewidth]{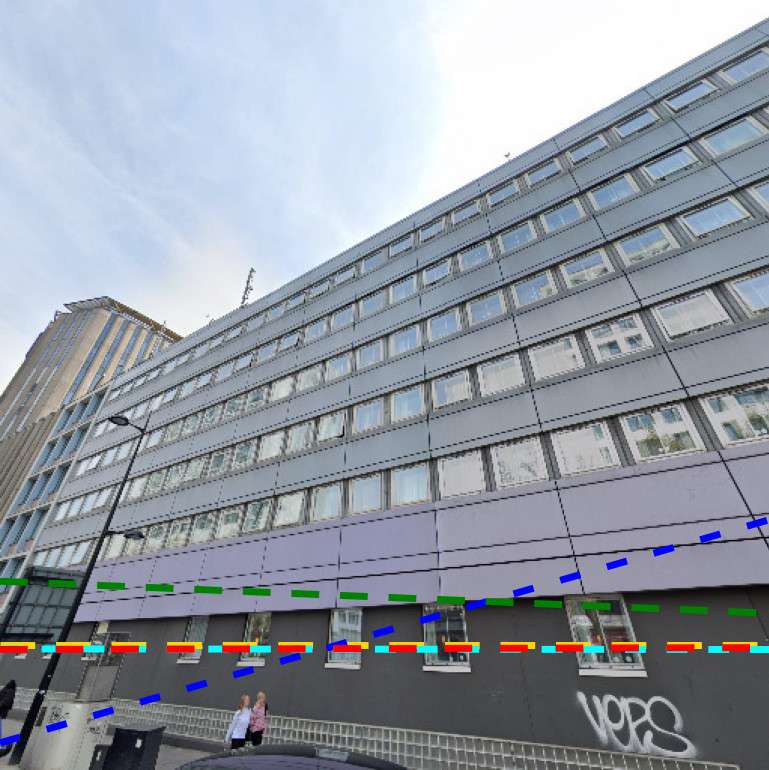} &
    \includegraphics[width=0.15\linewidth]{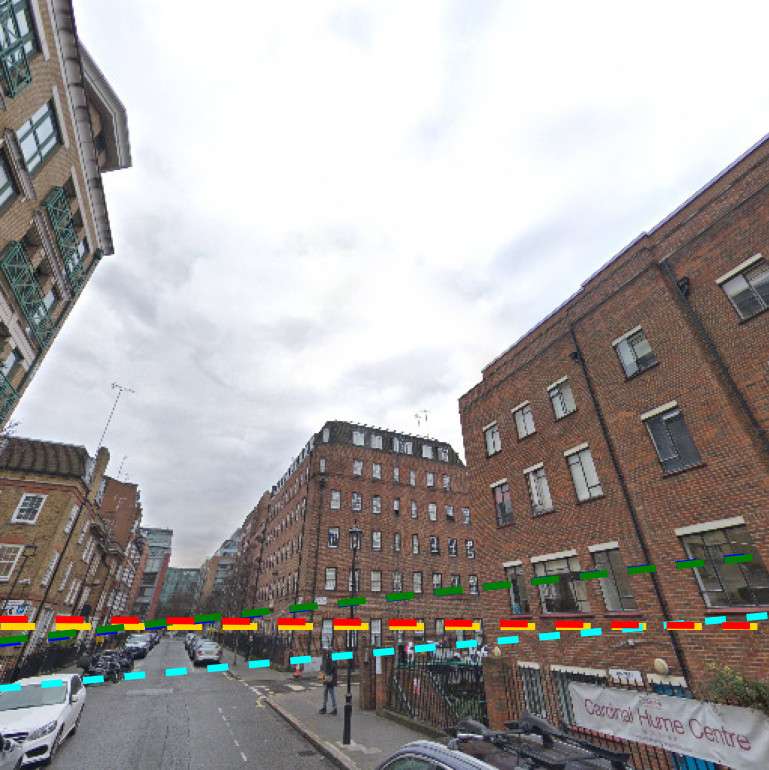} \\
    \includegraphics[width=0.15\linewidth]{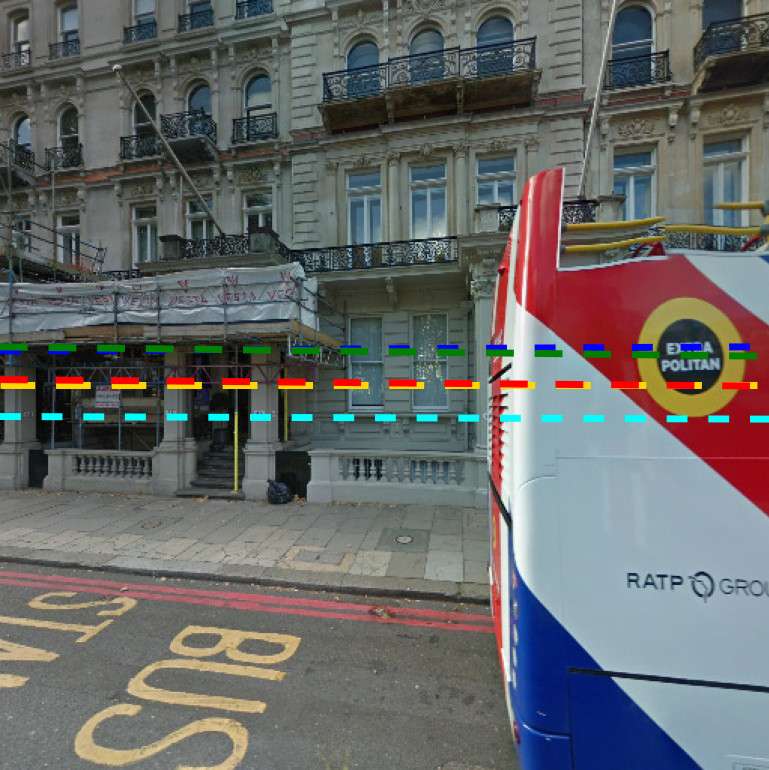} &
    \includegraphics[width=0.15\linewidth]{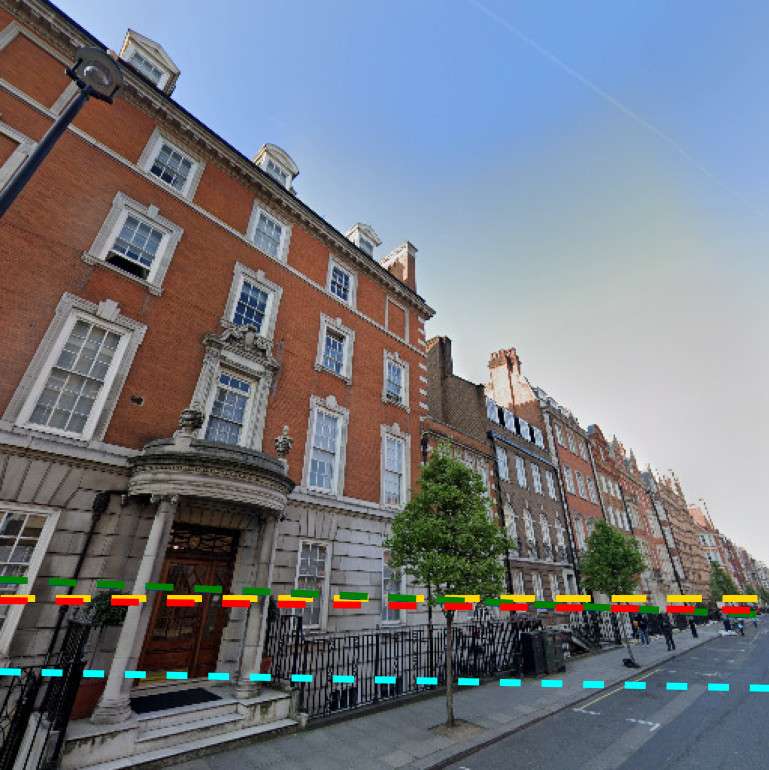} &
    \includegraphics[width=0.15\linewidth]{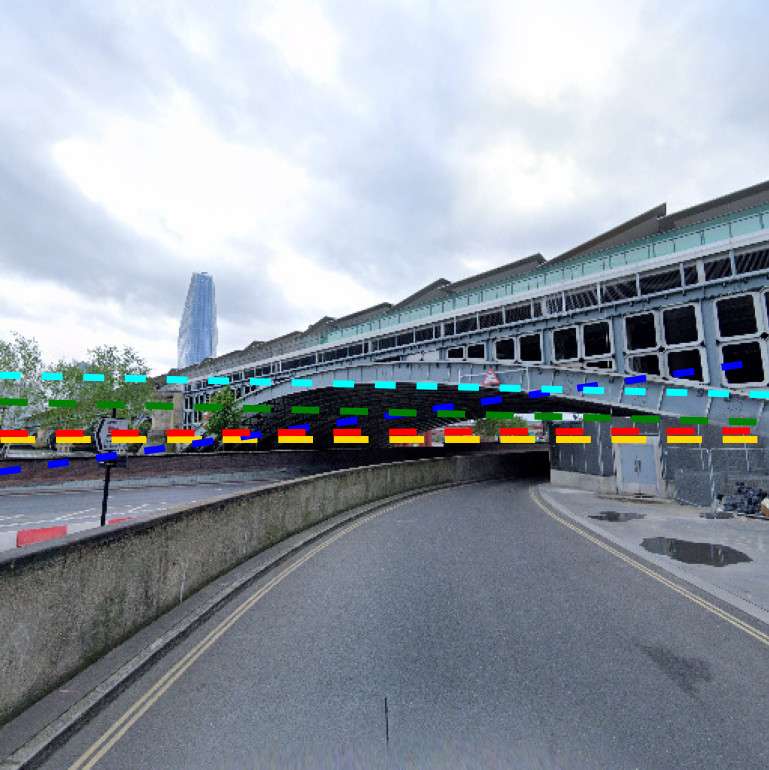} &
    \includegraphics[width=0.15\linewidth]{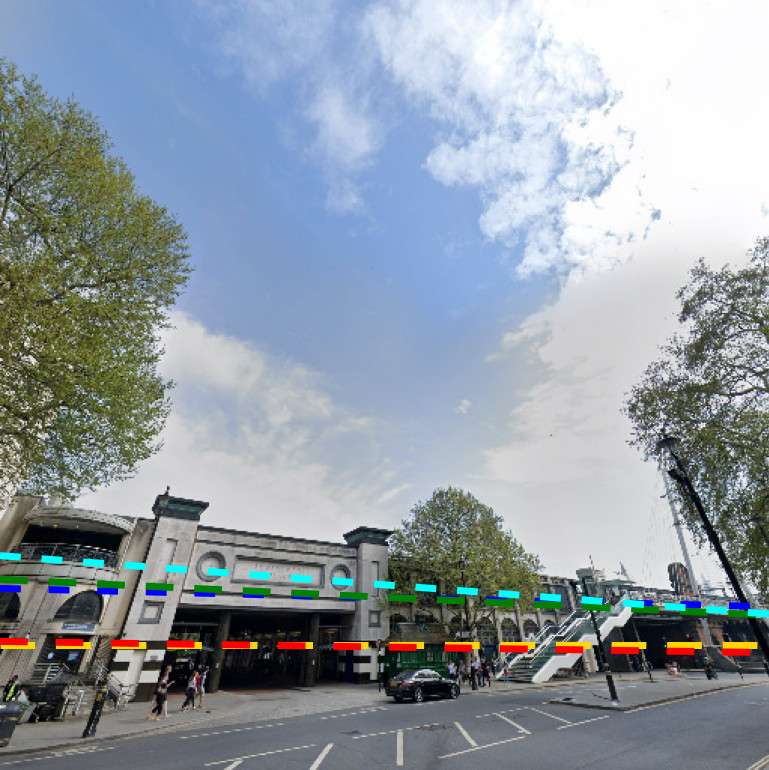} &
    \includegraphics[width=0.15\linewidth]{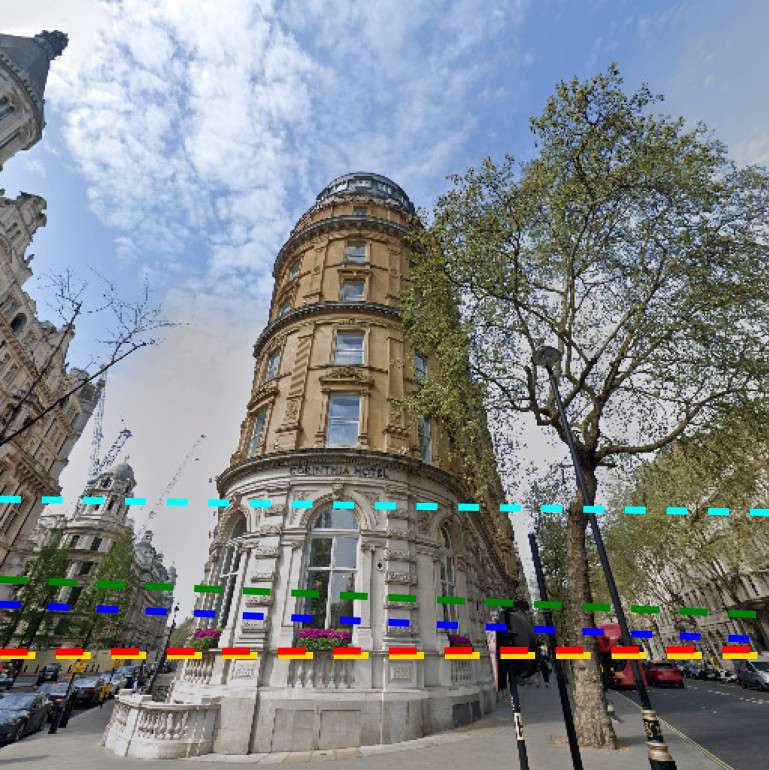} &
    \includegraphics[width=0.15\linewidth]{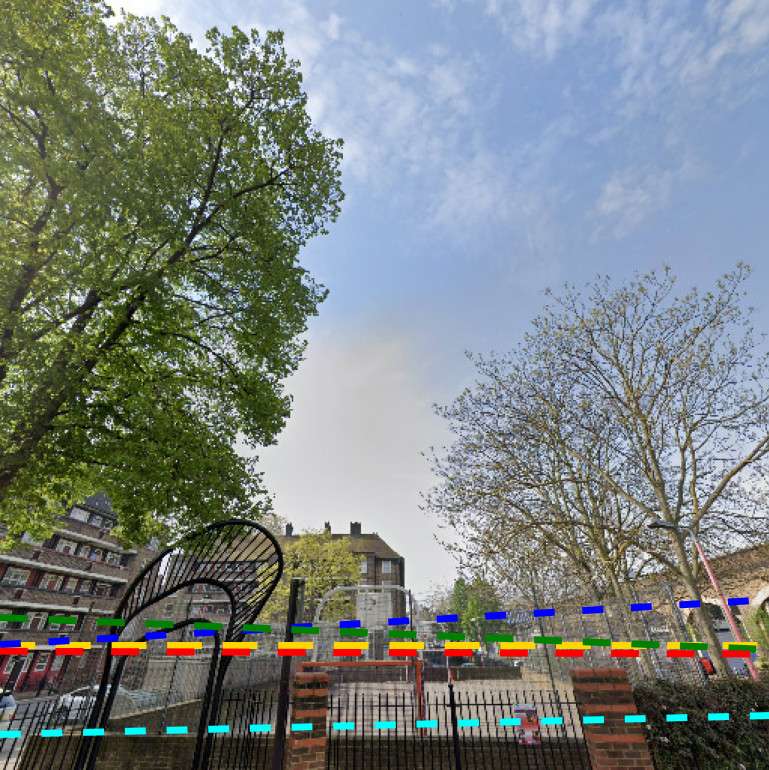} \\
    \end{tabular}
    \newcommand{\crule}[3][red]{\textcolor{#1}{\rule{#2}{#3} \rule{#2}{#3} \rule{#2}{#3} \rule{#2}{#3}}}
    {\scriptsize
    \begin{tabular}{lllll}
    \crule[Goldenrod]{0.01\linewidth}{0.01\linewidth} Ground Truth & 
    \crule[blue]{0.01\linewidth}{0.01\linewidth} DeepHorizon \cite{Workman:2016} & 
    \crule[OliveGreen]{0.01\linewidth}{0.01\linewidth} Perceptual \cite{Hold-Geoffroy:2018} & 
    \crule[SkyBlue]{0.01\linewidth}{0.01\linewidth} GPNet \cite{Lee:2020:ECCV} &
    \crule[red]{0.01\linewidth}{0.01\linewidth} CTRL-C (Ours) \\
    \end{tabular}
    }
    \caption{Examples of horizon line prediction on the HoliCity~\cite{HoliCity:2020:arXiv} test set.}
    \label{fig:horizon_line_predictions_holicity}
\end{figure*}

\begin{figure*}[h!]
    \centering
    \setlength\tabcolsep{1.5pt} 
    \begin{tabular}{ccccccc}
    (a) & 
    \includegraphics[width=0.15\linewidth]{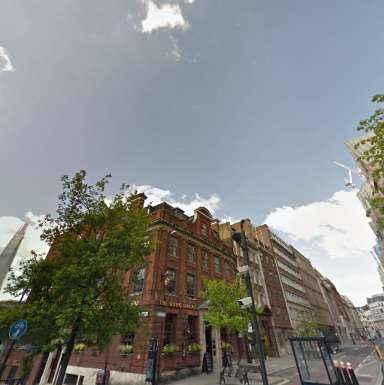} &
    \includegraphics[width=0.15\linewidth]{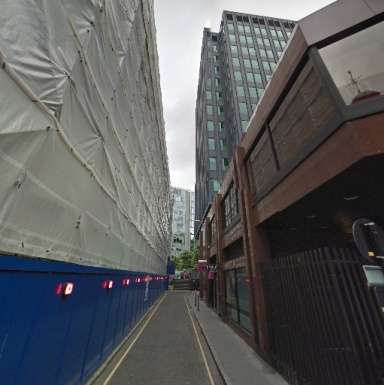} &
    \includegraphics[width=0.15\linewidth]{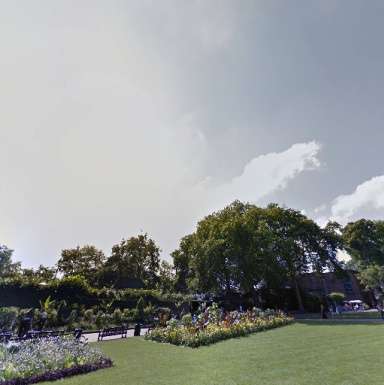} &
    \includegraphics[width=0.15\linewidth]{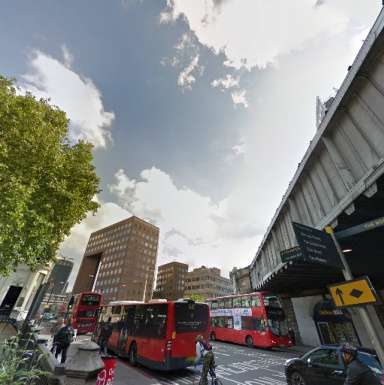} &
    \includegraphics[width=0.15\linewidth]{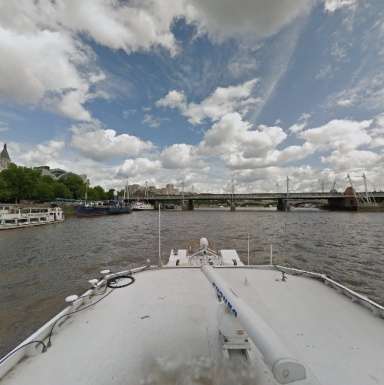} &
    \includegraphics[width=0.15\linewidth]{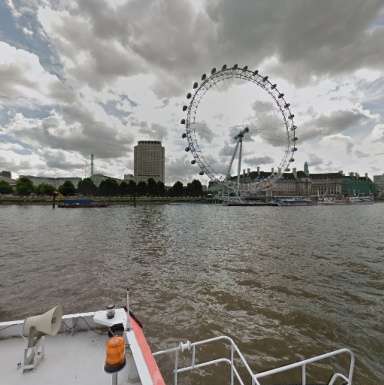} \\
    (b) & 
    \includegraphics[width=0.15\linewidth]{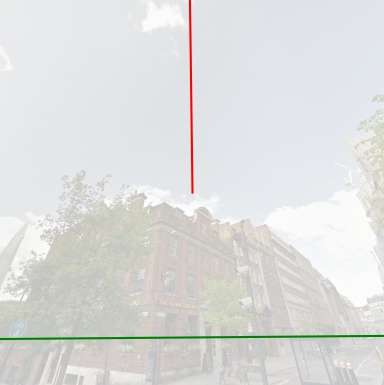} &
    \includegraphics[width=0.15\linewidth]{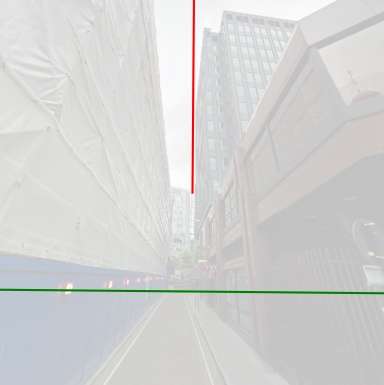} &
    \includegraphics[width=0.15\linewidth]{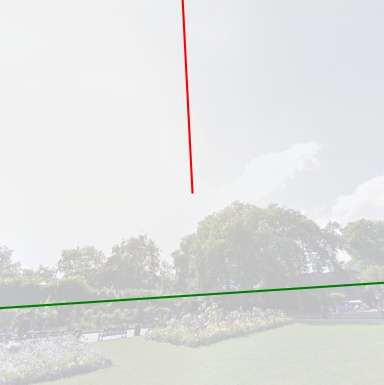} &
    \includegraphics[width=0.15\linewidth]{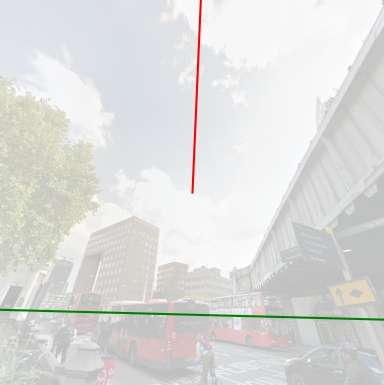} &
    \includegraphics[width=0.15\linewidth]{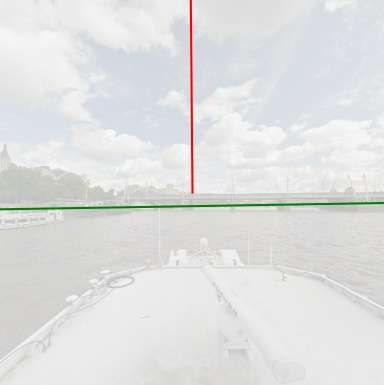} &
    \includegraphics[width=0.15\linewidth]{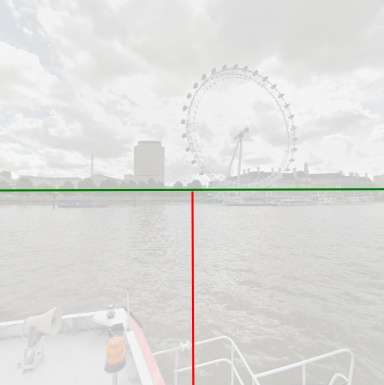} \\
    (c) & 
    \includegraphics[width=0.15\linewidth]{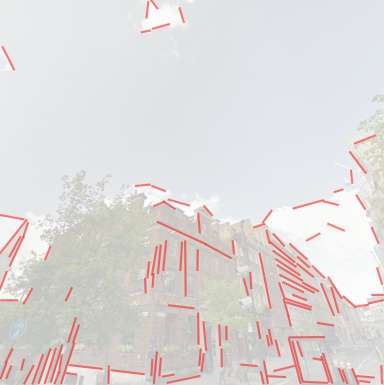} &
    \includegraphics[width=0.15\linewidth]{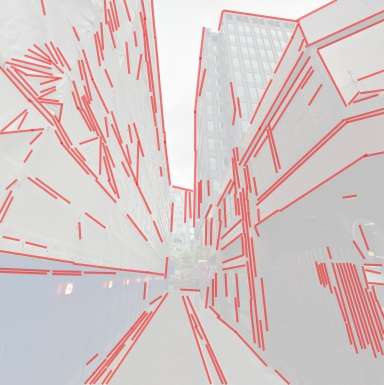} &
    \includegraphics[width=0.15\linewidth]{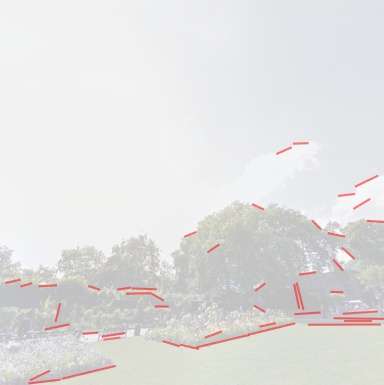} &
    \includegraphics[width=0.15\linewidth]{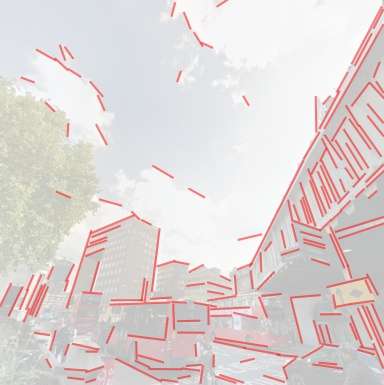} &
    \includegraphics[width=0.15\linewidth]{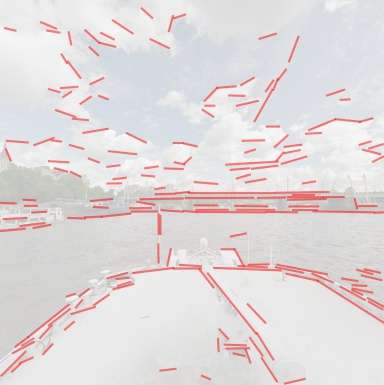} &
    \includegraphics[width=0.15\linewidth]{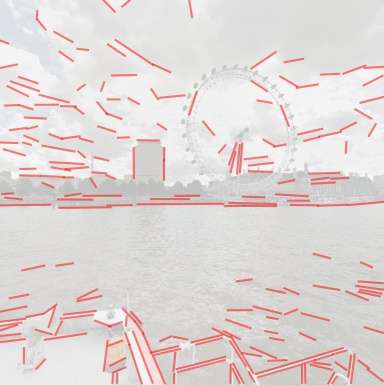} \\
    (d) &
    \includegraphics[width=0.15\linewidth]{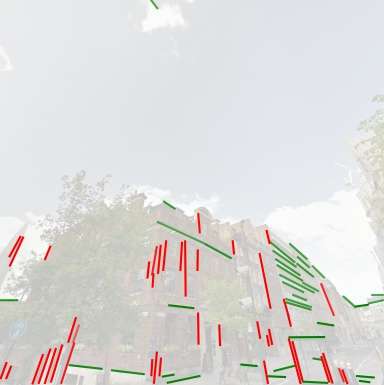} &
    \includegraphics[width=0.15\linewidth]{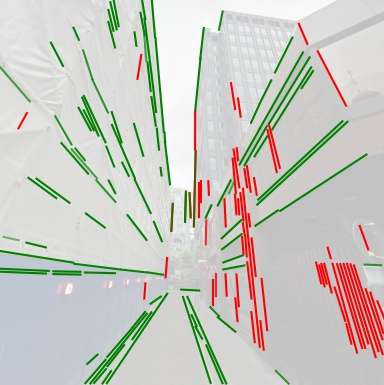} &
    \includegraphics[width=0.15\linewidth]{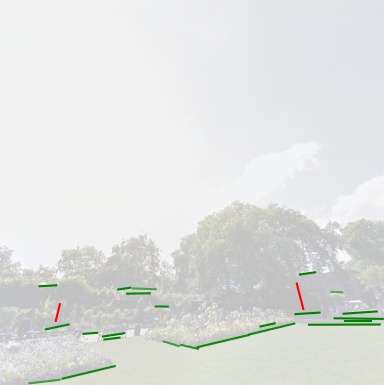} &
    \includegraphics[width=0.15\linewidth]{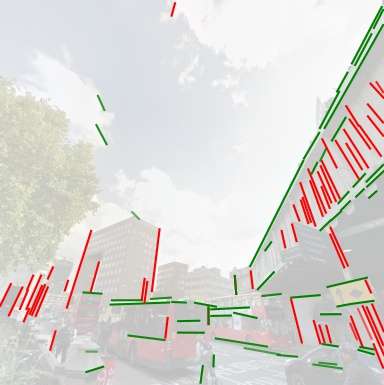} &
    \includegraphics[width=0.15\linewidth]{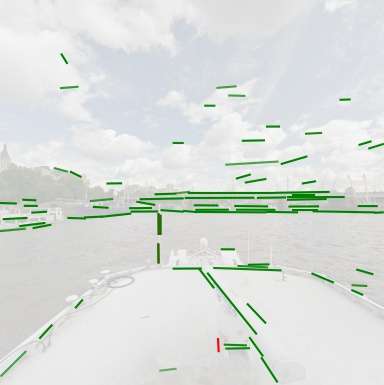} &
    \includegraphics[width=0.15\linewidth]{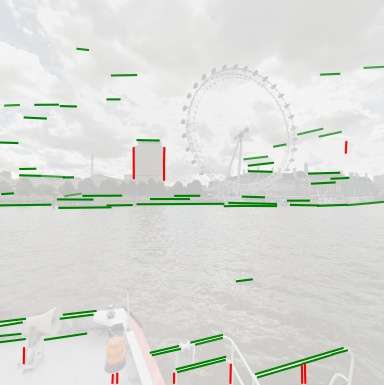} \\
    \end{tabular}
    \caption{More results with HoliCity~\cite{HoliCity:2020:arXiv} test set: (a) input image, (b) estimated horizon line (green) and vertical direction along with the zenith VP (red), 
    (c) detected lines with LSD \cite{Gioi:2010}, 
    (d) estimated vertical (red) and horizontal (green) convergence line segments of (c).}
    \label{fig:additional_results_holicity}
\end{figure*}

\clearpage
\begin{figure*}[t!]
    \centering
    \setlength\tabcolsep{1.5pt} 
    \begin{tabular}{cccccc}
    \includegraphics[width=0.15\linewidth]{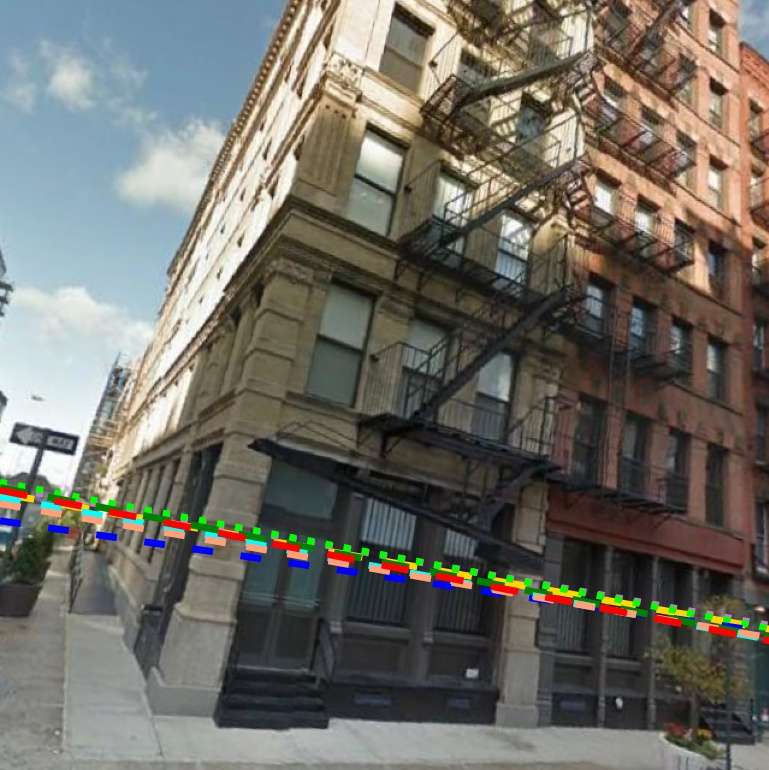} &
    \includegraphics[width=0.15\linewidth]{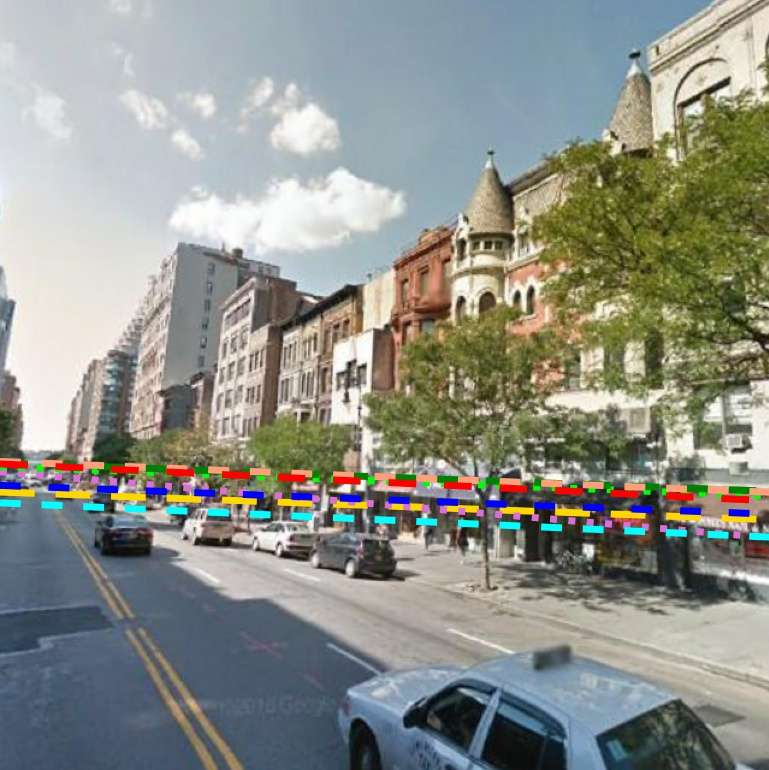} &
    \includegraphics[width=0.15\linewidth]{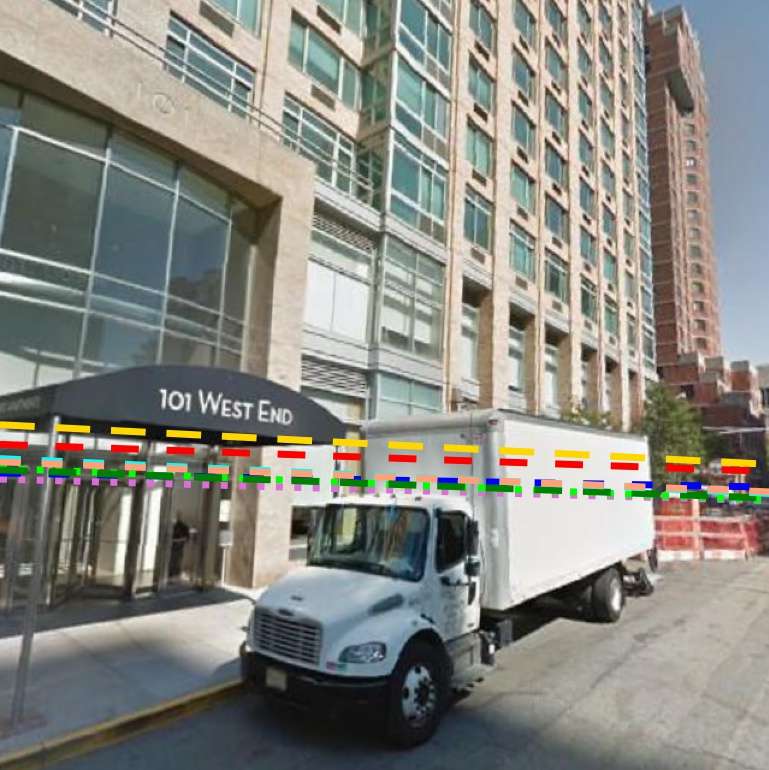} &
    \includegraphics[width=0.15\linewidth]{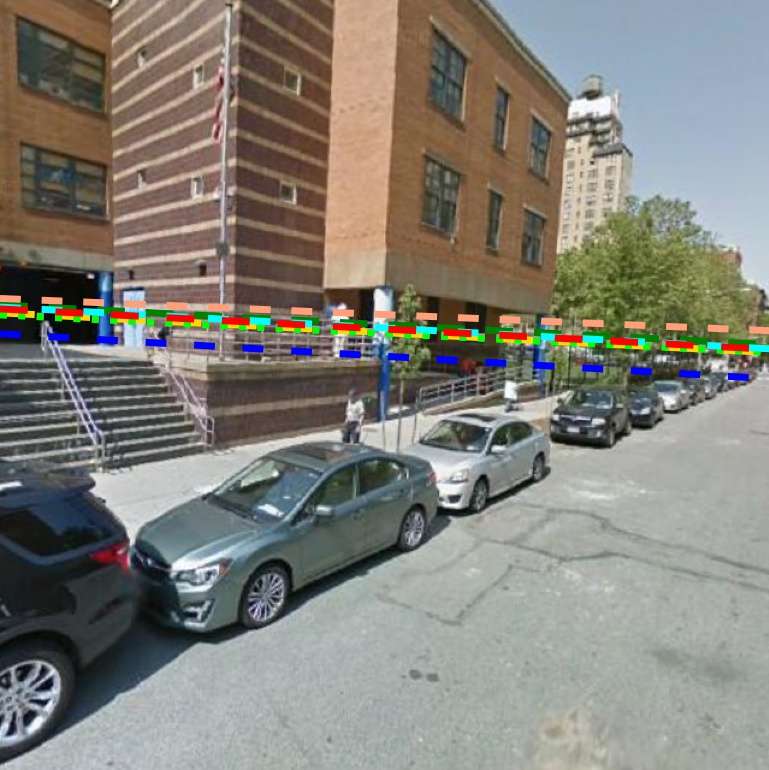} &
    \includegraphics[width=0.15\linewidth]{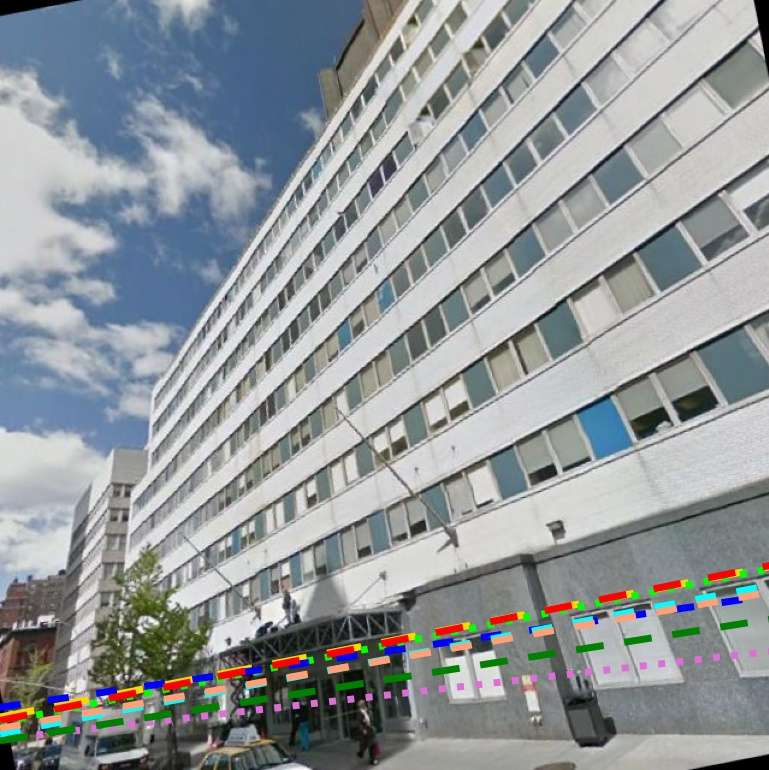} &
    \includegraphics[width=0.15\linewidth]{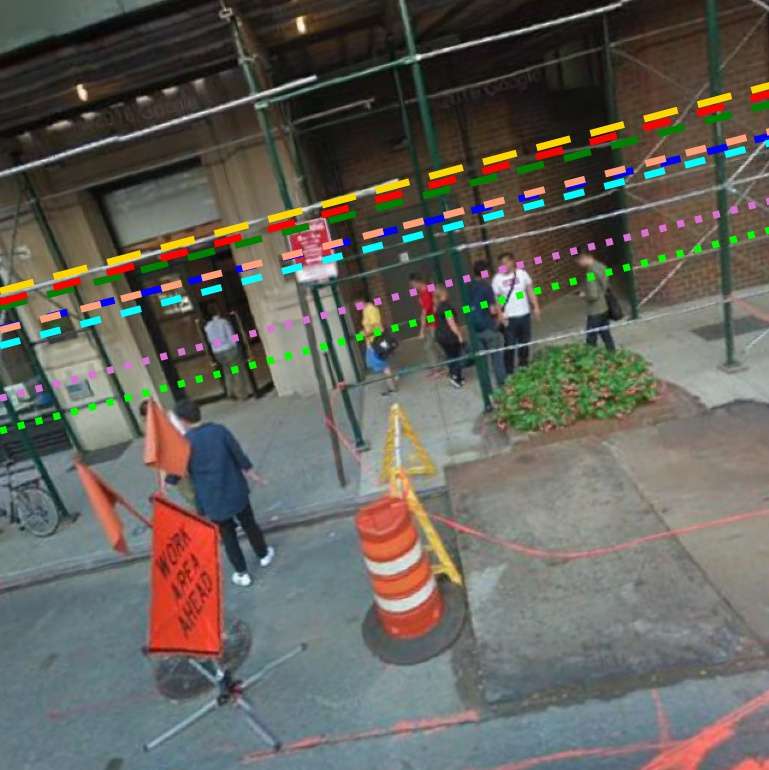} \\
    \includegraphics[width=0.15\linewidth]{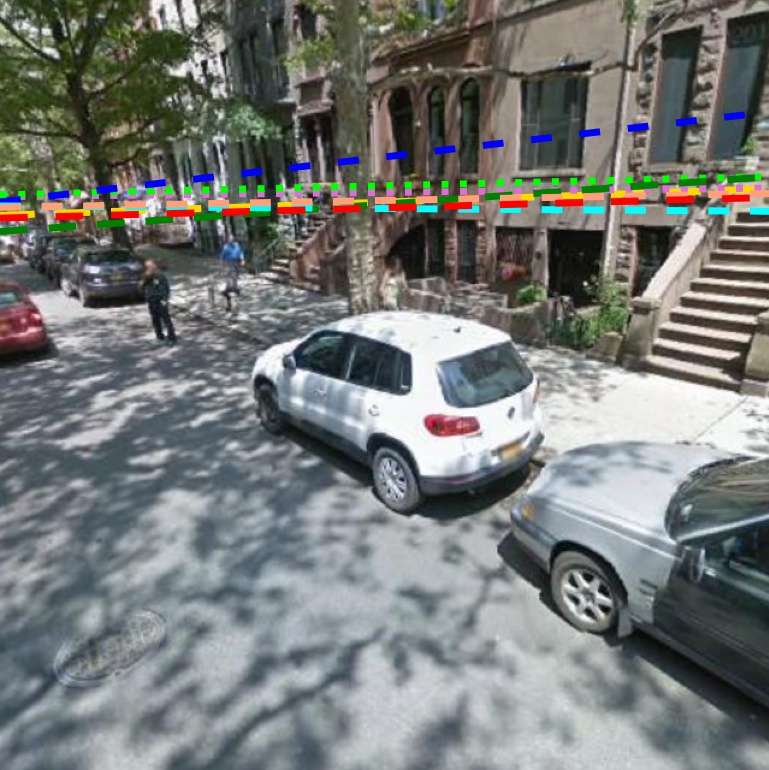} &
    \includegraphics[width=0.15\linewidth]{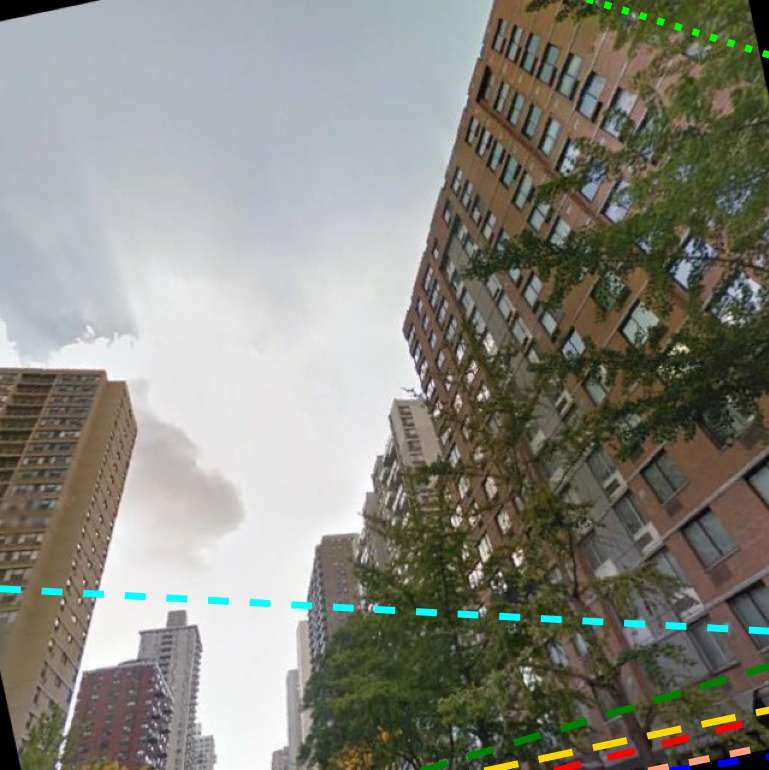} &
    \includegraphics[width=0.15\linewidth]{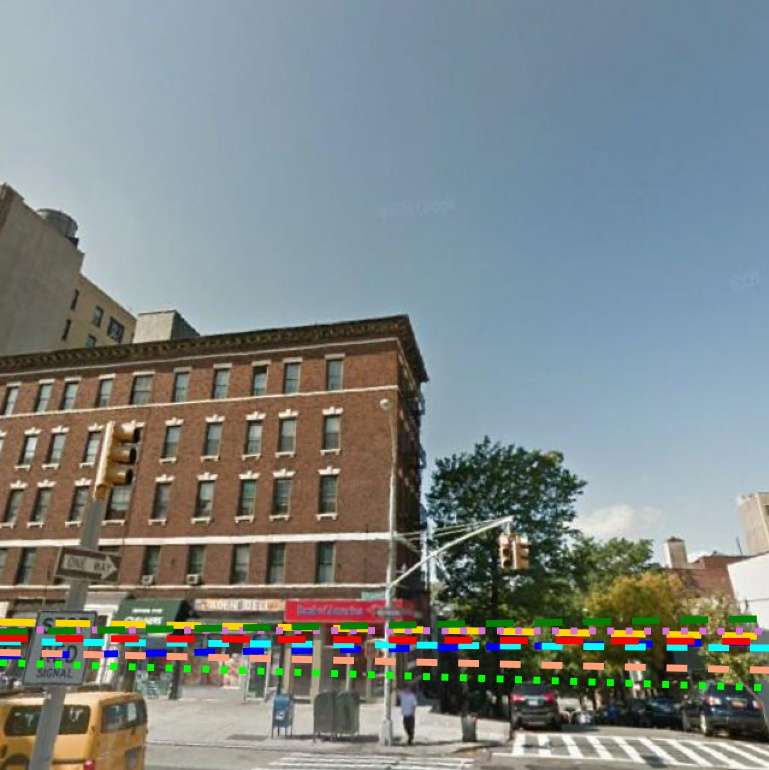} &
    \includegraphics[width=0.15\linewidth]{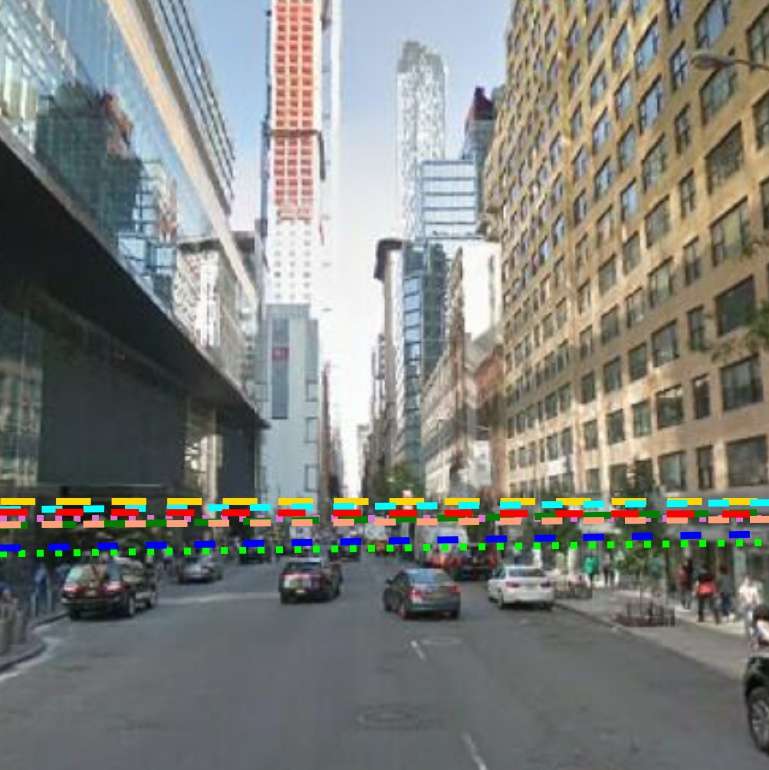} &
    \includegraphics[width=0.15\linewidth]{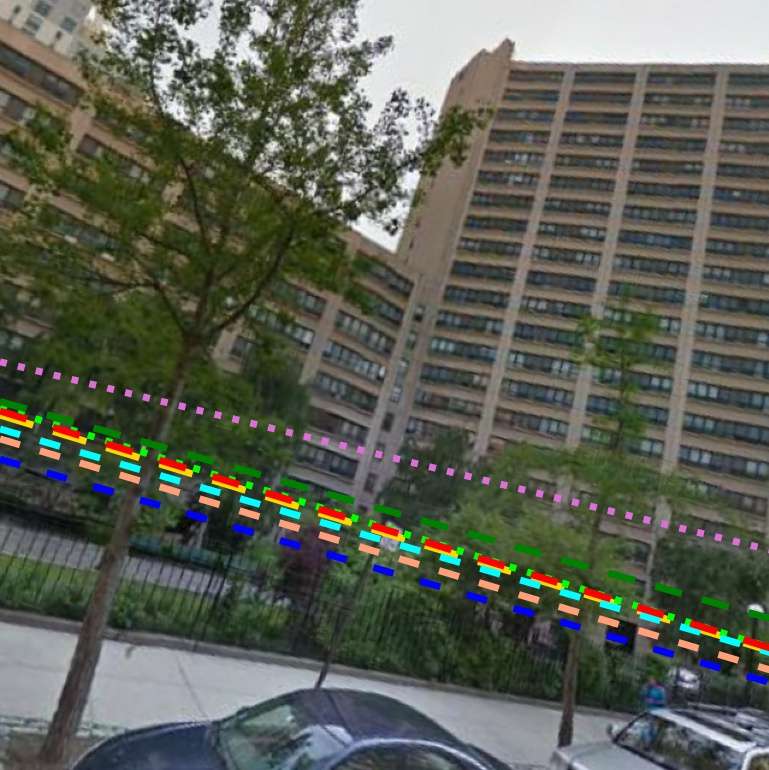} &
    \includegraphics[width=0.15\linewidth]{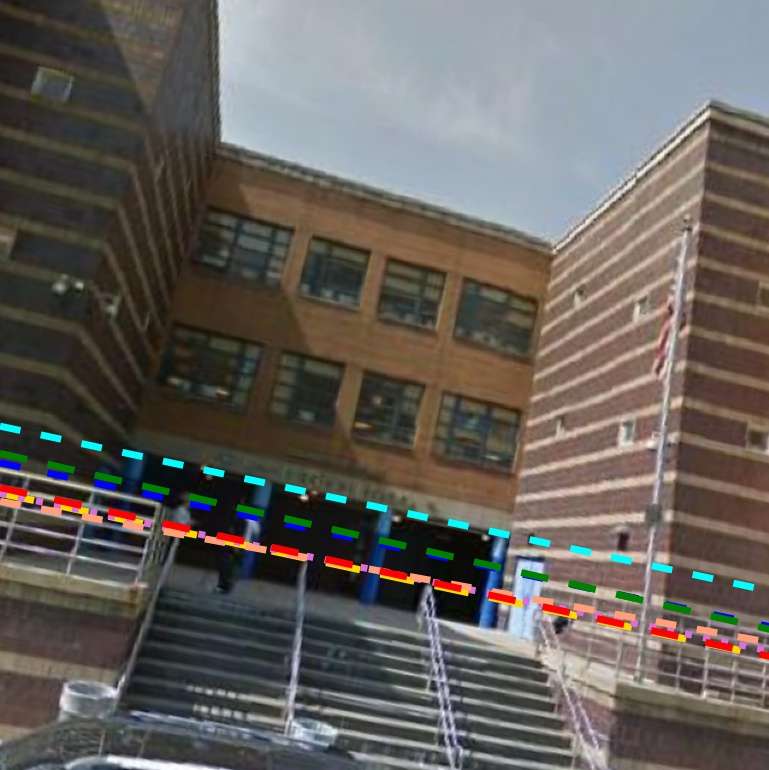} \\
    \includegraphics[width=0.15\linewidth]{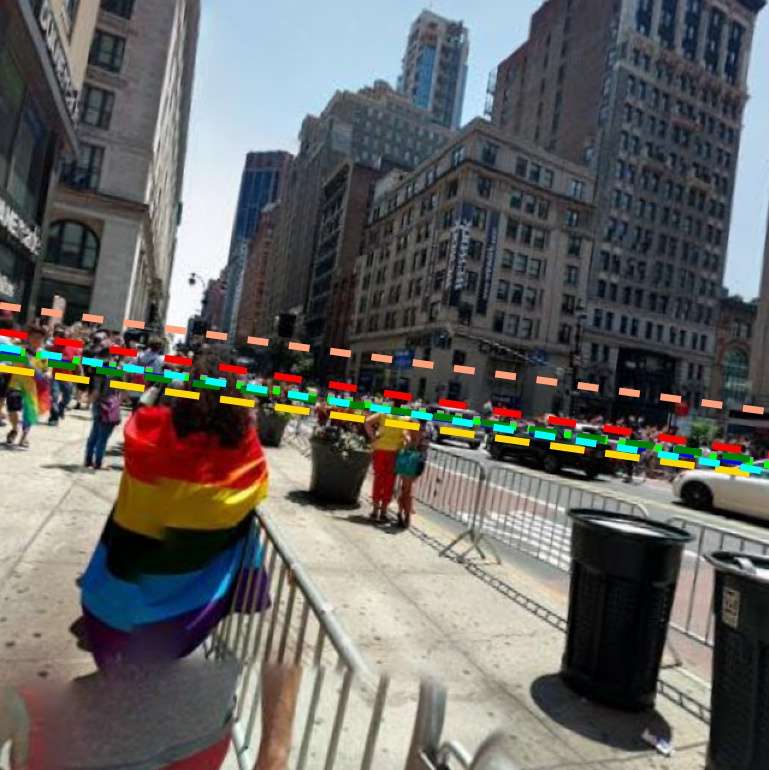} &
    \includegraphics[width=0.15\linewidth]{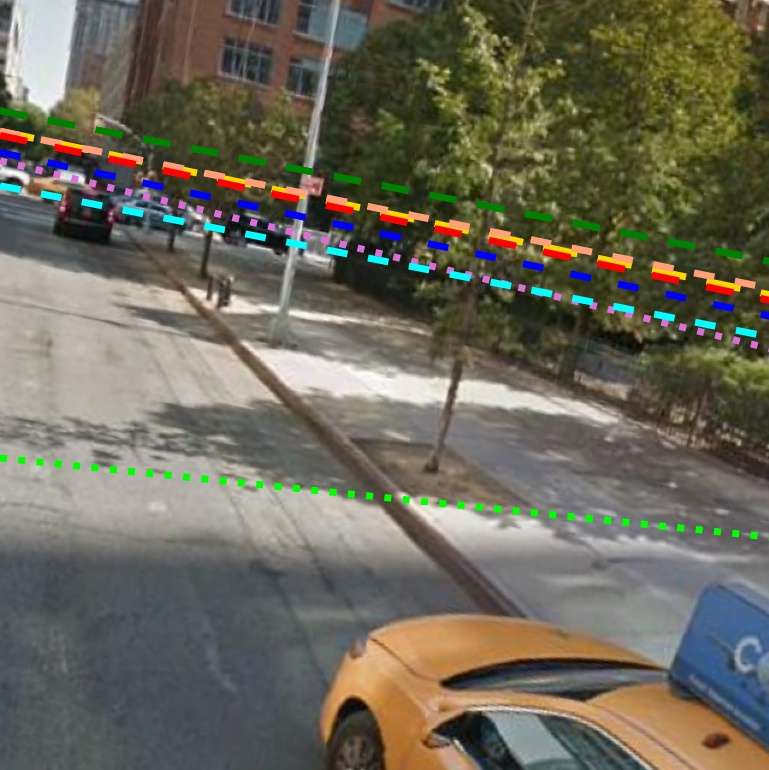} &
    \includegraphics[width=0.15\linewidth]{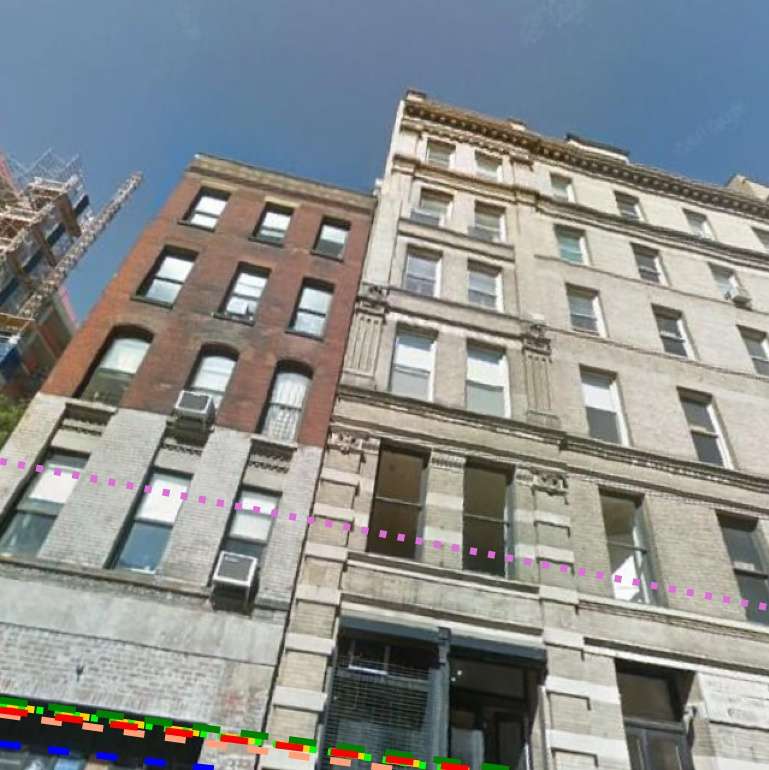} &
    \includegraphics[width=0.15\linewidth]{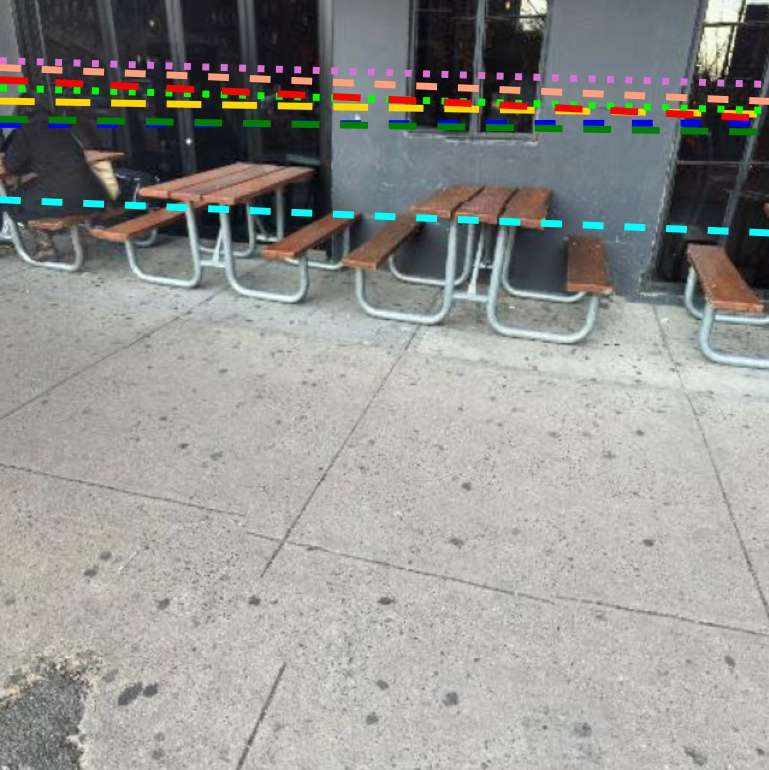} &
    \includegraphics[width=0.15\linewidth]{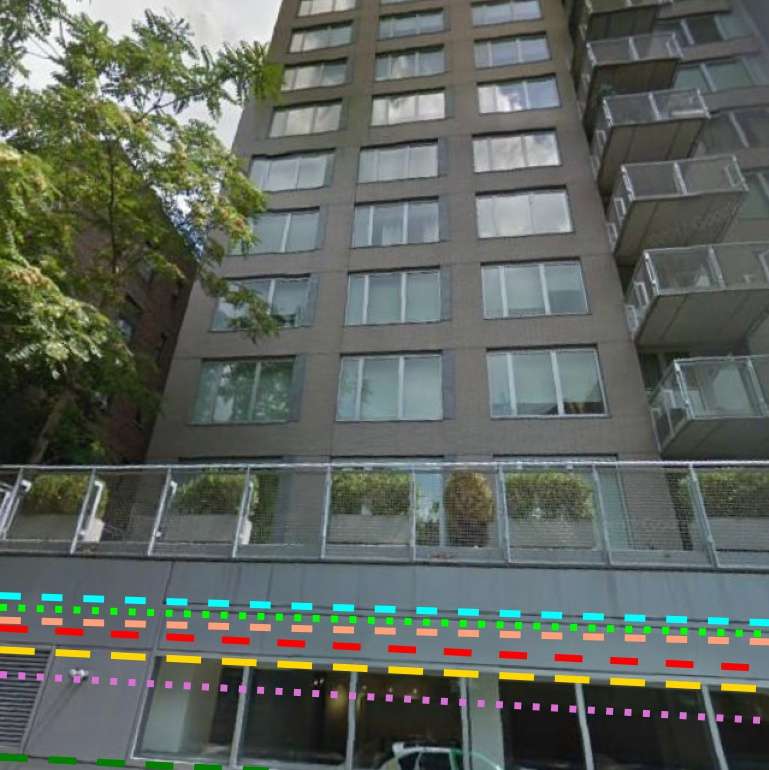} &
    \includegraphics[width=0.15\linewidth]{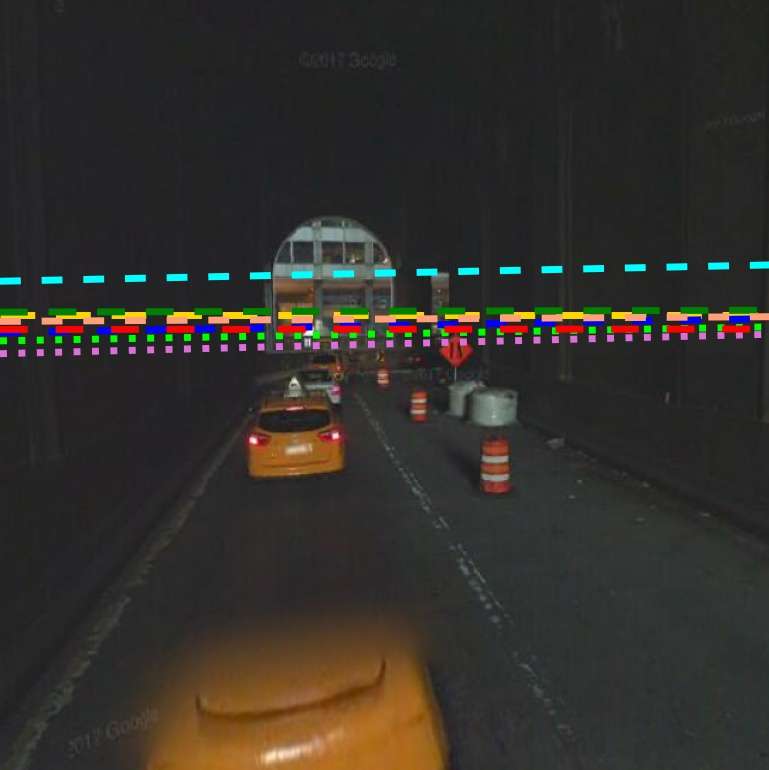} \\
    \end{tabular}
    \newcommand{\crule}[3][red]{\textcolor{#1}{\rule{#2}{#3} \rule{#2}{#3} \rule{#2}{#3} \rule{#2}{#3}}}
    {\scriptsize
    \begin{tabular}{llll}
    \crule[Goldenrod]{0.01\linewidth}{0.01\linewidth} Ground Truth & 
    \crule[Orchid]{0.01\linewidth}{0.01\linewidth} Upright \cite{Lee:2014} & 
    \crule[LimeGreen]{0.01\linewidth}{0.01\linewidth} A-Contario \cite{Simon:2018} & 
    \crule[blue]{0.01\linewidth}{0.01\linewidth} DeepHorizon \cite{Workman:2016} \\ 
    \crule[OliveGreen]{0.01\linewidth}{0.01\linewidth} Perceptual \cite{Hold-Geoffroy:2018} & 
    \crule[SkyBlue]{0.01\linewidth}{0.01\linewidth} GPNet \cite{Lee:2020:ECCV} &
    \crule[LightSalmon]{0.01\linewidth}{0.01\linewidth} ResNet &
    \crule[red]{0.01\linewidth}{0.01\linewidth} CTRL-C (Ours) \\
    \end{tabular}
    }
    \caption{Examples of horizon line prediction on the Google Street View~\cite{GSVI} test set.}
    \label{fig:horizon_line_predictions_gsv}
\end{figure*}

\begin{figure*}[h!]
    \centering
    \setlength\tabcolsep{1.5pt} 
    \begin{tabular}{ccccccc}
    (a) & 
    \includegraphics[width=0.15\linewidth]{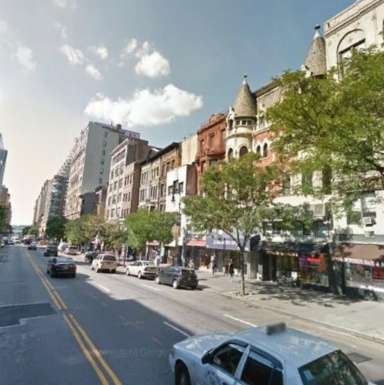} &
    \includegraphics[width=0.15\linewidth]{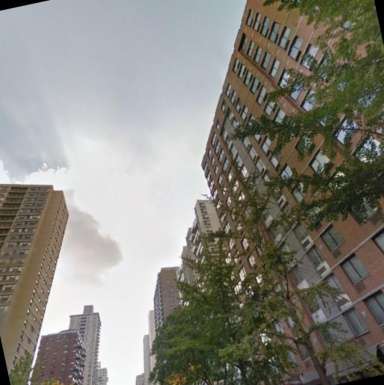} &
    \includegraphics[width=0.15\linewidth]{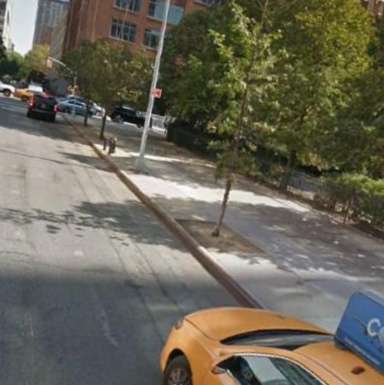} &
    \includegraphics[width=0.15\linewidth]{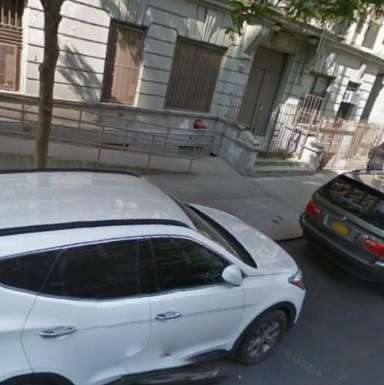} &
    \includegraphics[width=0.15\linewidth]{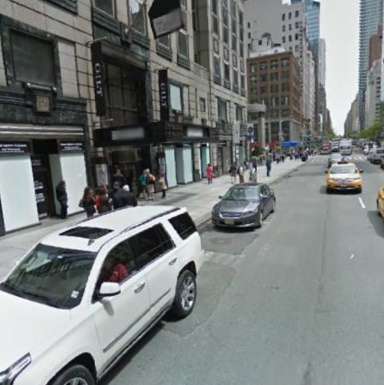} &
    \includegraphics[width=0.15\linewidth]{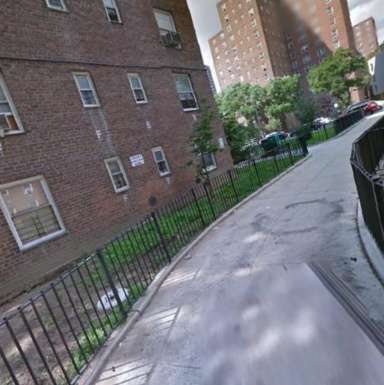} \\
    (b) &
    \includegraphics[width=0.15\linewidth]{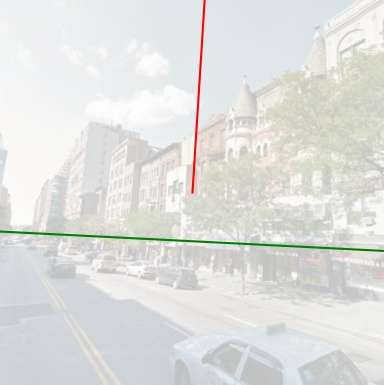} &
    \includegraphics[width=0.15\linewidth]{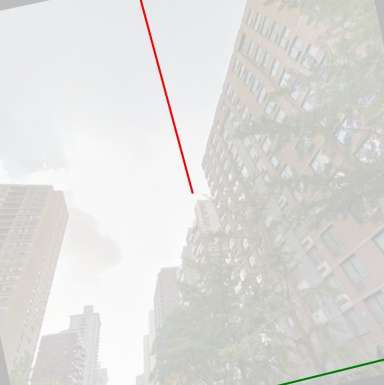} &
    \includegraphics[width=0.15\linewidth]{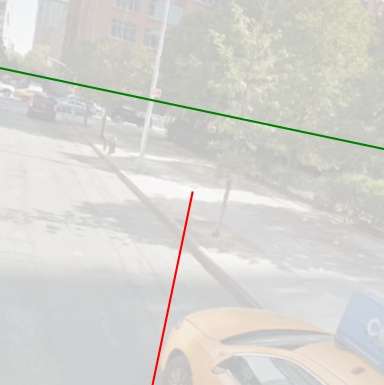} &
    \includegraphics[width=0.15\linewidth]{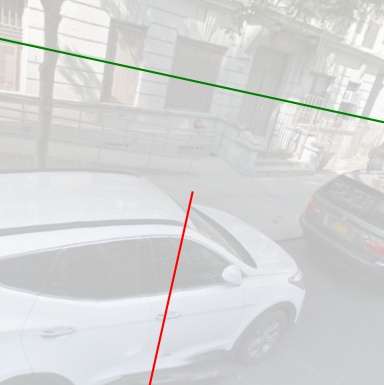} &
    \includegraphics[width=0.15\linewidth]{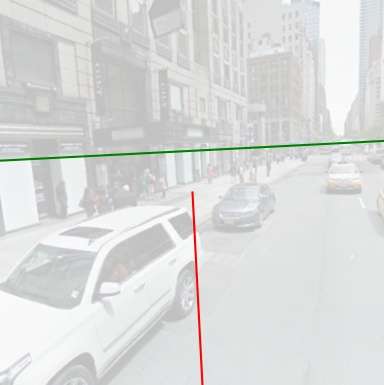} &
    \includegraphics[width=0.15\linewidth]{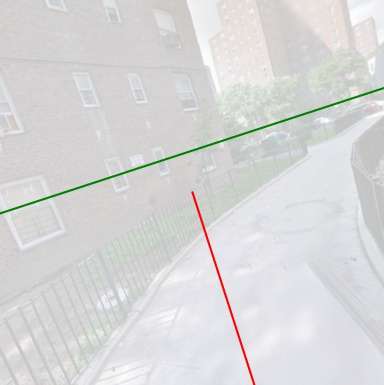} \\
    (c) &
    \includegraphics[width=0.15\linewidth]{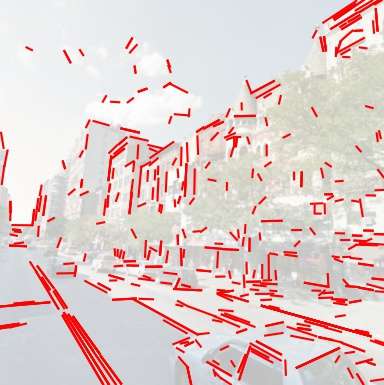} &
    \includegraphics[width=0.15\linewidth]{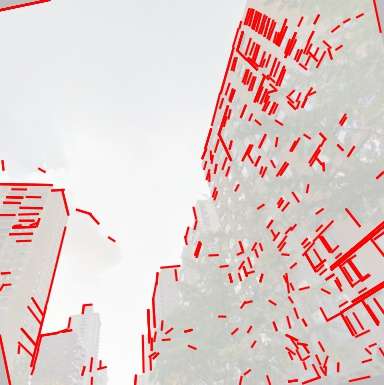} &
    \includegraphics[width=0.15\linewidth]{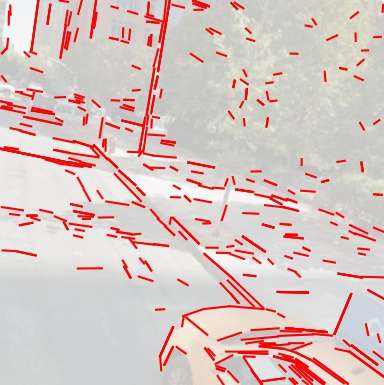} &
    \includegraphics[width=0.15\linewidth]{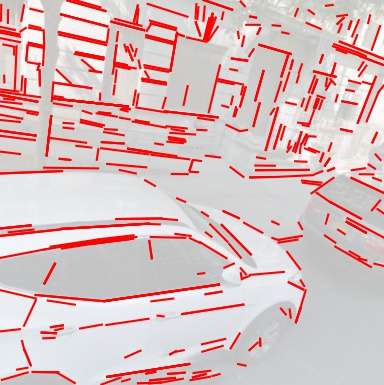} &
    \includegraphics[width=0.15\linewidth]{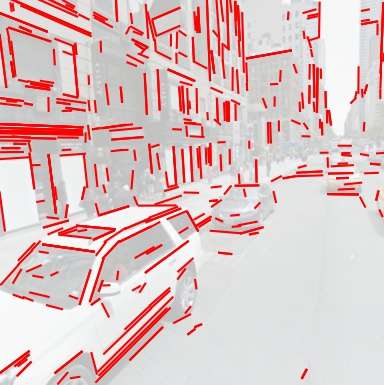} &
    \includegraphics[width=0.15\linewidth]{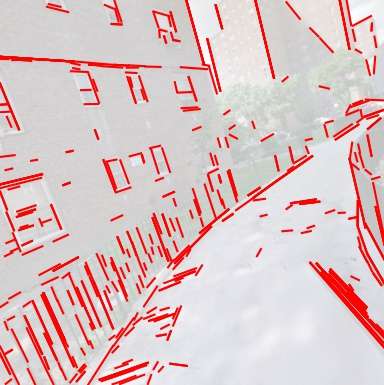} \\
    (d) & 
    \includegraphics[width=0.15\linewidth]{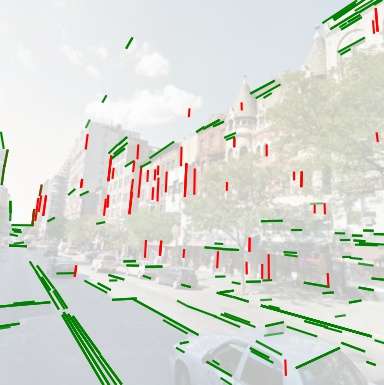} &
    \includegraphics[width=0.15\linewidth]{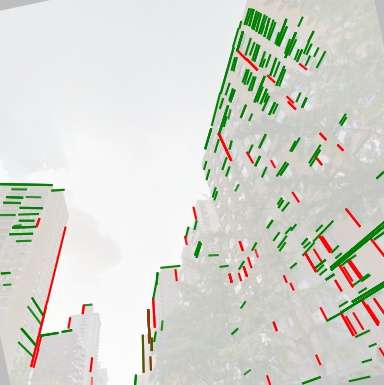} &
    \includegraphics[width=0.15\linewidth]{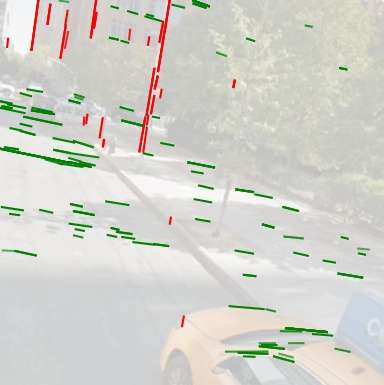} &
    \includegraphics[width=0.15\linewidth]{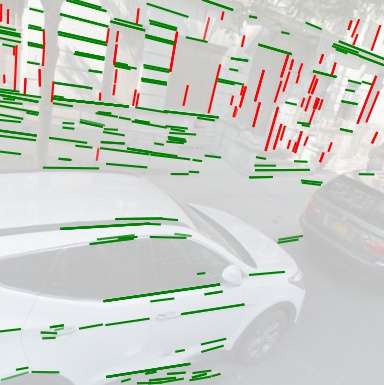} &
    \includegraphics[width=0.15\linewidth]{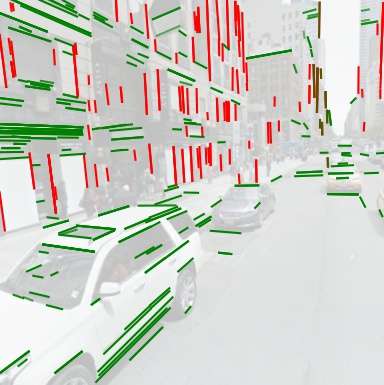} &
    \includegraphics[width=0.15\linewidth]{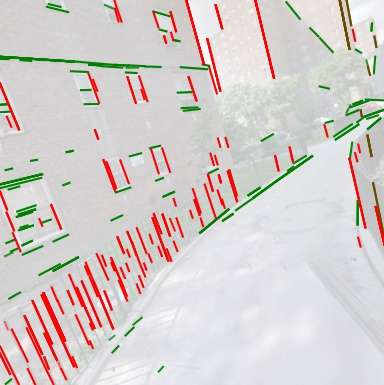} \\
    \end{tabular}
    \caption{More results with Google Street View~\cite{GSVI} test set: (a) input image, (b) estimated horizon line (green) and vertical direction along with the zenith VP (red), 
    (c) detected lines with LSD \cite{Gioi:2010}, 
    (d) estimated vertical (red) and horizontal (green) convergence line segments of (c).}
    \label{fig:additional_results_gsv}
\end{figure*}
\clearpage
\begin{figure*}[t!]
    \centering
    \setlength\tabcolsep{1.5pt} 
    \begin{tabular}{cccccc}
    \includegraphics[width=0.15\linewidth]{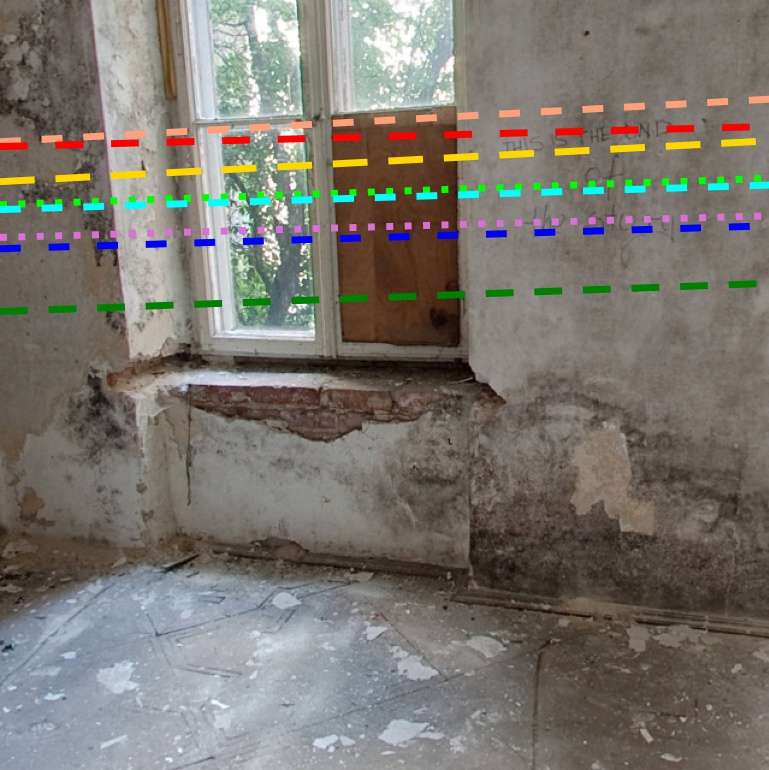} &
    \includegraphics[width=0.15\linewidth]{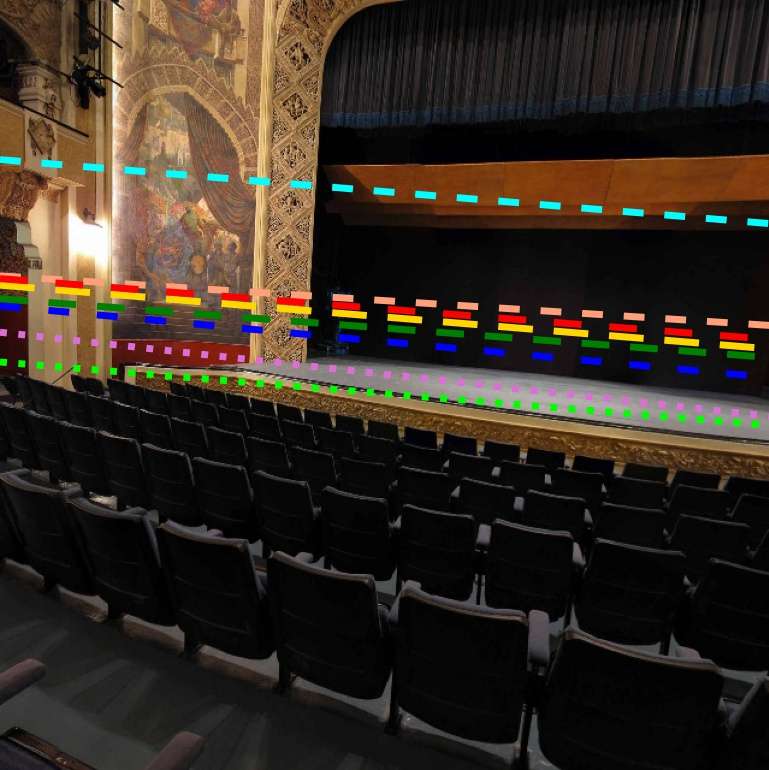} &
    \includegraphics[width=0.15\linewidth]{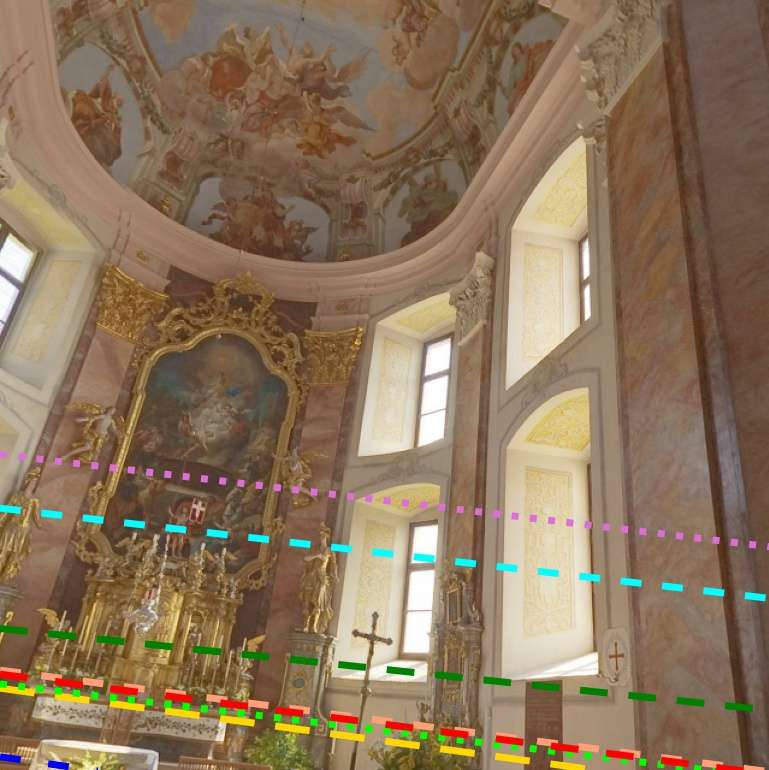} &
    \includegraphics[width=0.15\linewidth]{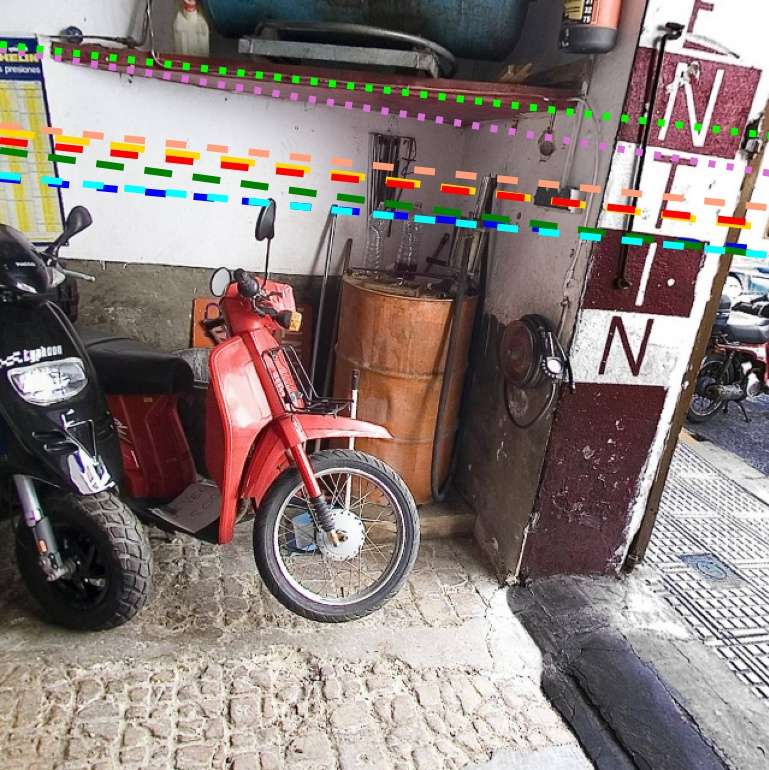} &
    \includegraphics[width=0.15\linewidth]{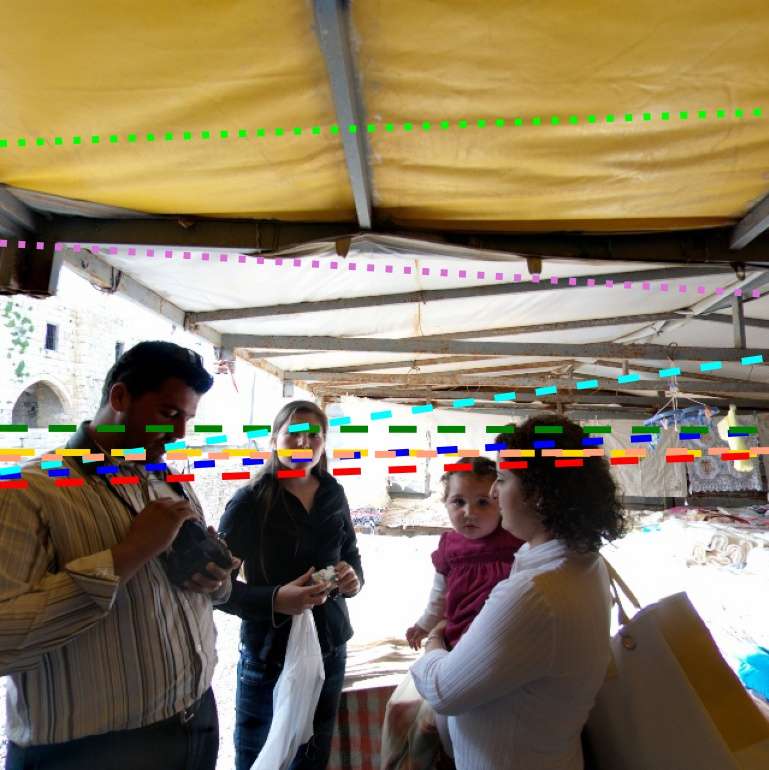} &
    \includegraphics[width=0.15\linewidth]{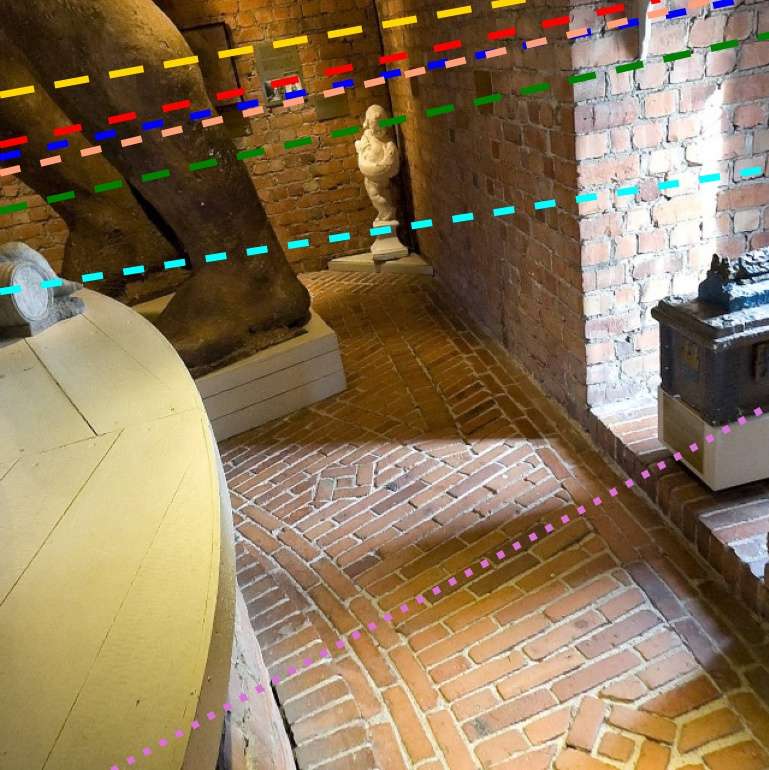} \\
    \includegraphics[width=0.15\linewidth]{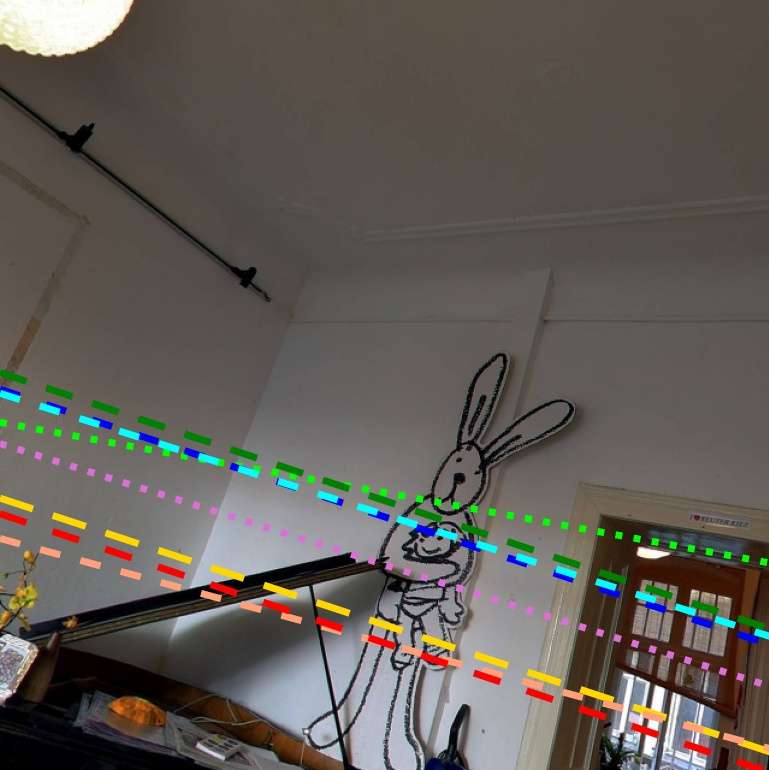} &
    \includegraphics[width=0.15\linewidth]{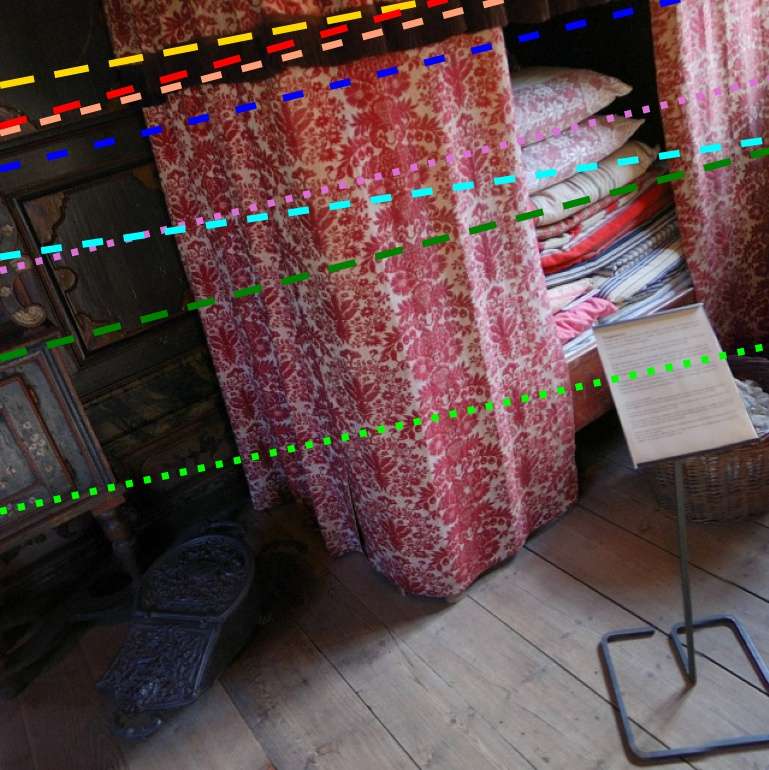} &
    \includegraphics[width=0.15\linewidth]{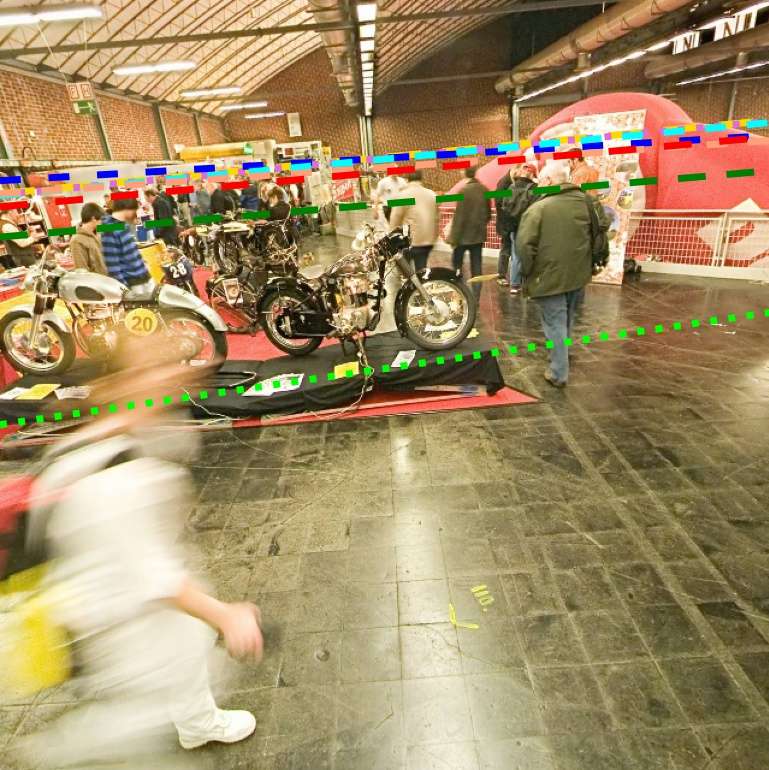} &
    \includegraphics[width=0.15\linewidth]{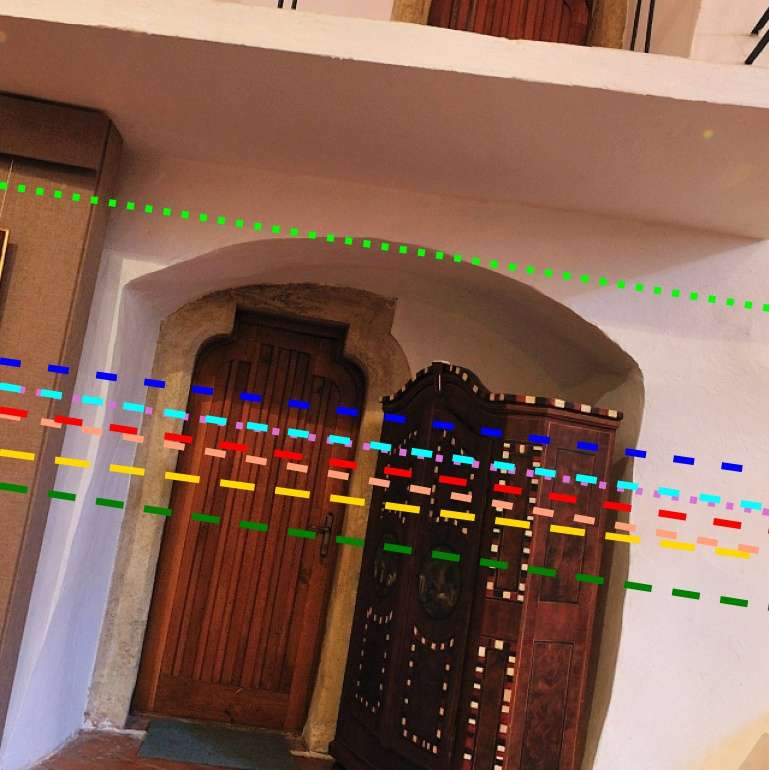} &
    \includegraphics[width=0.15\linewidth]{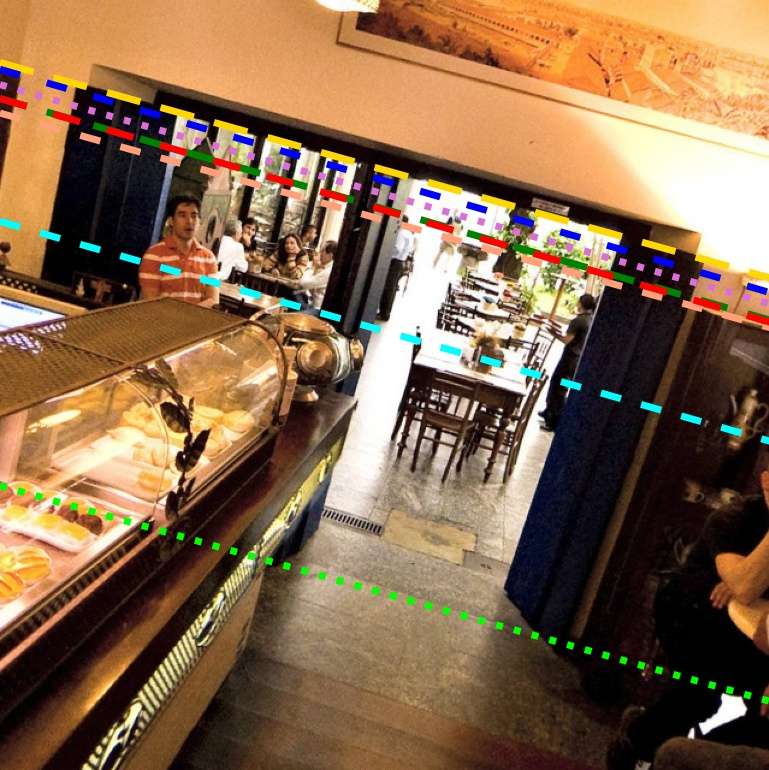} &
    \includegraphics[width=0.15\linewidth]{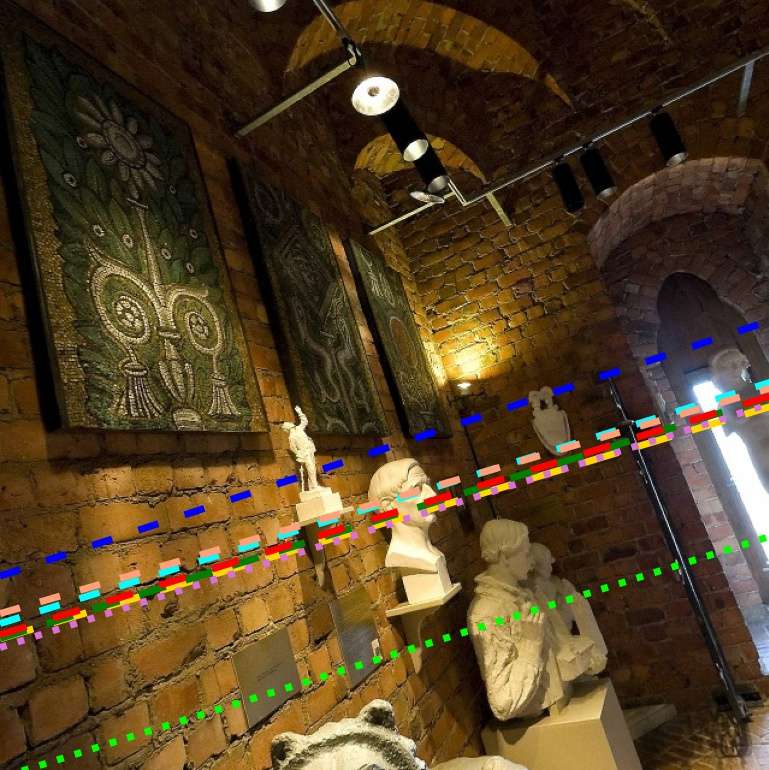} \\
    \includegraphics[width=0.15\linewidth]{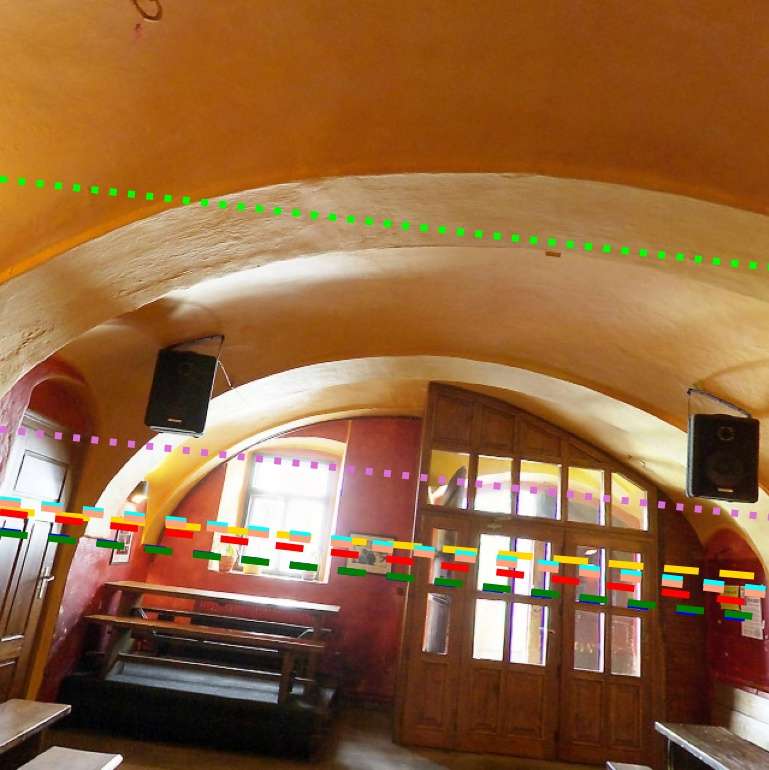} &
    \includegraphics[width=0.15\linewidth]{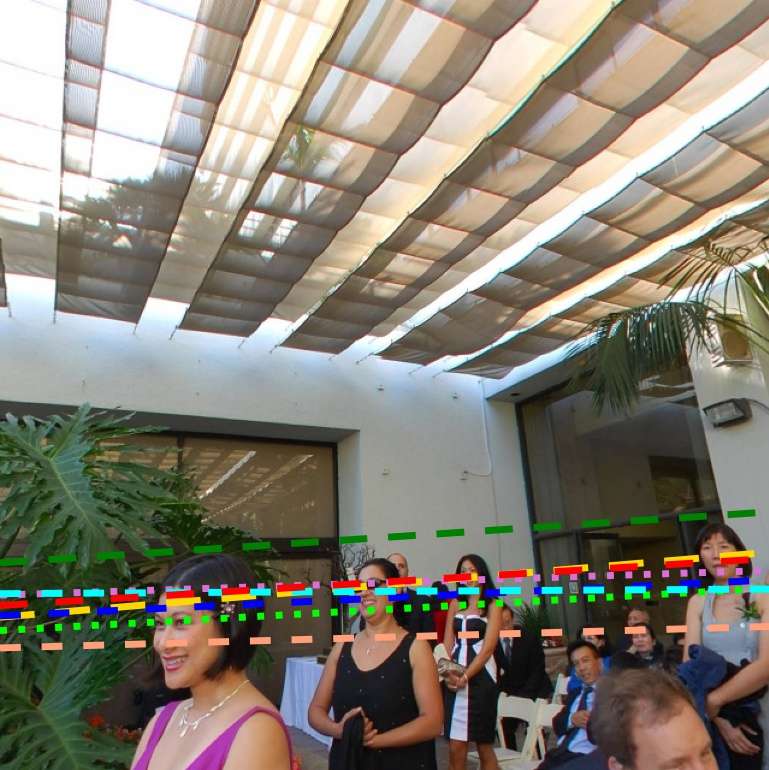} &
    \includegraphics[width=0.15\linewidth]{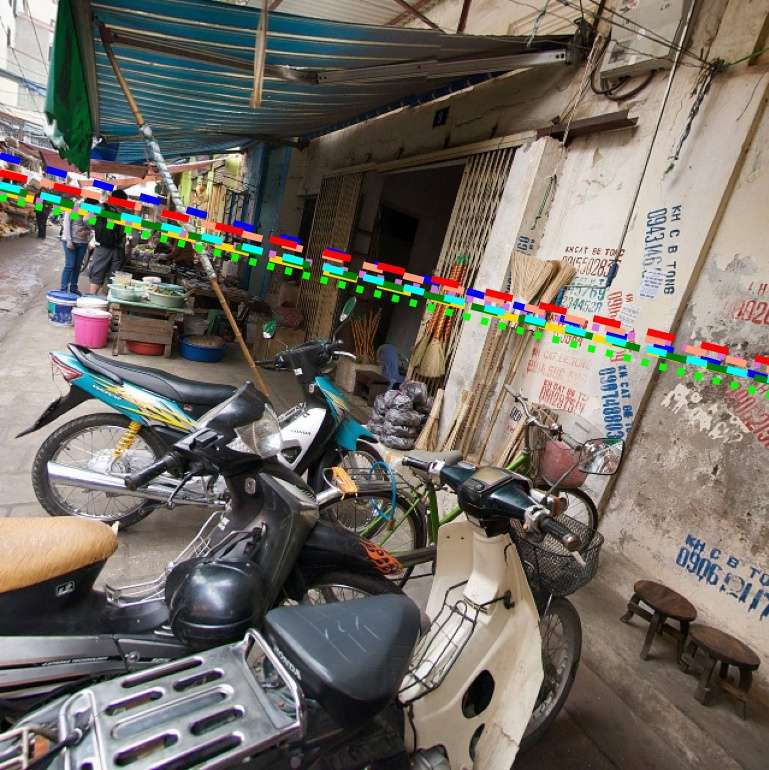} &
    \includegraphics[width=0.15\linewidth]{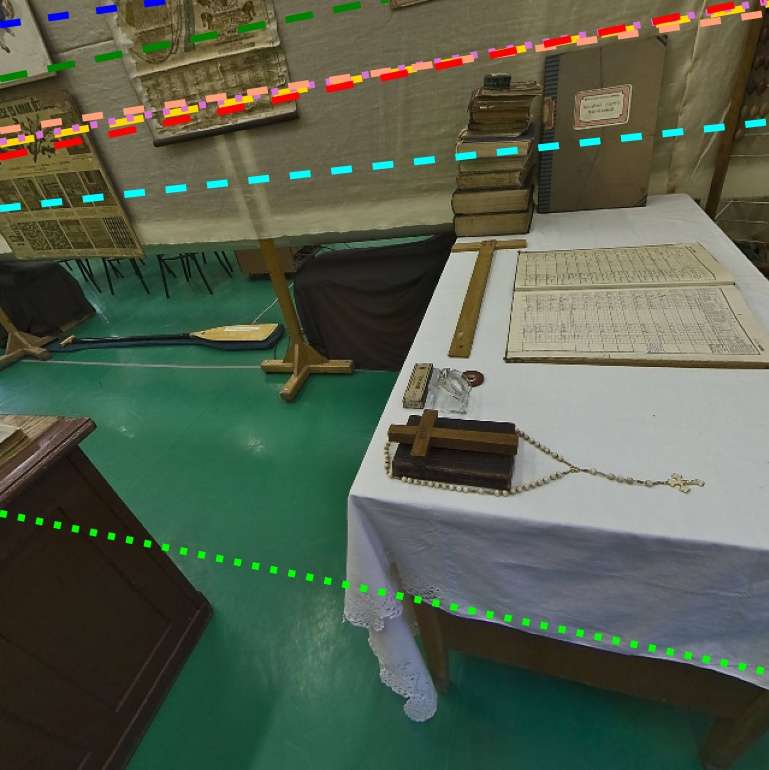} &
    \includegraphics[width=0.15\linewidth]{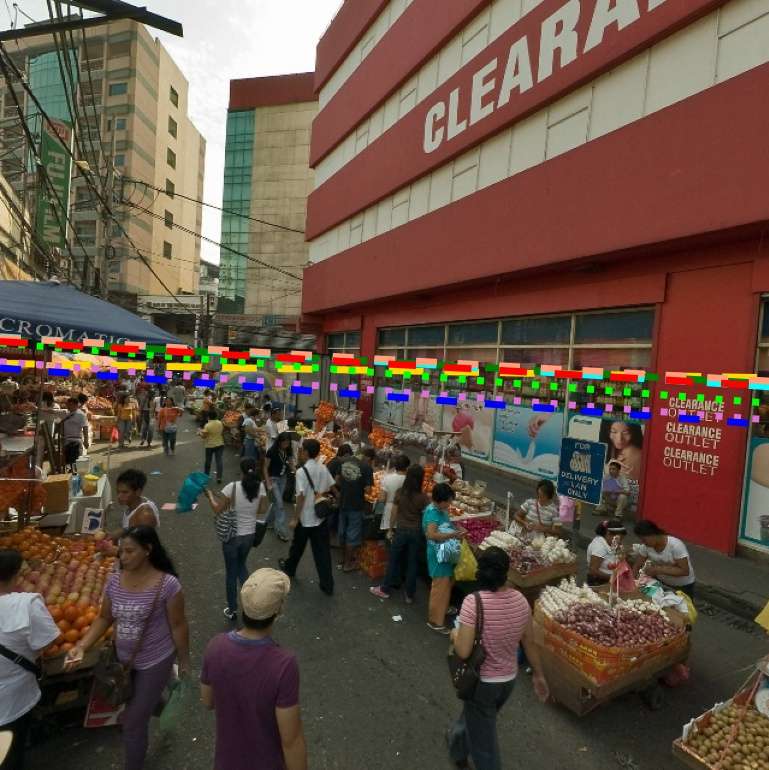} &
    \includegraphics[width=0.15\linewidth]{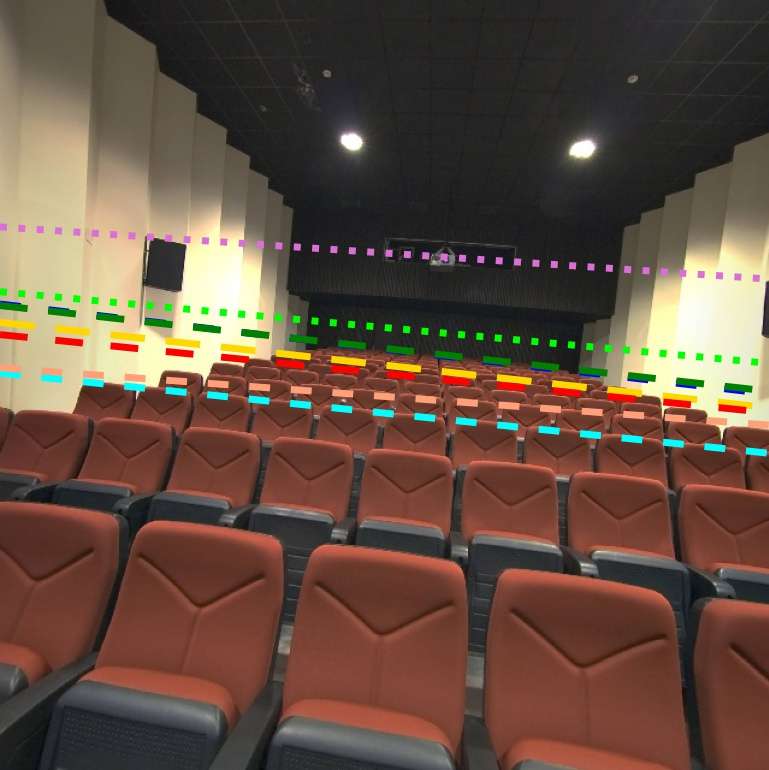} \\
    \end{tabular}
    \newcommand{\crule}[3][red]{\textcolor{#1}{\rule{#2}{#3} \rule{#2}{#3} \rule{#2}{#3} \rule{#2}{#3}}}
    {\scriptsize
    \begin{tabular}{llll}
    \crule[Goldenrod]{0.01\linewidth}{0.01\linewidth} Ground Truth & 
    \crule[Orchid]{0.01\linewidth}{0.01\linewidth} Upright \cite{Lee:2014} & 
    \crule[LimeGreen]{0.01\linewidth}{0.01\linewidth} A-Contario \cite{Simon:2018} & 
    \crule[blue]{0.01\linewidth}{0.01\linewidth} DeepHorizon \cite{Workman:2016} \\
    \crule[OliveGreen]{0.01\linewidth}{0.01\linewidth} Perceptual \cite{Hold-Geoffroy:2018} & 
    \crule[SkyBlue]{0.01\linewidth}{0.01\linewidth} GPNet \cite{Lee:2020:ECCV} &
    \crule[LightSalmon]{0.01\linewidth}{0.01\linewidth} ResNet &
    \crule[red]{0.01\linewidth}{0.01\linewidth} CTRL-C (Ours) \\
    \end{tabular}
    }
    \caption{Examples of horizon line prediction on the SUN360~\cite{SUN360:2012} test set.}
    \label{fig:horizon_line_predictions_sun360}
\end{figure*}

\begin{figure*}[h!]
    \centering
    \setlength\tabcolsep{1.5pt} 
    \begin{tabular}{ccccccc}
    (a) &
    \includegraphics[width=0.15\linewidth]{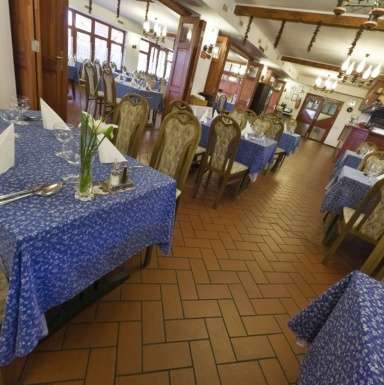} &
    \includegraphics[width=0.15\linewidth]{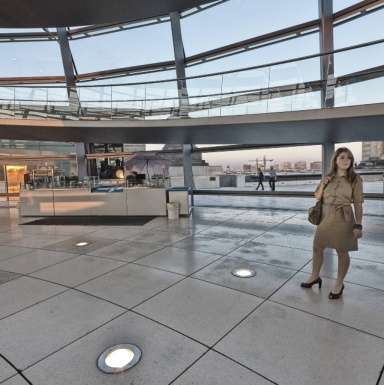} &
    \includegraphics[width=0.15\linewidth]{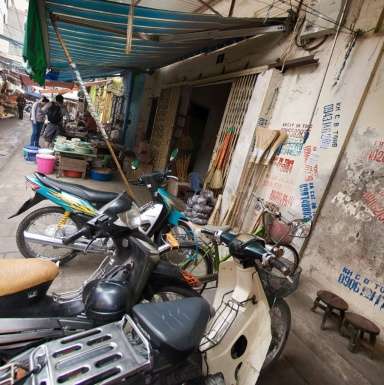} &
    \includegraphics[width=0.15\linewidth]{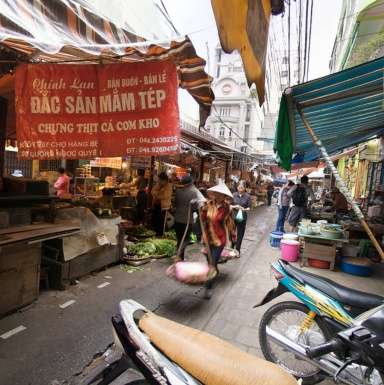} &
    \includegraphics[width=0.15\linewidth]{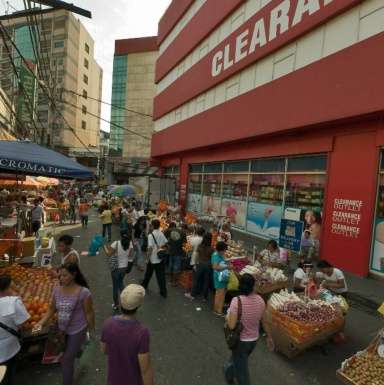} &
    \includegraphics[width=0.15\linewidth]{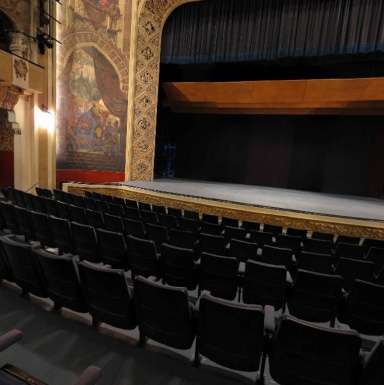} \\
    (b) &
    \includegraphics[width=0.15\linewidth]{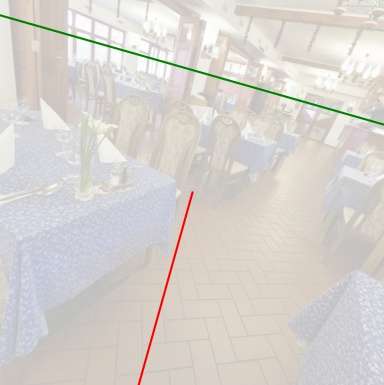} &
    \includegraphics[width=0.15\linewidth]{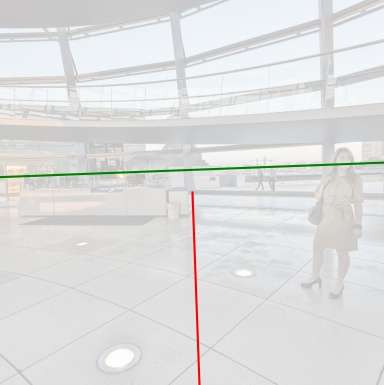} &
    \includegraphics[width=0.15\linewidth]{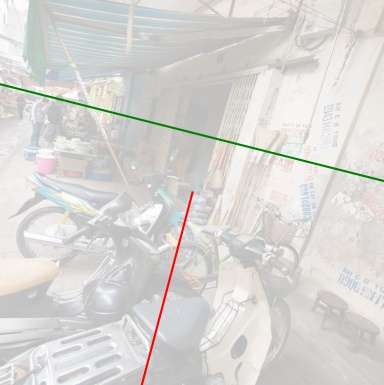} &
    \includegraphics[width=0.15\linewidth]{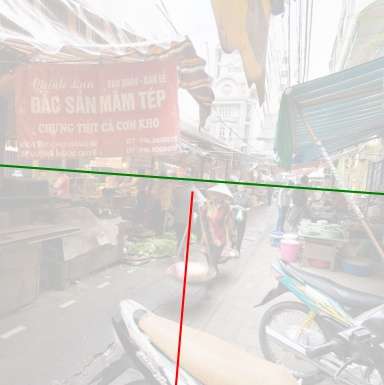} &
    \includegraphics[width=0.15\linewidth]{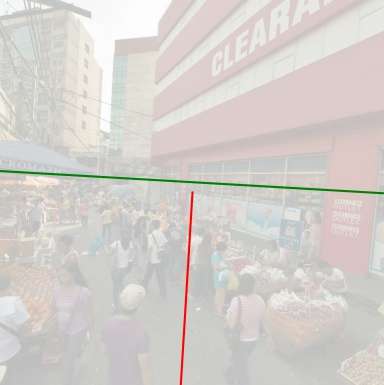} &
    \includegraphics[width=0.15\linewidth]{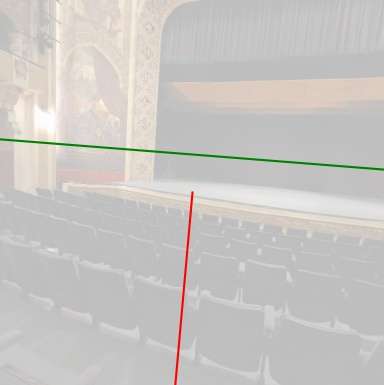} \\
    (c) &
    \includegraphics[width=0.15\linewidth]{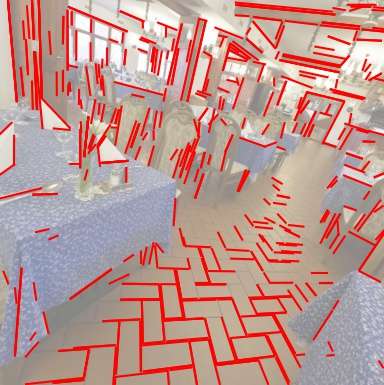} &
    \includegraphics[width=0.15\linewidth]{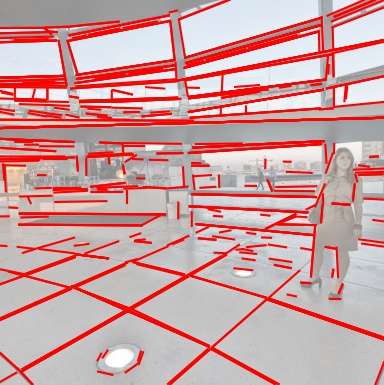} &
    \includegraphics[width=0.15\linewidth]{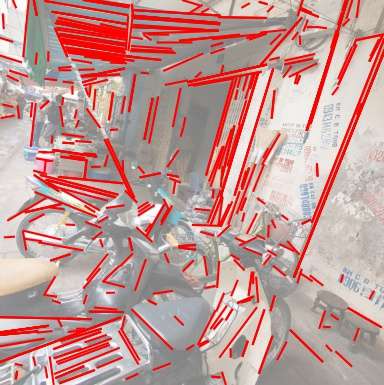} &
    \includegraphics[width=0.15\linewidth]{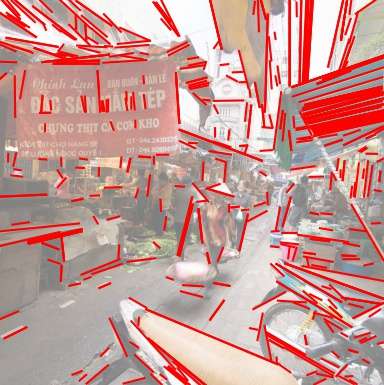} &
    \includegraphics[width=0.15\linewidth]{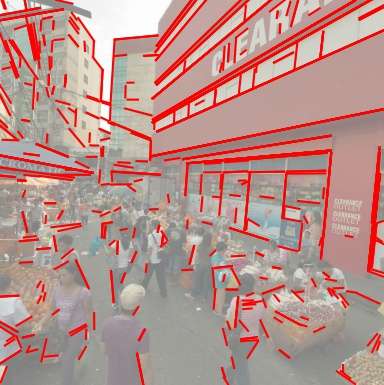} &
    \includegraphics[width=0.15\linewidth]{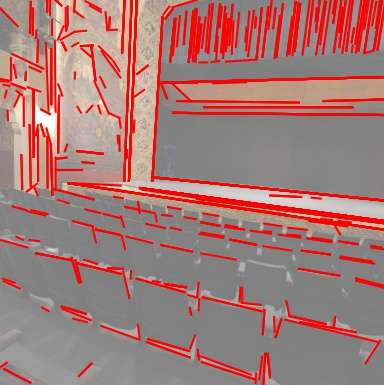} \\
    (d) &
    \includegraphics[width=0.15\linewidth]{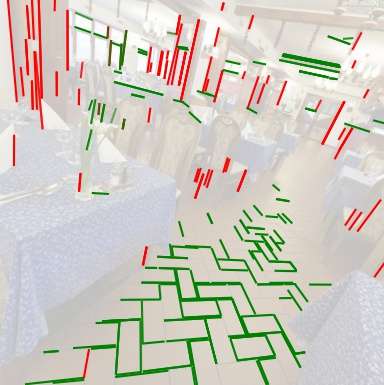} &
    \includegraphics[width=0.15\linewidth]{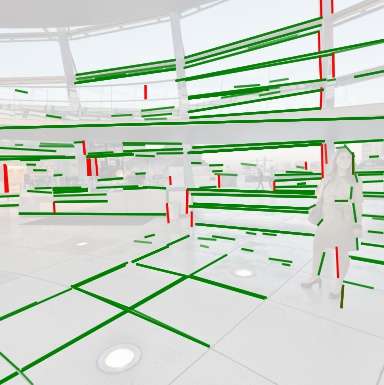} &
    \includegraphics[width=0.15\linewidth]{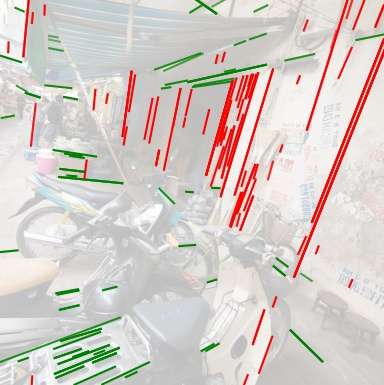} &
    \includegraphics[width=0.15\linewidth]{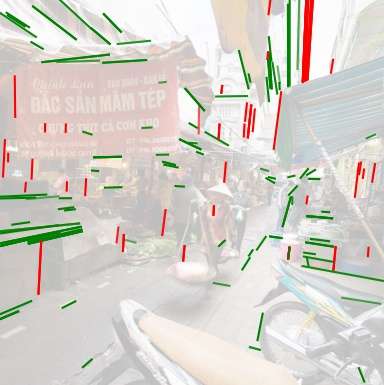} &
    \includegraphics[width=0.15\linewidth]{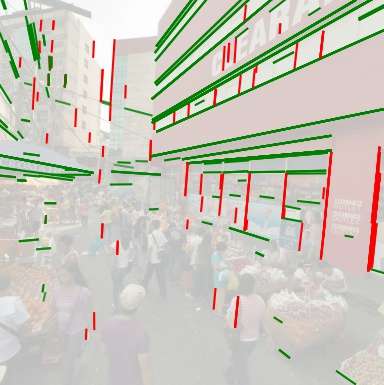} &
    \includegraphics[width=0.15\linewidth]{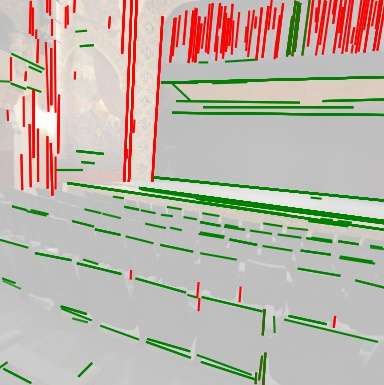} \\
    \end{tabular}
    \caption{More results with SUN360~\cite{SUN360:2012} test set: (a) input image, (b) estimated horizon line (green) and vertical direction along with the zenith VP (red), 
    (c) detected lines with LSD \cite{Gioi:2010}, 
    (d) estimated vertical (red) and horizontal (green) convergence line segments of (c).}
    \label{fig:additional_results_sun360}
\end{figure*}

\fi

\end{document}